\documentclass{article} 
\usepackage[accepted]{icml2024}

\usepackage{microtype}
\usepackage{graphicx}
\usepackage{booktabs} 
\usepackage{hyperref}

\usepackage{mathtools}
\usepackage[utf8]{inputenc} 
\usepackage[T1]{fontenc}    
\usepackage{url}            
\usepackage{amsfonts}       
\usepackage{nicefrac}       
\usepackage{xcolor}         
\usepackage{colortbl}
\usepackage{subcaption}
\usepackage{enumitem}
\usepackage{multirow}
\usepackage{capt-of}
\usepackage{wrapfig}
\usepackage{makecell}
\usepackage{bm}
\usepackage{arydshln}
\usepackage{placeins}
\hypersetup{
    colorlinks = true,
    citecolor = blue,
    linkcolor = red
}
\usepackage{comment}
\usepackage{algorithm}
\usepackage{algorithmic}
\usepackage{eqparbox}
\usepackage{amsmath, amsfonts, amsmath, amssymb, amsthm, bm}
\usepackage{pifont} 

\theoremstyle{plain}

\theoremstyle{definition}

\theoremstyle{remark}

\renewcommand{\algorithmiccomment}[1]{\hfill{ $\rhd$ #1}}

\newenvironment{sizeddisplay}[1]
 {\par\nopagebreak#1\noindent\ignorespaces}
 {\nopagebreak\ignorespacesafterend}

\newcommand{\overbar}[1]{\mkern 1.5mu\overline{\mkern-1.5mu#1\mkern-1.5mu}\mkern 1.5mu}

\newcommand{\greencheck}{\textcolor{green!80!black}{\ding{51}}}
\newcommand{\redx}{\textcolor{red}{\ding{55}}}

\usepackage{adjustbox}

\definecolor{gg}{gray}{0.92}
\newcolumntype{a}{>{\columncolor{gg}}c}
\definecolor{figred}{RGB}{192, 0, 0}
\definecolor{figblue}{RGB}{48, 132, 194}
\definecolor{figgreen}{RGB}{99, 154, 63}

\definecolor{rebuttal_edit}{RGB}{255,132,0}

\usepackage[capitalize,noabbrev]{cleveref}
\usepackage[textsize=tiny]{todonotes}
\usepackage[hang,flushmargin]{footmisc} 

\begin{document}
\twocolumn[

\icmltitle{Graph Generation with Diffusion Mixture}

\icmlsetsymbol{equal}{*}

\begin{icmlauthorlist}
\icmlauthor{Jaehyeong Jo}{kaist,equal}
\icmlauthor{Dongki Kim}{kaist,equal}
\icmlauthor{Sung Ju Hwang}{kaist,deepauto}
\end{icmlauthorlist}

\icmlaffiliation{kaist}{Korea Advanced Institute of Science and Technology (KAIST)}
\icmlaffiliation{deepauto}{DeepAuto.ai}

\icmlcorrespondingauthor{Jaehyeong Jo}{harryjo97@kaist.ac.kr}
\icmlcorrespondingauthor{Dongki Kim}{cleverki@kaist.ac.kr}
\icmlcorrespondingauthor{Sung Ju Hwang}{sjhwang82@kaist.ac.kr}

\icmlkeywords{Machine Learning, ICML}
\vskip 0.3in
]
\printAffiliationsAndNotice{\;\;\icmlEqualContribution}

\begin{abstract}
Generation of graphs is a major challenge for real-world tasks that require understanding the complex nature of their non-Euclidean structures. Although diffusion models have achieved notable success in graph generation recently, they are ill-suited for modeling the topological properties of graphs since learning to denoise the noisy samples does not explicitly learn the graph structures to be generated. To tackle this limitation, we propose a generative framework that models the topology of graphs by explicitly learning the final graph structures of the diffusion process. 
Specifically, we design the generative process as a mixture of endpoint-conditioned diffusion processes which is driven toward the predicted graph that results in rapid convergence. 
We further introduce a simple parameterization of the mixture process and develop an objective for learning the final graph structure, which enables maximum likelihood training.
Through extensive experimental validation on general graph and 2D/3D molecule generation tasks, we show that our method outperforms previous generative models, generating graphs with correct topology with both continuous (e.g. 3D coordinates) and discrete (e.g. atom types) features. Our code is available at https://github.com/harryjo97/GruM.
\end{abstract}

\section{Introduction}
\vspace{-0.075in}
Generation of graph-structured data has emerged as a crucial task for real-world problems such as drug discovery~\citep{drug-design/1}, protein design~\citep{protein-design/0}, and program synthesis~\citep{program-synthesis/0}. To tackle the challenge of learning the underlying distribution of graphs, deep generative models have been proposed, including models based on generative adversarial networks (GANs)~\citep{de2018molgan, martinkus22spectre}, recurrent neural networks (RNNs)~\citep{you2018graphrnn}, and variational autoencoders (VAEs)~\citep{jin18jtvae}. 

\begin{figure*}[t!]
    \centering
    \includegraphics[width=1\linewidth]{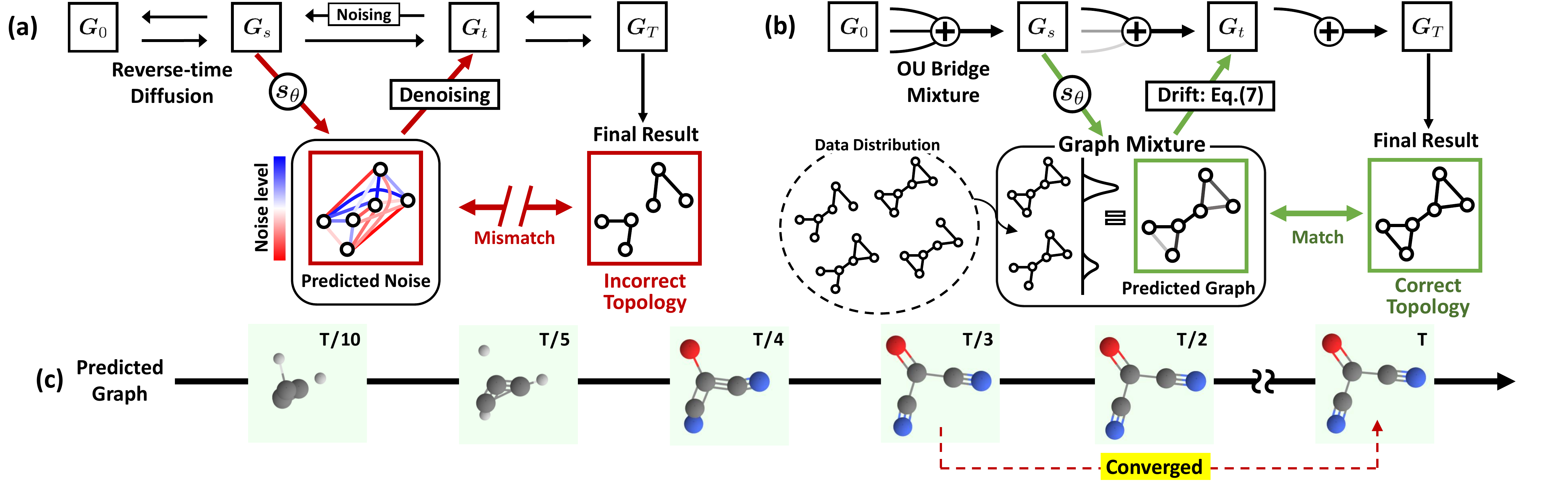}
    \vspace{-0.3in}
    \caption{\textbf{Illustration of the graph generative process.} 
    (a: Denoising diffusion model, b: GruM (ours), c: Graph mixture) For GruM, we design the generative process as a mixture of endpoint-conditioned diffusion processes (Eq.~\eqref{eq:ou_bridge}), namely the OU bridge mixture (Eq.~\eqref{eq:mixture}), which is driven toward the graph mixture (\textcolor{figgreen}{\textbf{green}}) by its drift (Eq.~\eqref{eq:drift_mixture}). Our GruM in (b) successfully generates graphs with valid topology by predicting the final result via learning the graph mixture as a weighted mean of data (Eq.~\eqref{eq:predicted_graph}). The predicted graph of GruM converges in an early stage to the correct topology as visualized in (c).
    In contrast, previous denoising diffusion models in (a) often fail to capture the correct topology as they learn the score or noise for denoising (\textcolor{figred}{\textbf{red}}), without explicit knowledge of final graph structure.}
    \label{fig:concept}
\vspace{-0.1in}
\end{figure*}

Recently, diffusion models have achieved state-of-the-art performance on the generation of graph-structured data~\citep{niu20edpgnn,jo22gdss,hoogeboom22edm}. These models learn the generation process as the time reversal of the forward process, which corrupts the graphs by gradually adding noise that destroys its topological properties. Since the generative process is derived from the unknown score function~\citep{song21sde} or noise~\citep{ho20ddpm}, existing graph diffusion models aim to estimate them in order to denoise the data from noise, which are commonly referred to as the \emph{denoising diffusion models} (Figure~\ref{fig:concept} (a)).

Despite their success, learning the score or noise is fundamentally ill-suited for the generation of graphs. In contrast to other types of data such as images, the key to generating valid graphs is accurately modeling the discrete structures that determine the topological properties such as connectivity or clusteredness. However, learning the score or noise does not explicitly model these features, as it aims to gradually denoise the corrupted structures. Thereby it is challenging for the diffusion models to recover the topological properties, which leads to failure cases even for small graphs. A way to more accurately generate graphs with correct topology would be directly learning the final graph and its structure, instead of learning how to denoise a noisy version of the original graphs.

However, predicting the final graph structure of the diffusion process is difficult since the prediction would be highly inaccurate in the early steps of the diffusion process, and such an inaccurate prediction may lead the process in the wrong direction resulting in invalid results. Few existing works~\citep{hoogeboom21argmax, austin21discrete, vignac22digress} based on denoising diffusion models aim to predict the probability of the final states by parameterizing the denoising process, but it is only applicable to categorical data with a finite number of states and thus cannot generate graphs with continuous features, which is not suitable for tasks such as 3D molecule generation.

To address these limitations of existing graph diffusion models, we propose a novel framework that explicitly models the graph topology, by learning the prediction of the resulting graph of the generative process which is represented as a weighted mean of graph data (Figure~\ref{fig:concept} (b)). 
Specifically, we construct a diffusion process via the mixture of endpoint-conditioned Ornstein-Uhlenbeck processes for which the drift drives the process in the direction of the predicted graph, differing from the denoising diffusion process used in previous works (Section~\ref{sec:method:design}). 
In order to model the mixture of the diffusion process, we develop a simple parameterization of the graph generative model with respect to the prediction of the final graph. 
We further derive an efficient training objective for learning the graph prediction, which guarantees to maximize the likelihood of our generative model (Section~\ref{sec:method:learn_mixture}). 
Thanks to its ability to capture accurate graph structures, our framework achieves fast convergence to the correct graph topology in an early sampling step (Figure~\ref{fig:concept} (c) and Figure~\ref{tab:3d_mol} (Right)).

We experimentally validate our method on diverse real-world graph generation tasks. We first validate it on general graph generation benchmarks with synthetic and real-world graphs, on which it outperforms previous deep graph generative models including graph diffusion models, by being able to generate valid graphs with correct topologies. We further validate our method on 2D and 3D molecule generation tasks to demonstrate its ability to generate graphs with both the continuous and discrete features, on which ours generates a significantly larger number of valid and stable molecules compared to the state-of-the-art baselines.
Our main contributions can be summarized as follows:
\vspace{-0.05in}
\begin{itemize}[itemsep=0.5mm, parsep=3pt, leftmargin=*]
    \item We observe that previous diffusion models cannot accurately model the graph structures as they learn to denoise at each step without considering the topology of the graphs to be generated.
    \item To fix such a myopic behavior of previous diffusion models, we propose a new graph generation framework that captures the graph structures by directly predicting the final graph of the diffusion process modeled by a mixture of endpoint-conditioned diffusion processes.
    \item We develop a simple parameterization of the graph generative model for modeling the mixture process and present a simulation-free training objective for graph prediction.
    \item Our method significantly outperforms previous graph diffusion models on the generation of diverse real and synthetic graphs, as well as on 2D/3D molecule generation tasks, by being able to generate graphs with accurate topologies, and both the discrete and continuous features.
\end{itemize}

\vspace{-0.1in}
\section{Related Work}

\paragraph{Diffusion Models}
Diffusion models have been shown to successfully generate high-quality samples from diverse data domains such as images~\citep{diffusion/images/1, diffusion/images/3} and videos~\citep{diffusion/video/1}. Despite their success, existing diffusion models for graphs~\citep{niu20edpgnn,jo22gdss} often fail to generate graphs with correct structures since they learn to estimate the score or noise for the denoising process which does not explicitly capture the final graph and its structure. To address these limitations, we propose a graph diffusion framework that models the generative process as a mixture of diffusion processes, which learns to predict the final graph with valid topology instead of predicting the denoising function at each step. This promotes our generative process to be driven toward the prediction of the final graph, resulting in generation of valid graphs with correct topology.

\vspace{-0.1in}
\paragraph{Diffusion Bridge Process}
A line of recent works has improved the generative framework of diffusion models by leveraging the diffusion bridge processes, i.e., processes conditioned to the endpoints. Schr\"odinger Bridge~\citep{Bortoli21dsb, chen22fbsde} aims to find both the forward and the backward process that transforms two distributions back and forth using iterative proportional fittings that require heavy computations. More recent works~\citep{peluchetti21mixture, wu22bridge, liu23bridge, jo2023riemannian} consider learning the generation process as a mixture of diffusion processes instead of reversing the noising process as in denoising diffusion models, which we describe in detail in Appendix~\ref{sec:app:derivation:comparison_bridge}. 
However, previous works aim to approximate the drifts of the diffusion processes which cannot accurately capture the discrete structure of graphs as it does not explicitly learn the graph structures to be generated. Moreover, learning the drift could be problematic as the drift of the diffusion process diverges near the terminal time. Instead, we present a new approach to parameterizing the mixture process with the prediction of the final graph which allows it to model valid graph topology.

\vspace{-0.1in}
\paragraph{Graph Generative Models}
Deep generative models for graphs either generate nodes and edges in an autoregressive manner or all the nodes and edges at once using VAE~\citep{jin18jtvae}, RNN~\citep{you2018graphrnn}, normalizing flow~\citep{zang2020moflow, shi2020graphaf, luo2021graphdf}, and GAN~\citep{de2018molgan, martinkus22spectre}. However, these models show poor performance due to restrictive model architectures for modeling the likelihood or their inability to model the permutation equivariant nature of graphs. Recently, diffusion models for graphs~\citep{jo22gdss, hoogeboom22edm, vignac22digress} have made large progress, but either fail to capture the graph topology or are not applicable to general tasks due to the architectural restriction of the framework. 
In our work, we introduce a diffusion framework that predicts the final graph structure instead of denoising noisy graphs. Our method largely outperforms existing models~\citep{jo22gdss, vignac22digress, hoogeboom22edm} on generation tasks including general graphs as well as 2D and 3D molecules.

\section{Graph Diffusion Mixture}
\vspace{-0.05in}
In this section, we present our graph generation framework \underbar{\textbf{Gr}}aph Diff\underbar{\textbf{u}}sion \underbar{\textbf{M}}ixture (GruM), for modeling valid topology of graphs using a mixture of diffusion processes.

\subsection{Designing Graph Generative Process \label{sec:method:design}}
The key to generating graph-structured data is understanding the underlying topology of graphs which is crucial to determining its validity, since a slight modification in the edges may significantly change its structure and the attributes, for example, planarity or molecular properties. However, previous diffusion models fail to do so as their objective is to denoise the noisy graphs, in which the topology is only implicitly captured (Figure~\ref{fig:concept} (a)) from the noisy structure. 
To overcome the limitation, we propose a graph diffusion framework that can directly learn the accurate graph structures and capture valid topology.

Throughout the paper, we represent a graph with $N$ nodes as a pair $\bm{G}\!=\!(\bm{X},\! \bm{A})$ where $\bm{X}\!\in\! \mathbb{R}^{N\!\times\! F}$ is the node features of feature dimension $F$ and $\bm{A}\!\in\! \mathbb{R}^{N\times\! N}$ is the weighted adjacency matrix that defines the connection between nodes.

\paragraph{Graph Mixture}
Our goal is to directly predict the final graph of the diffusion process that transports a prior distribution to the data distribution $\Pi^{\ast}\!$. To be specific, for a graph diffusion process represented as a trajectory of random variables $\{\bm{G}_{\tau}\!=\!(\bm{X}_{\tau},\! \bm{A}_{\tau})\}_{\tau\in[0,T]}$, we aim to predict the terminus of the process $\bm{G}_T$ in $\Pi^{\ast}\!$ given the current state $\bm{G}_{t}$. However, identifying the exact $\bm{G}_T$ at the early stage of the process is problematic since the prediction based on $\bm{G}_{t}$ of almost no information would be highly inaccurate, and could lead the process in the wrong direction.

To address this problem, we present a different approach to predicting the probable graph, which we define as a weighted mean of all the possible final results (Figure~\ref{fig:concept} (b)). Since the probability of a graph $\bm{g}$ being the final result is equal to the transition probability of the process denoted as $p_{T|t}(\bm{g}|\cdot)$, we define the probable graph given the current state $\bm{G}_t$ via the expectation of the graphs as follows:
\begin{align}
    \bm{D}(\bm{G}_t,t) = \int \bm{g} \cdot p_{T|t}(\bm{g}|\bm{G}_t) \; \mathrm{d}\bm{g} ,
    \label{eq:predicted_graph}
\end{align}
which we refer to as the \emph{graph mixture} of the process, visualized in Figure~\ref{fig:concept}. In order to explicitly model this, we construct a generative process driven toward the graph mixture using a mixture of diffusion processes, which we describe in the following paragraphs.

\paragraph{Ornstein-Uhlenbeck Bridge Process \label{sec:method:ou}}
As a building block of our generative framework, we leverage diffusion processes with fixed endpoints, namely the \emph{diffusion bridge} processes. 
We propose to use a family of bridge processes, namely the \emph{Ornstein-Uhlenbeck} (OU) bridge process that enables simulation-free training for our generative model while providing flexibility for modeling the complex generative process for graphs. 

Given an OU process $\mathbb{Q}$ modeled by the following SDE:
\begin{equation}
    \mathbb{Q}\!:\; \mathrm{d}\bm{G}_t = \alpha\sigma_t^2\bm{G}_t\mathrm{d}t + \sigma_t\mathrm{d}\mathbf{W}_t ,
    \label{eq:ou_process}
\end{equation}
where $\alpha$ is a constant, $\sigma_t$ is a noise schedule, and $\mathbf{W}_t$ is the standard Wiener process, the OU bridge process $\mathbb{Q}^{\bm{g}}$ is the process $\mathbb{Q}$ pinned at a fixed terminal point $g$.
Using the Doob's h-transform~\citep{doob1984h-transform}, we can derive the OU bridge process $\mathbb{Q}^{\bm{g}}$ as follows (we provide detailed derivation of the bridge process in Appendix~\ref{sec:app:derivation:bridge}):
\begin{align}
    \mathrm{d}\bm{G}_t =\! \bigg[ \underbrace{\alpha\sigma_{t}^2\bm{G}_t + \frac{\sigma_t^2}{v_t} \!\left( \frac{\bm{g}}{u_t}  - \bm{G}_t \right)}_{\eta^{\bm{g}}(\bm{G}_t,t)} \bigg]\mathrm{d}t + \sigma_t\mathrm{d}\mathbf{W}_t ,
    \label{eq:ou_bridge}
\end{align}
where the scalar functions $u_t$ and $v_t$ are defined as follows:
\begin{align}
    u_t = \exp\Big(\alpha\int^T_t\sigma_{\tau}^2\mathrm{d}\tau\Big) \;,\; v_t = \frac{1}{2\alpha}\left( 1-u_t^{-2} \right) .
    \label{eq:drift_scalar_functions}
\end{align}
The endpoint of $\mathbb{Q}^{\bm{g}}$ is fixed to $\bm{G}_T\!=\!\bm{g}$, since the drift $\eta^{\bm{g}}(\cdot,t)$ of the process forces the trajectory $\bm{G}_t$ in the direction of $\bm{g}$. Although there exists a more general class of bridge processes with non-linear drift (see Appendix~\ref{sec:app:derivation:bridge}), they have intractable transition probability and require expensive SDE simulation to obtain trajectories. In contrast, the OU bridge processes yield tractable transition probabilities due to their affine nature and allow the training of our generative model to be simulation-free, which we further discuss in Section~\ref{sec:method:objective}. Note that the Brownian bridge process used in previous works~\citep{wu22bridge,liu22bridge} is a special case of the OU bridge process when $\alpha\rightarrow 0$ (see Appendix~\ref{sec:app:derivation:bridge}). Especially, we can write the OU bridge process of Eq.~\eqref{eq:ou_bridge} for graphs $\bm{g}\!=\!(\bm{x},\! \bm{a})$ as a system of SDEs:
\begin{align}
\hspace*{-2mm}
\begin{cases}
    \! \mathrm{d}\bm{X}_t \!=\! \left[ \alpha_{1}\sigma_{1,t}^2 \bm{X}_t \!+\! \frac{\sigma_{1,t}^2}{v_{1,t}} \! \left(\frac{\bm{x}}{u_{1,t}} \!-\! \bm{X}_t \right) \right]\! \mathrm{d}t \!+\! \sigma_{1,t} \mathrm{d}\mathbf{W}_{\!1,t} \\[5pt]
    \! \mathrm{d}\bm{A}_t \!=\! \left[ \alpha_{2}\sigma_{2,t}^2 \,\bm{A}_t \!+\! \frac{\sigma_{2,t}^2}{v_{2,t}} \! \left(\frac{\bm{a}}{u_{2,t}} \!-\! \,\bm{A}_t \right)  \right]\! \mathrm{d}t \!+\! \sigma_{2,t} \mathrm{d}\mathbf{W}_{\!2,t}
\end{cases}
\hspace*{-5mm}
\label{eq:graph_ou_bridge}
\end{align}
With the OU bridge processes in hand, we develop a framework for predicting the final graph via the graph mixture.

\paragraph{Diffusion Mixture for Graph Generation \label{sec:method:mixture}}
As the graph mixture in Eq.~\eqref{eq:predicted_graph} is a weighted mean of the final graphs, conceptually, this can be modeled by aggregating the endpoint-conditioned processes with respect to the weights from the graph mixture. Inspired by the diffusion mixture framework~\citep{peluchetti21mixture,wu22bridge,liu22bridge}, we design the generation process by mixing the OU bridge processes with the endpoints from the data distribution, where we leverage the \emph{diffusion mixture representation}~\citep{brigo08mixture,peluchetti21mixture}. This yields the SDE representation of a mixture process as a weighted mean of the SDEs of the diffusion processes (we provide a formal definition of the mixture representation in Appendix~\ref{sec:app:derivation:representation}).

Specifically, we mix a collection of OU bridge processes $\{ \mathbb{Q}^{\bm{g}}: \bm{g}\!=\!(\bm{x},\! \bm{a})\!\sim\!\Pi^{\ast} \}$ to construct a generation process, for which the \emph{mixture process} is modeled by the SDE:
\begin{align}
\mathbb{Q}^{\Pi^{\ast}\!\!}:
\begin{cases}
    \mathrm{d}\bm{X}_t = \eta_1(\bm{X}_t,\bm{A}_t,t)\mathrm{d}t + \sigma_{1,t}\mathrm{d}\mathbf{W}_{\!1,t} \\[2pt]
    \mathrm{d}\bm{A}_t \,= \eta_2(\bm{X}_t,\bm{A}_t,t)\mathrm{d}t + \sigma_{2,t}\mathrm{d}\mathbf{W}_{\!2,t}
\end{cases}
\label{eq:mixture}
\end{align}
with $\bm{G}_0\!=\!(\bm{X}_0,\!\bm{A}_0)$ following an arbitrary prior distribution $\Gamma$ and the drifts $\eta_1$ and $\eta_2$ defined as follows:
\begin{align}
    \begin{pmatrix}
         \eta_1(\bm{X}_t,\!\bm{A}_t,t) \\ \eta_2(\bm{X}_t,\!\bm{A}_t,t)
     \end{pmatrix} = \int\! \begin{pmatrix}
         \eta^{\bm{x}}_1(\bm{X}_t,t) \\ \eta^{\bm{a}}_2(\bm{A}_t,t)
     \end{pmatrix} \frac{p^{\bm{g}}_t(\bm{G}_t)}{p_t(\bm{G}_t)}\Pi^{\ast}(\mathrm{d}\bm{g})
\end{align}
for $\bm{G}_t\!=\!(\bm{X}_t,\!\bm{A}_t)$, where $p^{\bm{g}}_t$ is the marginal density of the bridge process $\mathbb{Q}^{\bm{g}}$, and $p_t(\cdot) \!\coloneqq\! \int p^{\bm{g}}_t(\cdot)\Pi^{\ast}(\mathrm{d}\bm{g})$ is the marginal density of the mixture process. Notably, the terminal distribution of the mixture process $\mathbb{Q}^{\Pi^{\ast}}\!$ is equal to the data distribution $\Pi^{\ast}\!$ by construction. We refer to this mixture process as the \emph{OU bridge mixture}.

Remarkably, the mixture process $\mathbb{Q}^{\Pi^{\ast}}\!$ can be explicitly represented in terms of the graph mixture. We derive a parameterization of $\mathbb{Q}^{\Pi^{\ast}}\!$ from the SDE representation of the OU bridge process in Eq.~\eqref{eq:ou_bridge} as follows (see Appendix~\ref{sec:app:derivation:mixture} for the derivation):
\begin{sizeddisplay}{\small}
\begin{equation}
\begin{split}
    & \eta_1(\bm{X}_t,\!\bm{A}_t,\!t) = \alpha_{1}\sigma_{1,t}^2 \bm{X}_t \!+\! \frac{\sigma_{1,t}^2}{v_{1,t}} \! \left(\frac{\bm{D}_{X}(\bm{X}_t,\!\bm{A}_t,\!t)}{u_{1,t}} \!-\! \bm{X}_t \right) \\
    & \eta_2(\bm{X}_t,\!\bm{A}_t,\!t) = \alpha_{2}\sigma_{2,t}^2 \bm{A}_t \!+\! \frac{\sigma_{2,t}^2}{v_{2,t}} \! \left(\frac{\bm{D}_{A}(\bm{X}_t,\!\bm{A}_t,\!t)}{u_{2,t}} \!-\! \bm{A}_t \right) \\
\end{split}
\label{eq:drift_mixture}
\end{equation}
\end{sizeddisplay}
where $\bm{D}_X$ and $\bm{D}_{A}$ are defined as a weighted mean of the node features and the adjacency matrices respectively:
\begin{equation}
\hspace{-2mm}
\begin{split}
    \bm{D}^{\Pi^{\ast}}\!\!(\bm{G}_t,t) \!\coloneqq \!
    \begin{pmatrix}
        \bm{D}_{X}\!(\bm{G}_t,t) \\ \bm{D}_{A}(\bm{G}_t,t)
    \end{pmatrix} \!=\!\!
    \int \!\! \begin{pmatrix}
        \bm{x} \\ \bm{a}
    \end{pmatrix} \!
    \frac{p^{\bm{g}}_t(\bm{G}_t)}{p_t(\bm{G}_t)}\Pi^{\ast}\!(\mathrm{d}\bm{g})
\end{split}
\label{eq:graph_mixture}
\hspace{-1mm}
\end{equation}
Notice that from the definition of the transition distribution, we can derive the following identity:
\begin{align*}
    \bm{D}^{\Pi^{\ast}}\!(\bm{G}_t,t) = \!\int \!\bm{g}\frac{p^{\bm{g}}_t(\bm{G}_t)}{p_t(\bm{G}_t)} \Pi^{\ast}(\mathrm{d}\bm{g})
    = \!\int\! \bm{g}\cdot p_{T|t}(\bm{g}|\bm{G}_t) \mathrm{d}\bm{g} ,
\end{align*}
which shows that $\bm{D}^{\Pi^{\ast}}\!(\cdot,t)$ coincides with the graph mixture of $\mathbb{Q}^{\Pi^{\ast}}\!$ as in Eq.~\eqref{eq:predicted_graph}. 
As a result, $\bm{D}^{\Pi^{\ast}}\!(\cdot,t)$ acts as the prediction of the final graph at time $t$, where $\bm{D}_X$ and $\bm{D}_A$ are the predicted node features and adjacency matrices, respectively, in the form of a weighted mean of data. 

In the view of the graph mixture as the weighted mean from $\bm{D}^{\Pi^{\ast}}$, it converges to the final graph of the mixture process since the marginal density $p^{\bm{g}}_t$ of the bridge process converges to one if $\bm{g}$ corresponds to the final graph while the probability becomes zero otherwise. 
This convergence is achieved at an early stage as visualized in Figure \ref{fig:concept} (c) and Appendix~\ref{sec:app:vis:process}, where we further analyze the convergence behavior with respect to the coefficient $\alpha$ and the noise schedule $\sigma_t$ in Appendix~\ref{sec:app:additional:coeff}. 

A key observation is that the drift of the OU bridge mixture in Eq.~\eqref{eq:drift_mixture} highly resembles the drift of the OU bridge process in Eq.~\eqref{eq:ou_bridge}, except that the final graph $\bm{g}$ is replaced by the graph mixture.
From this observation, we can see that the trajectory of the mixture process is guided by the drift in the direction of $\bm{D}^{\Pi^{\ast}}\!(\cdot,t)$, driven toward the graph mixture that converges to a graph in the data distribution $\Pi^{\ast}$.
Therefore, if we could estimate the graph mixture of this process, we can build a generative model upon the mixture process without relying on score function or noise, where the graph structures and the topological attributes can be explicitly modeled by the graph mixture.

Before introducing the training objective for learning the graph mixture, we discuss the difference between our framework and the denoising diffusion models. Our generative process is modeled by the mixture of bridge processes that describes the exact transport from the prior distribution to the data distribution by construction, whereas the time reversal of denoising diffusion models is not an exact transport to the data distribution for finite time~\citep{franzese2023much}. We provide further discussion on the difference in the characteristics of our mixture process and the denoising diffusion processes in Appendix~\ref{sec:app:derivation:comparison}.

\subsection{Generation Framework Using Graph Mixture\label{sec:method:learn_mixture}}

\paragraph{Training Objectives\label{sec:method:objective}}
Our goal is to design a generative model that explicitly learns the graph topology. To this end, we leverage the OU bridge mixture parameterized by the graph mixture, where we estimate the graph mixture using a neural network $\bm{s}_{\theta}(\cdot,t)$ that corresponds to directly learning the graph structures.
In particular, we show that estimating the graph mixture guarantees to maximize the likelihood of our generative model. For the rest of the section, we represent the system of SDEs of Eq.~\eqref{eq:mixture} as a SDE with respect to $\bm{G}_t$ for notational simplicity.

We propose to define the generative model $\mathbb{P}^{\theta}$ to approximate the mixture process $\mathbb{Q}^{\Pi^{\ast}}\!$ as follows:
\begin{equation}
\begin{aligned}
    &\mathbb{P}^{\theta}: \mathrm{d}\bm{G}_t = \eta_{\theta}(\bm{G}_t,t)\mathrm{d}t + \sigma_t\mathrm{d}\mathbf{W}_t , \\
    &\eta_{\theta}(\bm{G}_t,t) = \alpha\sigma_t^2\bm{G}_t + \frac{\sigma_t^2}{v_t}\left(\frac{1}{u_t}\bm{s}_{\theta}(\bm{G}_t,t) - \bm{G}_t\right) ,
\end{aligned} \label{eq:generative_model}
\end{equation}
where $\bm{s}_{\theta}$ is desired to estimate the graph mixture $\bm{D}^{\Pi^{\ast}}$. 

In order to model the drift $\eta_{\theta}$, we provide a tractable objective for estimating the graph mixture, which guarantees to maximize the likelihood of our generative model $\mathbb{P}^{\theta}$. Leveraging the Girsanov theorem~\citep{oksendal03sde}, we upper-bound the KL divergence between $\Pi^{\ast}\!$ and the terminal distribution of $\mathbb{P}^{\theta}$ denoted as $p^{\theta}_T$ as follows (see Appendix~\ref{sec:app:derivation:objective} for a detailed derivation of the objective): 
\begin{align}
    &D_{KL}(\Pi^{\ast}\|p^{\theta}_T) \leq D_{KL}(\mathbb{Q}^{\Pi^{\ast}} \|\mathbb{P}^{\theta}) \notag \\
    &= \mathbb{E}_{\bm{G}\sim\mathbb{Q}^{\Pi^{\ast}}}\! \bigg[ \frac{1}{2}\int^T_0 \!\! \gamma_t^2 \left\| \bm{s}_{\theta}(\bm{G}_t,t) - \bm{D}^{\Pi^{\ast}}\!\!(\bm{G}_t,t) \right\|^2 \mathrm{d}t  \bigg] + C_1 \notag \\
    &= \mathbb{E}_{\bm{G}\sim\mathbb{Q}^{\Pi^{\ast}}}\! \bigg[ \frac{1}{2}\int^T_0 \!\! \gamma_t^2 \left\| \bm{s}_{\theta}(\bm{G}_t,t) - \bm{G}_T \right\|^2 \! \mathrm{d}t  \bigg] + C_2 , 
    \label{eq:objective}
\end{align}
where $\gamma_t\coloneqq\sigma_t/(u_tv_t)$, $C_1$ and $C_2$ are constants independent of $\theta$, and the expectation is computed over the samples $\bm{G}$ from the OU bridge mixture $\mathbb{Q}^{\Pi^{\ast}}\!$.

During training, $\bm{G}_t$ from the mixture process $\mathbb{Q}^{\Pi^{\ast}}\!$ can be easily obtained without simulating the SDE.
Notice that $\bm{G}_t$ follows the distribution $p_{t|0,T}(\bm{G}_t|\bm{G}_0,\bm{G}_T)$, which is the distribution of the mixture process $\mathbb{Q}^{\Pi^{\ast}}\!$ at time $t$ given the endpoints $\bm{G}_0$ and $\bm{G}_T$.
By construction, the OU bridge mixture with fixed endpoints $\bm{G}_0$ and $\bm{G}_T$ coincides with the OU process (Eq.~\eqref{eq:ou_process}) with these fixed endpoints, and $p_{t|0,T}(\bm{G}_t|\bm{G}_0,\bm{G}_T)$ corresponds to the marginal probability of the OU process with the fixed endpoints $\bm{G}_0$ and $\bm{G}_T$. 

Using the Bayes theorem, we derive that the distribution $p(\bm{G}_t|\bm{G}_0,\bm{G}_T)$ is also Gaussian that results from the product of Gaussian distributions, where the mean $\mu^{\ast}_t$ and the covariance $\mathbf{\Sigma}^{\ast}_t$ have analytical forms as follows (see Appendix~\ref{sec:app:derivation:prob} for the derivation):
\begin{equation}
\begin{aligned}
    \mu^{\ast}_t &= \frac{\sinh\left( \varphi_{T}-\varphi_{t} \right)}{\sinh\left( \varphi_{T} \right)} \bm{G}_0 
    + \frac{\sinh\left( \varphi_{t} \right)}{\sinh\left( \varphi_{T} \right)} \bm{G}_T ,\\
    \mathbf{\Sigma}^{\ast}_t &=  \frac{1}{\alpha}\frac{\sinh\left( \varphi_{T}-\varphi_{t} \right) \sinh\left( \varphi_{t} \right)}{ \sinh\left( \varphi_{T} \right)} \mathbf{I} ,        
\end{aligned}
\label{eq:transition_mean_variance_main}
\end{equation}
where $\varphi_t \coloneqq \alpha\int^{t}_{0}\sigma_{\tau}^2\mathrm{d}\tau $.
Thereby the training of GruM is simulation-free, and our approach is 17.5 times faster compared to the training that relies on expensive SDE simulation~\citep{wu22bridge}.

In particular, Eq.~\eqref{eq:transition_mean_variance_main} shed light on the connection of the OU bridge mixture and the stochastic interpolant~\citep{albergo23interpolant} between the distributions $\Gamma$ and $\Pi^{\ast}$ as follows:
\begin{align}
    \bm{G}_t =& \frac{\sinh\left( \varphi_{T}-\varphi_{t} \right)}{\sinh\left( \varphi_{T} \right)} \bm{G}_0 
    + \frac{\sinh\left( \varphi_{t} \right)}{\sinh\left( \varphi_{T} \right)} \bm{G}_T \notag \\
    &+ \left( \frac{1}{\alpha}\frac{\sinh\left( \varphi_{T}-\varphi_{t} \right) \sinh\left( \varphi_{t} \right)}{ \sinh\left( \varphi_{T} \right)} \right)^{1/2} \bm{Z} .
    \label{eq:interpolant_main}
\end{align}
where $\bm{G}_0\!\sim\!\Gamma$, $\bm{G}_T\!\sim\!\Pi^{\ast}\!$, and $\bm{Z}\!\sim\!\mathcal{N}(\bm{0},\mathbf{I})$, respectively.

\begin{figure}[t!]
\centering
\begin{algorithm}[H]
    \caption{ Training}\label{alg:training_main}
        \textbf{Input:} Model $\bm{s}_{\theta}$, constant $\epsilon$ \\
        \textbf{For each epoch:} \phantom{a}
    \begin{algorithmic}[1]
        \STATE Sample graph $\bm{G}$ from the training set
        \STATE $N \leftarrow$ \text{number of nodes of} $\bm{G}$
        \STATE Sample $t\sim[0,T-\epsilon]$ and $\bm{G}_0\sim\mathcal{N}(0,\mathbf{I}_N)$
        \STATE Sample $\bm{G}_t\sim p_{t|0,T}(\bm{G}_t|\bm{G}_0,\bm{G})$ 
        \COMMENT{Eq.~\eqref{eq:interpolant_main}}
        \STATE $\gamma_t \leftarrow {\sigma_{t}}/{u_tv_t}$
        \STATE $\mathcal{L}_{\theta} \leftarrow \gamma_t^2 \| \bm{s}_{\theta}(\bm{G}_t,t) - \bm{G} \|^2$ 
        \STATE Update $\theta$ using $\mathcal{L}_{\theta}$
    \end{algorithmic}
\end{algorithm}
\vspace{-0.275in}
\begin{algorithm}[H]
    \caption{Sampling}\label{alg:sampling_main}
        \textbf{Input:} Trained model $\bm{s}_{\theta}$, number of sampling steps $K$, diffusion step size $\mathrm{d}t$
    \begin{algorithmic}[1]
        \STATE Sample number of nodes $N$ from the training set.
        \STATE $\bm{G}_0\sim \mathcal{N}(0,\mathbf{I}_N)$
        \COMMENT{Start from noise}
        \STATE $t\leftarrow 0$
        \FOR{$k=1$ \textbf{to} $K$}
            \STATE $\eta\leftarrow \alpha\sigma_t^2\bm{G}_t + \frac{\sigma_t^2}{v_t} \left(\frac{1}{u_t}\bm{s}_{\theta}(\bm{G}_t,t) - \bm{G}_t\right)$ 
            \STATE $\mathbf{w}\sim\mathcal{N}(0,\mathbf{I}_N)$ 
            \STATE $\bm{G}_{t+\mathrm{d}t} \!\leftarrow \eta\mathrm{d}t +  \sigma_t \sqrt{\mathrm{d} t} \mathbf{w}$
            \COMMENT{Euler-Maruyama Step}
            \STATE $t \leftarrow t+\mathrm{d}t$
        \ENDFOR
        \STATE $\bm{G} \leftarrow \texttt{quantize}(\bm{G}_t)$
        \COMMENT{Quantize if necessary}
        \STATE \textbf{Return:} Graph $\bm{G}$
    \end{algorithmic}
\end{algorithm}
\vspace{-0.2in}
\end{figure}

Note that the goal of Eq.~\eqref{eq:objective} is to model the drift of the OU bridge mixture parametrized by the graph mixture, where $\bm{s}_{\theta}$ is trained to estimate the graph mixture instead of the exact graph $\bm{G}_T$, and we refer to this objective as the \emph{graph mixture matching}. 
Learning the graph mixture not only allows us to directly model the structures of the final graph and their topological properties, but further guarantees our generative model to closely approximate the data distribution. 
Additionally, we discuss the difference between learning the graph mixture and the training objectives of denoising diffusion models in Appendix~\ref{sec:app:derivation:probability_flow} and \ref{sec:app:derivation:comparison}.
We summarize the training process in Algorithm~\ref{alg:training_main} and provide the details in Appendix~\ref{sec:app:details:objective}.

\paragraph{Sampling from GruM}
Using the trained model $\bm{s}_{\theta}$ to compute the drift $\eta_{\theta}$ of the parameterized mixture process in Eq.~\eqref{eq:generative_model}, we generate samples by simulating $\mathbb{P}^{\theta}$ from time $t=0$ to $t=T$ with initial samples drawn from the prior distribution. 
Note that we generate the node features and the adjacency matrices simultaneously using the system of SDEs in the form of Eq.~\eqref{eq:mixture}, and solving the SDEs is similar to that of denoising diffusion models which does not require additional time.
We summarize the sampling process in Algorithm~\ref{alg:sampling_main} and describe the details in Appendix~\ref{sec:app:details:sampler}.

\paragraph{Advantages of Our Framework \label{sec:method:advantages}} 
We conclude this section by explaining the advantages of our framework. First, GruM can directly model the graph topology by predicting the graph structures via the graph mixture, instead of implicitly capturing via noise or score. 
Furthermore, our framework is not restricted to the type of data to be generated, allowing it to be applicable to both continuous and discrete data, for example, 3D molecules with both discrete atom types and continuous coordinates.

From the perspective of the model hypothesis space, learning the graph mixture is considerably easier compared to previous objectives such as learning the score function or the drift of the diffusion process.
While the graph mixture is supported inside the bounded data space, the score function or the drift tends to diverge near the terminal time which could be problematic for the model to learn. Furthermore, we can exploit the inductive bias of the graph data for learning the graph mixture, which is critical as it dramatically reduces the hypothesis space. To be specific, we can leverage the prior knowledge of the graph representation such as one-hot encoding or the categorical type by adding an additional function at the last layer of the model $\bm{s}_{\theta}$, for instance, softmax function for the one-hot encoded node features and the sigmoid function for the 0-1 adjacency matrices (we provide more details in Appendix~\ref{sec:app:details:model}). We experimentally verify these advantages in Section~\ref{sec:exp:analysis}.

\begin{table*}[t]
    \caption{\textbf{Generation results on the general graph datasets.} Best results are highlighted in bold, where smaller MMD and larger V.U.N. indicate better results. Hyphen(-) denotes out-of-resources that take more than 2 weeks.}
    \label{tab:general_graph}
\vspace{-0.075in}
    \centering
    \resizebox{\textwidth}{!}{
    \renewcommand{\arraystretch}{1.1}
    \renewcommand{\tabcolsep}{4pt}
    \begin{tabular}{l c c c c a c c c c a c c c c}
    \toprule
        &
        \multicolumn{5}{c}{Planar} &
        \multicolumn{5}{c}{SBM} &
        \multicolumn{4}{c}{Proteins} \\
    \cmidrule(l{2pt}r{2pt}){2-6}    
    \cmidrule(l{2pt}r{2pt}){7-11}
    \cmidrule(l{2pt}r{2pt}){12-15}
        &
        \multicolumn{5}{c}{Synthetic, $|V| = 64$} &
        \multicolumn{5}{c}{Synthetic, $44\leq|V|\leq187$} &
        \multicolumn{4}{c}{Real, $100\leq|V|\leq500$} \\
    \cmidrule(l{2pt}r{2pt}){2-6}    
    \cmidrule(l{2pt}r{2pt}){7-11}
    \cmidrule(l{2pt}r{2pt}){12-15}
        & Deg. & Clus. & Orbit & Spec. & V.U.N.  & Deg. & Clus. & Orbit & Spec. & V.U.N.  & Deg. & Clus. & Orbit & Spec. \\
    \midrule
        Training set & 0.0002 & 0.0310 & 0.0005 & 0.0052 & 100.0 & 0.0008 & 0.0332 & 0.0255 & 0.0063 & 100.0 & 0.0003 & 0.0068 & 0.0032 & 0.0009 \\
    \midrule
        GraphRNN & 0.0049 & 0.2779 & 1.2543 & 0.0459 & 0.0 & 0.0055 & 0.0584 & 0.0785 & 0.0065 & 5.0 & 0.0040 & 0.1475 & 0.5851 & 0.0152  \\
        GRAN & 0.0007 & 0.0426 & 0.0009 & 0.0075 & 0.0 & 0.0113 & 0.0553 & 0.0540 & 0.0054 & 25.0 & 0.0479 & 0.1234 & 0.3458 & 0.0125 \\
        SPECTRE & 0.0005 & 0.0785 & 0.0012 & 0.0112 & 25.0 & 0.0015 & 0.0521 & \textbf{0.0412} & 0.0056 & 52.5 & 0.0056 & 0.0843 & \textbf{0.0267} & 0.0052 \\
    \midrule
        EDP-GNN & 0.0044 & 0.3187 & 1.4986 & 0.0813 & 0.0 & 0.0011 & 0.0552 & 0.0520 & 0.0070 & 35.0 & - & - & - & - \\
        GDSS & 0.0041 & 0.2676 & 0.1720 & 0.0370 & 0.0 & 0.0212 & 0.0646 & 0.0894 & 0.0128 & 5.0 & 0.0861 & 0.5111 & 0.732 & 0.0748 \\
        ConGress & 0.0048 & 0.2728 & 1.2950 & 0.0418 & 0.0 & 0.0273 & 0.1029 & 0.1148 & - & 0.0 & -  & - & - & - \\
        DiGress & \textbf{0.0003} & 0.0372 & \textbf{0.0009} & 0.0106 & 75 & 0.0013 & 0.0498 & 0.0434 & 0.0400 & 74 & - & - & - & - \\
    \midrule
        \textbf{GruM (Ours)} & 0.0005 & \textbf{0.0353} & \textbf{0.0009} & \textbf{0.0062} & \textbf{90.0} & \textbf{0.0007} & \textbf{0.0492} & 0.0448 & \textbf{0.0050} & \textbf{85.0} & \textbf{0.0019} & \textbf{0.0660} & 0.0345 & \textbf{0.0030} \\
    \bottomrule
    \end{tabular}}
\end{table*}
\begin{figure*}[t]
\vspace{-0.105in}
    \centering
    \includegraphics[width=0.95\linewidth]{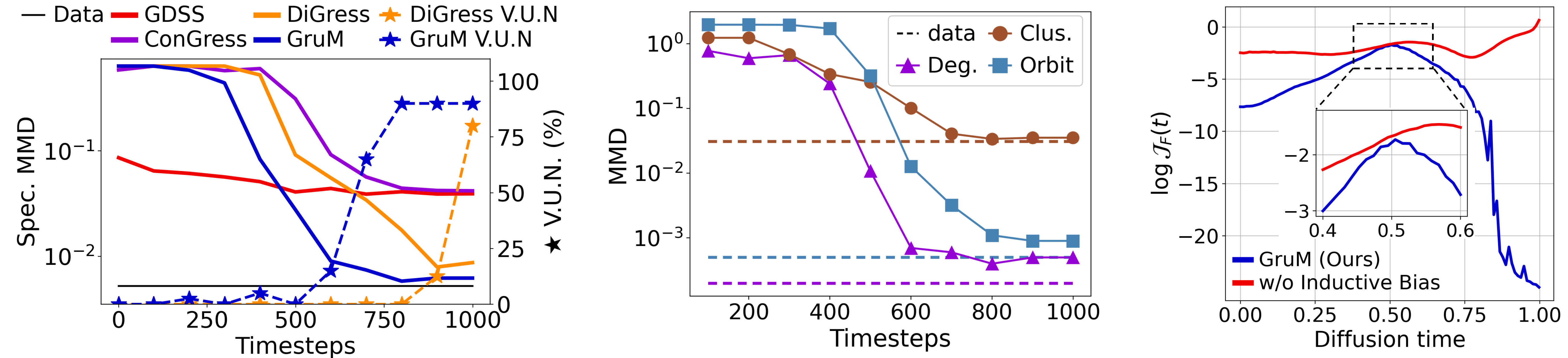}
\vspace{-0.105in}
\caption{\textbf{(Left) Topology analysis}. We compare Spec. MMD and V.U.N of the graph mixture from GruM against the implicit prediction computed from GDSS, ConGress, and DiGress which we provide details in Appendix~\ref{sec:app:exp:general_implementation}. \textbf{(Middle) MMD between the test set and the graph mixture of GruM} through the generative process. \textbf{(Right) The complexity of GruM} with and without using the inductive bias, measured by the Frobenius norm of the Jacobian of the models.}
\vspace{-0.1in}
\label{fig:analysis}    
\end{figure*}


\section{Experiments~\label{sec:experiments}}

\subsection{General Graph Generation~\label{subsec:exp:general}}
\vspace{-0.05in}
We validate GruM on general graph generation tasks to show that it can generate valid graph topology.

\vspace{-0.05in}
\paragraph{Datasets and Metrics}
We evaluate the quality of generated graphs on three synthetic and real datasets used as benchmarks in previous works~\citep{martinkus22spectre, vignac22digress}: \textbf{Planar}, Stochastic Block Model (\textbf{SBM}), and \textbf{Proteins}~\citep{protein}. We follow the evaluation setting of \citet{martinkus22spectre} using the same data split. We measure the maximum mean discrepancy (MMD) of four graph statistics between the set of generated graphs and the test set: degree (\textbf{Deg.}), clustering coefficient (\textbf{Clus.}), count of orbits with 4 nodes (\textbf{Orbit}), and the eigenvalues of the graph Laplacian (\textbf{Spec.}). To verify that the model truly learns the distribution, we report the percentage of valid, unique, and novel (\textbf{V.U.N.}) graphs for which the validness is defined as satisfying the specific property of each dataset. We provide further details in Appendix~\ref{sec:app:exp:general}.

\vspace{-0.05in}
\paragraph{Baselines}
We compare our method against the following graph generative models: \textbf{GraphRNN}~\citep{you2018graphrnn} an autoregressive model based on RNN, \textbf{GRAN}~\citep{liao2019gran} an autoregressive model with attention, \textbf{SPECTRE}~\citep{martinkus22spectre} a one-shot model based on GAN, \textbf{EDP-GNN}~\citep{niu20edpgnn} a score-based model for adjacency matrix, \textbf{GDSS}~\citep{jo22gdss} and \textbf{ConGress}~\citep{vignac22digress} a continuous diffusion model, and \textbf{DiGress}~\citep{vignac22digress}, a discrete diffusion model. We provide the details of training and sampling of our GruM in Appendix~\ref{sec:app:details} and describe further implementation details including the hyperparameters in Appendix~\ref{sec:app:exp:general}.

\vspace{-0.05in}
\paragraph{Results}
Table~\ref{tab:general_graph} shows that our method outperforms all the baselines on all datasets. Especially, ours achieves the highest validity (V.U.N.) metric, as it accurately learns the underlying topology of the graphs. Notably, our method outperforms DiGress by a large margin in V.U.N., even though we do not use specific prior distributions or structural feature augmentation that are utilized in DiGress.  
We provide an ablation study on the model architecture in Appendix~\ref{sec:app:additional:arch} to validate that the superior performance of GruM comes from its ability to accurately model the graph topology by predicting the graph mixture.
We provide the visualization of the generated graphs and the generative process of GruM in Appendix~\ref{sec:app:vis}, showing that it can accurately capture the attributes of each dataset.

\begin{table*}[t]
    \caption{\textbf{Generation results on the 2D molecule datasets.} We report the mean of 3 different runs. Best results are highlighted in bold. We provide the results of uniqueness, novelty and variance in Appendix \ref{sec:app:additional_2d_mol}.} \label{tab:2d_mol}
    \vspace{-0.075in}
    \centering
    \resizebox{\textwidth}{!}{
    \renewcommand{\arraystretch}{0.95}
    \renewcommand{\tabcolsep}{8pt}
    \begin{tabular}{l c c c c c c c c}
    \toprule
    & \multicolumn{4}{c}{\phantom{\scriptsize{($|V|\leq 9$)}}QM9 ~\scriptsize{($|V|\leq 9$)}} & \multicolumn{4}{c}{\phantom{\scriptsize{($|V|\leq 38$)}}ZINC250k \scriptsize{($|V|\leq 38$)}} \\
    \cmidrule(l{2pt}r{2pt}){2-5}
    \cmidrule(l{2pt}r{2pt}){6-9}
        Method & Valid (\%)$\uparrow$ & FCD$\downarrow$ & NSPDK$\downarrow$ & Scaf.$\uparrow$ & Valid (\%)$\uparrow$ & FCD$\downarrow$ & NSPDK$\downarrow$ &  Scaf.$\uparrow$ \\
    \midrule
    Training set & 100.0 & 0.0398 & 0.0001 & 0.9719 & 100.0 & 0.0615 & 0.0001 & 0.8395 \\
    \midrule
        MoFlow~\citep{zang2020moflow} & 91.36 & 4.467 & 0.0169 & 0.1447 & 63.11 & 20.931 & 0.0455 & 0.0133 \\
        GraphAF~\citep{shi2020graphaf} & 74.43 & 5.625 & 0.0207 & 0.3046 & 68.47 & 16.023 & 0.0442 & 0.0672 \\
        GraphDF~\citep{luo2021graphdf} & 93.88 & 10.928 & 0.0636 & 0.0978 & 90.61 & 33.546 & 0.1770 & 0.0000 \\
    \midrule
        EDP-GNN~\citep{niu20edpgnn} & 47.52 & 2.680 & 0.0046 & 0.3270 & 82.97 & 16.737 & 0.0485 & 0.0000 \\
        GDSS~\citep{jo22gdss} & 95.72 & 2.900 & 0.0033 & 0.6983 & 97.01 & 14.656 & 0.0195 & 0.0467 \\
        DiGress~\citep{vignac22digress} & 98.19 & \textbf{0.095} & 0.0003 & 0.9353 & 94.99 & \phantom{0}3.482 & 0.0021 & 0.4163 \\
    \midrule
        \textbf{GruM (Ours)} & \textbf{99.69} & 0.108 & \textbf{0.0002} & \textbf{0.9449} & \textbf{98.65} & \phantom{0}\textbf{2.257} & \textbf{0.0015} & \textbf{0.5299} \\
    \bottomrule
    \end{tabular}}
\end{table*}
\begin{figure*}[t]
\centering
\vspace{-0.2in}
\begin{minipage}{0.69\linewidth}
    \resizebox{\textwidth}{!}{
    \renewcommand{\arraystretch}{1.0}
    \renewcommand{\tabcolsep}{4pt}
    \begin{tabular}{l c c c c}
    \toprule
    & \multicolumn{2}{c}{\phantom{\scriptsize{($|V|\leq 29$)}}QM9 ~\scriptsize{($|V|\leq 29$})} & \multicolumn{2}{c}{GEOM-DRUGS ~\scriptsize{($|V|\leq 181$)}} \\
    \cmidrule(l{2pt}r{2pt}){2-3}
    \cmidrule(l{2pt}r{2pt}){4-5}
        Method & Atom Stab.(\%) & Mol. Stab.(\%) & Atom Stab.(\%) & Mol. Stab.(\%) \\
    \midrule
    G-Schnet~\citep{gebauer19gschnet} & 95.7 & 68.1 & - & - \\
    EN-Flow~\citep{satorras21enflow} & 85.0 & \phantom{0}4.9  & 75.0 & 0.0 \\
    GDM~\citep{hoogeboom22edm} & 97.0 & 63.2 & 75.0 & 0.0 \\
    EDM~\citep{hoogeboom22edm} & 98.7~\small{$\pm$0.1} & 82.0~\small{$\pm$0.4} & 81.3 & 0.0  \\
    Bridge~\citep{wu22bridge} & 98.7~\small{$\pm$0.1} & 81.8~\small{$\pm$0.2} & 81.0~\small{$\pm$0.7} & 0.0 \\
    Bridge+Force~\citep{wu22bridge} & 98.8~\small{$\pm$0.1} & 84.6~\small{$\pm$0.2} & 82.4~\small{$\pm$0.7} & 0.0 \\
    \midrule
    \textbf{GruM (Ours)} & \textbf{98.81}~\small{$\pm$0.03} & \textbf{87.34}~\small{$\pm$0.19} & \textbf{82.96}~\small{$\pm$0.12} & \textbf{0.51}~\small{$\pm$0.03} \\
    \bottomrule
    \end{tabular}}
\end{minipage}
\hfill
\begin{minipage}{0.30\linewidth}
    \vspace{0.05in}
    \includegraphics[width=1\linewidth]{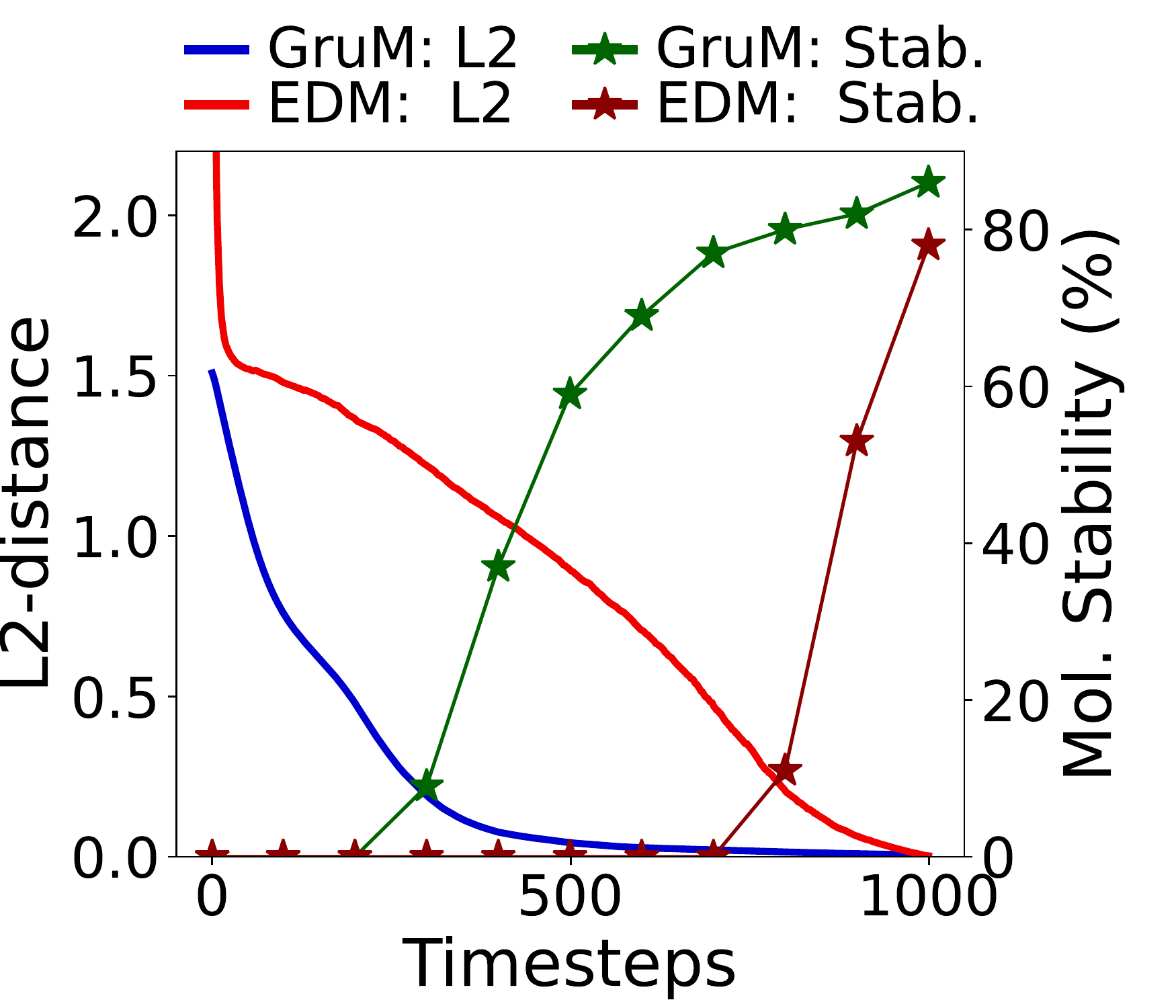}
\end{minipage}
\vspace{-0.13in}
\caption{\textbf{(Left) Generation results on the 3D molecule datasets.} Best results are highlighted in bold which is the average of 3 different runs. The baseline results are taken from \citet{hoogeboom22edm} and \citet{wu22bridge}. \textbf{(Right) Convergence of the generative process.} We compare the convergence of the graph mixture from GruM and the implicit prediction computed from EDM. We measure the convergence (L2 distance) and report the molecule stability of the predictions. 
} 
\label{tab:3d_mol}
\vspace{-0.15in}
\end{figure*}

\vspace{-0.05in}
\paragraph{Topology Analysis}
To show how learning the graph mixture results in graphs with correct topology, we conduct an analysis of the graph mixture. Figure~\ref{fig:analysis} (Left) demonstrates that GruM can achieve the spectral property of the final graph at an early stage by explicitly modeling the topology via learning the graph mixture. In contrast, GDSS and ConGress fail to recover the spectral properties as they implicitly model the topology via predicting the noise or score functions. Further, ours recovers the spectral property faster than DiGress, resulting in graphs with higher validity. In particular, we observe that the V.U.N. of the estimated graph mixture increases after achieving the desired spectral property, resulting in 90\% V.U.N. This shows that predicting the final graph allows us to better capture the global topologies. Moreover, we plot the MMD results of GruM through the generative process in Figure~\ref{fig:analysis} (Middle), which demonstrates that the local characteristics of the predicted graph rapidly converge to that of the graphs from the training set.

\subsection{2D Molecule Generation~\label{subsec:exp:2d_mol}}
\vspace{-0.05in}
We further validate GruM on 2D molecule generation tasks to show that it can accurately generate graphs with both the node features and the topologies of the target graphs.

\vspace{-0.05in}
\paragraph{Datasets and Metrics}
We evaluate the quality of generated 2D molecules on two molecule datasets used as benchmarks in \citet{jo22gdss}: \textbf{QM9}~\citep{ramakrishnan2014qm9} and \textbf{ZINC250k}~\citep{irwin2012zinc}. 
Following the evaluation setting of \citet{jo22gdss}, we evaluate the models with four metrics: \textbf{Validity} is the percentage of the valid molecules among the generated without any posthoc correction. \textbf{FCD}~\citep{preuer2018frechet} measures the distance between the sets of molecules in the chemical space. \textbf{NSPDK MMD}~\citep{nspdk} evaluates the quality of the graph structure compared to the test set. \textbf{Scaffold similarity} (Scaf.) evaluates the ability to generate similar substructures. We provide more details in Appendix~\ref{sec:app:exp:2d_mol}.

\vspace{-0.05in}
\paragraph{Baselines}
We compare to the following molecular graph generative models: \textbf{MoFlow}~\citep{zang2020moflow} is a one-shot flow-based model. \textbf{GraphAF}~\citep{shi2020graphaf} and \textbf{GraphDF}~\citep{luo2021graphdf} are autoregressive flow-based model. \textbf{EDP-GNN}, \textbf{GDSS}, \textbf{ConGress}, and \textbf{DiGress} are diffusion models previously explained. We describe further implementation details in Appendix~\ref{sec:app:exp:2d_mol}.

\vspace{-0.05in}
\paragraph{Results}
Table~\ref{tab:2d_mol} shows that our method achieves the highest validity on all datasets verifying that GruM can generate valid molecules without correction. Further, ours outperforms the baselines in FCD and NSPDK metrics demonstrating that the molecules synthesized by GruM are similar to the molecule from the training set in both chemical and graph-structure aspects. Especially, ours achieves the highest scaffold similarity indicating that it is able to generate similar substructures from that of the training set. We visualize the generated molecules in Appendix~\ref{sec:app:vis:samples}.

\subsection{3D Molecule Generation~\label{subsec:exp:3d_mol}}
\vspace{-0.05in}
To show that GruM is able to generate graphs with both continuous and discrete features, we validate it on 3D molecule generation tasks, which come with discrete atom types and continuous coordinates.

\vspace{-0.05in}
\paragraph{Datasets and Metrics}
We evaluate the generated 3D molecules on two standard molecule datasets used as benchmarks in \citet{hoogeboom22edm}: \textbf{QM9}~\citep{ramakrishnan2014qm9} (up to 29 atoms) and \textbf{GEOM-DRUGS}~\citep{axelrod22geom-drugs} (up to 181 atoms). Following \citet{hoogeboom22edm}, both datasets include hydrogen atoms. For GEOM-DRUGS, we select 30 conformations for each molecule with the lowest energy. We evaluate the quality of the generated molecules with two stability metrics: \textbf{Atom stability} is the percentage of the atoms with valid valency. \textbf{Molecule stability} is the percentage of the generated molecules that consist of stable atoms. We provide more details in Appendix~\ref{sec:app:exp:3d_mol}.

\vspace{-0.05in}
\paragraph{Baselines}
We compare GruM against 3D molecule generative models: \textbf{G-Schnet}~\citep{gebauer19gschnet} is an autoregressive model based on the 3d point sets. \textbf{EN-Flow}~\citep{satorras21enflow} is a flow-based model. \textbf{GDM} and \textbf{EDM}~\citep{hoogeboom22edm} are denoising diffusion models. \textbf{Bridge}~\citep{wu22bridge} is a diffusion model based on the diffusion mixture that learns to approximate the drift and \textbf{Bridge+Force}~\citep{wu22bridge} adds physical force to the drift. For ours, we follow the training setting of \citet{hoogeboom22edm} using the same architecture of EGNN~\citep{satorras21enflow}. We describe further implementation details in Appendix~\ref{sec:app:exp:3d_mol}.

\vspace{-0.05in}
\paragraph{Results}
As shown in the table of Figure~\ref{tab:3d_mol}, our method yields the highest atom stability compared to all the baselines on both datasets. Furthermore, ours achieves higher molecule stability since we directly model the topology by learning the graph mixture. Moreover, GruM outperforms Bridge+Force~\citep{wu22bridge} even though GruM does not require task-dependent prior force while trained in a simulation-free manner. Notably, our method achieves non-zero molecule stability in the GEOM-DRUGS dataset consisting of large molecules with up to 181 atoms. We visualize the generated molecules and the generative process of GruM in Appendix~\ref{sec:app:vis}, demonstrating that we can predict the final molecule at an early stage of the process leading to stable molecules. We further observe that GruM generates $\times$1.5 more number of connected molecules compared to EDM as shown in Table~\ref{tab:geom_conn} of the Appendix.

\vspace{-0.05in}
\paragraph{Stability Analysis}
To further investigate the superior performance of our framework in generating more stable molecules, we conduct an analysis of the convergence and stability.
Figure~\ref{tab:3d_mol} (Right) shows the convergence of the predicted graph from GruM and the implicit prediction from EDM computed from the estimated noise. We observe that for GruM, the predicted graphs converge rapidly to the final result. After the convergence, the stability of GruM increases as it has sufficient steps to calibrate the details to produce valid molecules, which is visualized in the generative process of Figure~\ref{fig:3d_process} of the Appendix. As for EDM, the implicit predictions converge slowly since EDM does not explicitly learn the information of the final result, which leads to lower stability. This analysis shows that learning the final graph is significantly superior in capturing the correct topology compared to previous diffusion models.

\subsection{Further Analysis~\label{sec:exp:analysis}}
\vspace{-0.05in}
We conduct an analysis to investigate the advantages of our framework explained in Section~\ref{sec:method:advantages}.

\vspace{-0.05in}
\paragraph{Exploiting Inductive Bias}
To validate that exploiting the inductive bias of the graph data is critical, we compare GruM against a variant of it without an additional function at the last layer in the model. Figure~\ref{fig:analysis} (Right) shows the complexity of the models $\bm{s}_{\theta}$ trained on the Planar dataset, where the transformation at the last layer significantly reduces the model complexity for predicting the final graph. Especially, the larger complexity gap at the late stage of the diffusion process suggests that exploiting the inductive bias is crucial for learning valid structures and their topology.

\vspace{-0.05in}
\paragraph{Comparison with Learning Drift}
To verify that learning the graph mixture as in our framework is superior to learning the drift, we compare with Bridge~\citep{wu22bridge} which models the drift of the mixture process. Table~\ref{tab:3d_mol} shows that ours outperforms Bridge, especially for the molecule stability, since learning the drift is challenging due to its diverging nature and unable to model the topology directly. We further validate that learning the drift performs poorly on general graph generation tasks and fails to generate the correct topology in Appendix~\ref{sec:app:additional:drift}.

\vspace{-0.05in}
\paragraph{Early stopping for the generative process}
In Figure~\ref{fig:analysis} (Left) and (Middle), the V.U.N. and the MMD results of DruM in the Planar dataset demonstrate that the estimated destination mixture converges to the exact destination at early sampling steps, accurately capturing both the global topology and local graph characteristics. This allows us to early-stop the diffusion process, which reduces the generation time by up to 20\% on this task. The generation results on SBM and Proteins datasets in Section~\ref{sec:app:additional:early_stopping} of the Appendix show a similar tendency. 
\section{Conclusion}
In this work, we proposed a new diffusion-based graph generation framework, GruM, that explicitly models the topology of the graphs. Unlike existing graph diffusion models that learn to denoise, our framework learns to predict the final graph of the generative process through the graph mixture, thereby accurately capturing the valid graph structure and its topological features. 
Specifically, we construct the generation process as a mixture of diffusion bridges, which differs from the denoising diffusion process, where the drift drives the generation process toward the predicted graph that converges in an early stage. 
We extensively validated our framework on diverse graph generation tasks, including 2D/3D molecular generation, on which ours significantly outperforms previous graph generation methods. 
A promising future direction would be the generalization to domains other than graphs where the topology of the data is important, such as proteins and manifolds.

\paragraph{Impact Statement}
This paper presents work whose goal is to advance the field of deep generative models, specifically for graph-structured data. We believe that our work can accelerate the discovery of feasible drugs and improve the quality of human lives by recommending drug candidates in silico, which reduces time-consuming wet lab experiments performed by experts. However, one might maliciously use our framework to generate toxic substances or narcotics harmful to humans or the environment. 

\paragraph{Acknowledgement}
This work was supported by Institute for Information \& communications Technology Promotion(IITP) grant funded by the Korea government(MSIT) (No.2019-0-00075 Artificial Intelligence Graduate School Program(KAIST)), Google Research Grant and Google Cloud Research Credits program with the award (XKCV-N0JU-8K3R-65LK, 40KR-GJX5-XH4X-PU3L), Institute of Information \& communications Technology Planning \& Evaluation (IITP) grant funded by the Korea government(MSIT) (No.2022-0-00713), and the National Research Foundation of Korea(NRF) grant funded by the Korea government(MSIT) (No. RS-2023-00256259).

\bibliography{references}
\bibliographystyle{icml2024}

\newpage
\newpage
\appendix
\onecolumn
\begin{center}{\bf {\LARGE Appendix}}\end{center}

\paragraph{Organization} The Appendix is organized as follows: In Section~\ref{sec:app:derivation}, we provide the derivations of the results from the main paper. In Section~\ref{sec:app:details}, we explain the details of our generative framework including the training objectives, the sampling method, and the model architectures. In Section~\ref{sec:app:exp}, we provide experimental details for the generation tasks and further present additional experimental results in Section~\ref{sec:app:additional}. In Section~\ref{sec:app:vis}, we visualize the generated graphs and molecules, with visualized generative processes. Finally, in Section~\ref{sec:app:limitation}, we discuss the limitations of our work.

\section{Derivations \label{sec:app:derivation}}

\subsection{Diffusion bridge processes \label{sec:app:derivation:bridge}}
Here we derive the Ornstein-Uhlenbeck (OU) bridge process using Doob's h-transform~\citep{doob1984h-transform} and show that the Brownian bridge process is a special case of the OU bridge process. We further discuss a general class of bridge processes and explain the advantage of the OU bridge process.

\paragraph{Ornstein-Uhlenbeck bridge process}
First, we consider the simple case when the reference process is given as a standard OU process without a time-dependent diffusion coefficient:
\begin{equation}
    \hat{\mathbb{Q}} \;:\; \mathrm{d}\bm{G}_t = \alpha\bm{G}_t\mathrm{d}t + \mathrm{d}\mathbf{W}_t ,
    \label{eq:standard_ou_process}
\end{equation}
where $\alpha$ is a constant. Then the Doob's h-transform on $\hat{\mathbb{Q}}$ yields the representation of an endpoint-conditioned process $\hat{\mathbb{Q}}^{\bm{g}} \coloneqq \hat{\mathbb{Q}}(\cdot|\bm{G}_T=\bm{g})$ defined by the following SDE:
\begin{align}
     \hat{\mathbb{Q}}^{\bm{g}} \;:\; \mathrm{d}\bm{G}_t = \Big[ \alpha\bm{G}_t + \nabla_{\!\bm{G}_t}\! \log \hat{p}_{T|t}(\bm{g}|\bm{G}_t) \Big]\mathrm{d}t + \mathrm{d}\mathbf{W}_t ,
\end{align}
where $\hat{p}_{T|t}(\bm{g}|\bm{G}_t)$ is the transition probability from time $t$ to $T$ of the standard OU process in Eq.~\eqref{eq:standard_ou_process}. Since the standard OU process has a linear drift, the transition probability is Gaussian, i.e. $\hat{p}_{T|t}(\bm{g}|\bm{G}_t)=\mathcal{N}(\bm{g}; \mu_t, \mathbf{\Sigma}_{t})$, where the mean $\mu_{t}$ and the covariance $\mathbf{\Sigma}_{t}$ satisfies the following ODEs (derived from the results of Eq.(5.50) and Eq.(5.51) of \citet{kernel_derivation}):
\begin{align}
    \frac{\mathrm{d}\mu_t}{\mathrm{d}t} = \alpha\mu_t \;\;,\;\; 
    \frac{\mathrm{d}\mathbf{\Sigma}_{t}}{\mathrm{d}t} = \mathbf{I} + 2\alpha\mathbf{\Sigma}_{t} .
    \label{eq:transition_ode}
\end{align}
The ODE with respect to $\mathbf{\Sigma}_{t}$ can be modified as:
\begin{align}
    \frac{\mathrm{d}}{\mathrm{d}t}e^{-2\alpha t}\mathbf{\Sigma}_{t} = e^{-2\alpha t}\mathbf{I} ,
\end{align}
which give the following closed-form solutions:
\begin{align}
    \mu_{t} = \hat{u}_t\bm{G}_t \;\;,\;\; \mathbf{\Sigma}_{t} = \frac{1}{2\alpha}\left( \hat{u}_t^2 - 1 \right) \mathbf{I} \;\;\text{for} \;\; \hat{u}_t = e^{\alpha(T\!\!-\!t)}.
\end{align}
Therefore, the SDE representation of the standard OU bridge process with fixed endpoint $\bm{g}$ is given as follows:
\begin{align}
    \hat{\mathbb{Q}}^{\bm{g}} \;:\; \mathrm{d}\bm{G}_t = \left[ \alpha\bm{G}_t + \frac{2\alpha}{1-\hat{u}_t^{-2}} \Big( \frac{\bm{g}}{\hat{u}_t}- \bm{G}_t \Big) \right]\mathrm{d}t + \mathrm{d}\mathbf{W}_t .
\end{align}
Now we derive the bridge process for the general OU process with a time-dependent diffusion coefficient defined by the following SDE: 
\begin{equation}
    \mathbb{Q}\!:\; \mathrm{d}\bm{G}_t = \alpha\sigma_t^2\bm{G}_t\mathrm{d}t + \sigma_t\mathrm{d}\mathbf{W}_t ,
    \label{eq:general_ou_process}
\end{equation}
where $\sigma_t$ is a scalar function. Since the time change (Section 8.5. of \citet{oksendal03sde}) with $\beta_t = \int^t_0\sigma_{\tau}^2\mathrm{d}\tau$ of $\hat{\mathbb{Q}}$ in Eq.~\eqref{eq:standard_ou_process} is equivalent to $\mathbb{Q}$ of Eq.~\eqref{eq:general_ou_process}, the transition probability $\tilde{p}_{T|t}(\bm{g}|\bm{G}_t)$ of the general OU process satisfies the following:
\begin{align}
    \tilde{p}_{T|t}(\bm{g}|\bm{G}_t) = \hat{p}_{\beta_T|\beta_t}(\bm{g}|\bm{G}_t)
\end{align}
Thereby, the OU bridge process conditioned on the endpoint $\bm{g}$ is defined by the following SDE:
\begin{align}
    \mathbb{Q}^{\bm{g}} \;:\; \mathrm{d}\bm{G}_t = \left[ \alpha\sigma_t^2\bm{G}_t + \frac{\sigma_t^2}{v_t}\Big( \frac{\bm{g}}{u_t} - \bm{G}_t \Big) \right]\mathrm{d}t + \sigma_t\mathrm{d}\mathbf{W}_t , 
    \label{eq:ou_bridge_app}
\end{align}
where the scalar function $u_t$ and $v_t$ are given as:
\begin{align}
    u_t = e^{\alpha(\beta_T-\beta_t)} = \exp\Big( \alpha\int^T_t \sigma^2_{\tau} \mathrm{d}\tau \Big) \;\;,\;\; v_t = \frac{1}{2\alpha}(1-u_t^{-2}) .
\end{align}
Note that the OU bridge process, also known as the constrained OU process, was studied theoretically in previous works~\citep{corlay13oubridge,peluchetti21mixture,debortoli21bridge}. However, we are the first to validate the effectiveness of the OU bridge processes for modeling the generative process through extensive experiments, especially for the generation of graphs in diverse tasks including the generation of general graphs as well as 2D and 3D molecular graphs.

\paragraph{Brownian bridge process}
We show that the Brownian bridge process is a special case of the OU bridge process. When the constant $\alpha$ of the OU bridge process approaches $0$, the scalar function $u_t$ converges to 1 that leads to the convergence of $v_t$ as follows: 
\begin{align*}
    v_t = \frac{1}{2\alpha}(1-u_t^{-2}) = \frac{1}{2\alpha}\left( 1 - e^{-2\alpha(\beta_T-\beta_t)} \right) \rightarrow \beta_T-\beta_t ,
\end{align*} 
which is due to the Taylor expansion of the exponential function. Therefore, the OU bridge process for $\alpha\rightarrow0$ is modeled by the following SDE:
\begin{align}
    \mathbb{Q}^{\bm{g}}_{bb} \;:\; \mathrm{d}\bm{G}_t =\! \frac{\sigma_t^2}{\beta_T - \beta_t} \!\left( \bm{g} - \bm{G}_t \right)\mathrm{d}t + \sigma_t\mathrm{d}\mathbf{W}_t ,
    \label{eq:brownian_bridge}
\end{align}
which is equivalent to the SDE representation of the Brownian bridge process.
Compared to the OU bridge process in Eq.~\eqref{eq:ou_bridge_app}, the Brownian bridge process has a simpler SDE representation with less flexibility for designing the generative process as the process is solely determined by the noise schedule $\sigma_t$.

Note that the Brownian bridge is an endpoint-conditioned process with respect to a reference Brownian Motion defined by the following SDE:
\begin{align}
    \mathrm{d}\bm{G}_t = \sigma_t\mathrm{d}\mathbf{W}_t ,
\end{align}
which is a diffusion process without drift, and also a special case of the OU process that is used for the reference process of the OU bridge process.

\paragraph{More bridge processes}
\citet{wu22bridge} proposes an approach for designing a more general class of diffusion bridges using the Lyapunov function method. Starting from a simple Brownian bridge $\mathbb{Q}^{\bm{g}}_{bb}$, we can create a new bridge process by adding an extra drift term as follows: 
\begin{align}
    \mathbb{Q}^{\bm{g}}_{bb,f} \;:\; &\mathrm{d}\bm{G}_t =\! \bigg[ \underbrace{\sigma_tf_t(\bm{G}_t)}_{\text{extra drift}} + \frac{\sigma_t^2}{\beta_T - \beta_t} \!\left( \bm{g} - \bm{G}_t \right) \bigg]\mathrm{d}t + \sigma_t\mathrm{d}\mathbf{W}_t , \label{eq:more_general_bridge} \\
    &\text{for} \;\; f_t \;\; \text{satisfying} \quad \mathbb{E}_{\bm{G}\sim\mathbb{Q}^{\bm{g}}_{bb,f}}[\|f_t(\bm{G}_t)\|^2]<\infty .
    \label{eq:function_condition}
\end{align}
$\mathbb{Q}^{\bm{g}}_{bb,f}$ of Eq.~\eqref{eq:more_general_bridge} is still a bridge process with endpoint $\bm{g}$ since the drift of the Brownian bridge (i.e. Eq.~\eqref{eq:brownian_bridge}) dominates the extra drift term due to the condition of Eq.~\eqref{eq:function_condition}. Moreover, \citet{wu22bridge} introduces problem-dependent prior $f$ inspired by physical energy functions.

These general bridge processes could be used for our framework to construct a mixture process for modeling the generative process, as described in Section~\ref{sec:method:mixture}. If the explicit SDE representation for the general bridges is accessible, the mixture process can be represented by leveraging the diffusion mixture representation, and further the Brownian bridge could be replaced with the OU bridge process. 

However, in contrast to constructing the generative process as a mixture of the OU bridge processes, using the mixture of the general bridge processes results in difficulty during training; Training a generative model that approximates the mixture of the general bridge processes requires expensive SDE simulation due to the intractable transition probability. We show through extensive experiments that for our approach, the family of OU bridge processes is sufficient to model the complex generation process while the generative model can be trained in a simulation-free manner.

\subsection{Diffusion mixture representation \label{sec:app:derivation:representation}}
In this section, we provide the formal definition of the diffusion mixture representation~\citep{brigo08mixture,peluchetti21mixture}. 

Consider a collection of diffusion processes $\{ \mathbb{Q}^{\lambda} : \lambda\in\Lambda \}$ defined by the SDEs:
\begin{align}
    \mathbb{Q}^{\lambda} : \mathrm{d}\bm{Z}^{\lambda}_t = \eta^{\lambda}(\bm{Z}_t,t)\mathrm{d}t + \sigma_t^{\lambda}\mathrm{d}\mathbf{W}^{\lambda}_t \;\;,\;\; \bm{Z}^{\lambda}_0\sim p^{\lambda}_0 
\end{align}
where $\mathbf{W}^{\lambda}_t$ are independent standard Wiener processes and $p^{\lambda}_0$ are the initial distributions. Denote $p^{\lambda}_t$ as the marginal density of the process $\mathbb{Q}^{\lambda}$. Further, define the mixture of marginal densities and the mixture of initial distributions with respect to a mixing distribution $\mathcal{L}$ on the collection $\Lambda$ as follows: 
\begin{align}
    p_t(z) = \int_{\Lambda}p^{\lambda}_t(z) \mathcal{L}(\mathrm{d}\lambda) \;\;,\;\; p_0(z) = \int_{\Lambda}p^{\lambda}_0(z) \mathcal{L}(\mathrm{d}\lambda) ,
    \label{eq:mixture_marginals}
\end{align}
Then there exists a diffusion process that induces a marginal density $p_t$, and the diffusion process is modeled by the following SDE:
\begin{equation}
    \mathbb{Q}^{\mathcal{L}} : \mathrm{d}\bm{Z}_t = \eta(\bm{Z}_t,t)\mathrm{d}t + \sigma_t\mathrm{d}\mathbf{W}_t \;\;,\;\; \bm{Z}_0\sim p_0,
    \label{eq:mixture_sde}
\end{equation}
where the drift and diffusion coefficients are given as the weighted mean of the corresponding coefficients of $\mathbb{Q}^{\lambda}$ as follows:
\begin{align}
    \eta(z,t) = \int_{\Lambda} \eta^{\lambda}(z,t)\frac{p^{\lambda}_t(z)}{p_t(z)}\mathcal{L}(\mathrm{d}\lambda) \;\;, \;\; 
    \sigma_t^2 = \int_{\Lambda} (\sigma^{\lambda}_t)^2\frac{p^{\lambda}_t(z)}{p_t(z)}\mathcal{L}(\mathrm{d}\lambda).
    \label{eq:mixture_drift_sigma}
\end{align}
Below, we provide a proof of this statement.

\paragraph{proof.}
It is enough to show that $p_t$ defined in Eq.~\eqref{eq:mixture_marginals} is the solution to the corresponding Fokker-Planck equation of Eq.~\eqref{eq:mixture_sde}, which is given as follows:
\begin{align}
    \frac{\partial q_t(z)}{\partial t} = -\nabla_{\!z}\cdot \left( q_t(z)\eta(z,t) - \frac{1}{2}\sigma_t^2\nabla_{\!z}q_t(z) \right),
    \label{eq:fokker_planck}
\end{align}
where $q_t$ denotes the marginal density of Eq.~\eqref{eq:mixture_sde}. Using the definition of Eq.~\eqref{eq:mixture_marginals} and the corresponding Fokker-Planck equations with respect to $\mathbb{Q}^{\lambda}$ for $\lambda\in\Lambda$, we derive the following result:
\begin{align}
    \frac{\partial p_t(z)}{\partial t} &= \frac{\partial}{\partial t} \int_{\Lambda} p^{\lambda}_t(z) \mathcal{L}(\mathrm{d}\lambda) = \int_{\Lambda} \frac{\partial}{\partial t} p^{\lambda}_t(z) \mathcal{L}(\mathrm{d}\lambda) \notag \\
    &= \int_{\Lambda} \left[ -\nabla_{\!z}\cdot \left( \eta^{\lambda}(z,t) p^{\lambda}_t(z) - \frac{1}{2}(\sigma^{\lambda}_t)^2\nabla_{\!z}p^{\lambda}_t(z) \right) \right] \mathcal{L}(\mathrm{d}\lambda) \notag \\
    &= -\nabla_{\!z}\cdot \int_{\Lambda} \left[ \eta^{\lambda}(z,t) p^{\lambda}_t(z) - \frac{1}{2}(\sigma^{\lambda}_t)^2\nabla_{\!z}p^{\lambda}_t(z) \right] \mathcal{L}(\mathrm{d}\lambda) \notag \\
    &= -\nabla_{\!z}\cdot \left( p_t(z) \int_{\Lambda} \eta^{\lambda}(z,t)\frac{p^{\lambda}_t(z)}{p_t(z)}\mathcal{L}(\mathrm{d}\lambda) - \frac{1}{2} \nabla_{\!z} \left[ p_t(z) \int_{\Lambda} (\sigma^{\lambda}_t)^2\frac{p^{\lambda}_t(z)}{p_t(z)}\mathcal{L}(\mathrm{d}\lambda) \right] \right) \notag \\
    &= -\nabla_{\!z}\cdot \left(  p_t(z)\eta(z,t) - \frac{1}{2}\sigma_t^2 \nabla_{\!z}p_t(z)\right) ,
\end{align}
which proves that $p_t$ is the solution to the Fokker-Planck equation of Eq.~\eqref{eq:fokker_planck}.

\subsection{OU bridge mixture \label{sec:app:derivation:mixture}}
Now we use the diffusion mixture representation described in Appendix~\ref{sec:app:derivation:representation} to derive the OU bridge mixture.
Consider a mixture of the collection of OU bridge processes with endpoints in the data distribution, i.e. $\{ \mathbb{Q}^{\bm{g}}:\bm{g}\sim\Pi^{\ast} \}$. 
We mix this collection of processes with the data distribution $\Pi^{\ast}$ as the mixing distribution, which is represented by the following SDE:
\begin{align}
    \mathbb{Q}^{\Pi^{\ast}}: \mathrm{d}\bm{G}_t &= \left[ \int \left( \alpha\sigma_t^2\bm{G}_t + \frac{\sigma_t^2}{v_t}\Big( \frac{\bm{g}}{u_t} - \bm{G}_t \Big) \right) \frac{p^{\bm{g}}_t(\bm{G}_t)}{p_t(\bm{G}_t)} \Pi^{\ast}(\mathrm{d}\bm{g}) \right]  \mathrm{d}t + \sigma_t\mathrm{d}\mathbf{W}_t \notag \\[5pt]
    &= \left[ \alpha\sigma_t^2\bm{G}_t + \frac{\sigma_t^2}{v_t} \left( \frac{1}{u_t}\int \bm{g} \frac{p^{\bm{g}}_t(\bm{G}_t)}{p_t(\bm{G}_t)} \Pi^{\ast}(\mathrm{d}\bm{g}) - \bm{G}_t \right) \right] \mathrm{d}t + \sigma_t\mathrm{d}\mathbf{W}_t \notag \\[5pt]
    &= \left[ \alpha\sigma_t^2\bm{G}_t + \frac{\sigma_t^2}{v_t} \left( \frac{1}{u_t} \bm{D}^{\Pi^{\ast}\!}(\bm{G}_t,t) - \bm{G}_t \right) \right] \mathrm{d}t + \sigma_t\mathrm{d}\mathbf{W}_t
    \label{eq:ou_bridge_mixture}
\end{align}
where $p_t(z) = \int p^{\bm{g}}_t(z)\Pi^{\ast}(\mathrm{d}\bm{g})$ is used for the second equality and the definition of the graph mixture (Eq.~\eqref{eq:graph_mixture}) is used for the last equality.

\subsection{Reverse-time diffusion process of the OU bridge mixture \label{sec:app:derivation:reverse}}
Here we derive the reverse-time diffusion process of GruM, i.e. the time reversal of the OU bridge mixture. Since the generative process of GruM transports the prior distribution $\Gamma$ to the data distribution $\Pi^{\ast}\!$, the time reversal of GruM transports $\Pi^{\ast}\!$ to $\Gamma$. We show that it has a similar SDE representation as Eq.~\eqref{eq:ou_bridge_mixture}.

We derive the reverse process of the OU bridge mixture by constructing a mixture of the reverse processes of each OU bridge process.
To be precise, for the mixture process $\mathbb{Q}\coloneqq \int \mathbb{Q}^{\bm{g}}\mathrm{d}\Pi^{\ast}$, the reverse process of $\mathbb{Q}$ denoted as $\overbar{\mathbb{Q}}$ is equal to the mixture process $\int \overbar{\mathbb{Q}}^{\bm{x}}\mathrm{d}\Gamma$ where $\overbar{\mathbb{Q}}^{\bm{x}}$ is the reverse process of the bridge process $\mathbb{Q}^{\bm{g}}$ with starting point $\bm{x}$.
For the simplicity of the representation, we first derive the time-reversal of general bridge processes, where the reference process is given as
\begin{align}
    \mathrm{d}\bm{G}^{ref}_t = \mu(\bm{G}^{ref}_t,t) + \sigma_t\mathbf{W}_t ,
    \label{eq:general_reference}
\end{align}
with the marginal density denoted as $\tilde{p}_t$.
In order to obtain the reverse-time diffusion process, we leverage the reverse-time SDE representation~\citep{anderson82reversesde,song21sde} as follows:
\begin{align}
    \mathrm{d}\overbar{\bm{G}}^{ref}_t = \Big[- \mu(\overbar{\bm{G}}^{ref}_t,T\!\!-\!t) + \sigma_{T\!\!-\!t}^2\nabla_{\overbar{\bm{G}}^{ref}_t}\log \tilde{q}_t(\overbar{\bm{G}}^{ref}) \Big]\mathrm{d}t + \sigma_{T\!\!-\!t}\mathrm{d}\mathbf{W}_t,
    \label{eq:reverse_reference}
\end{align}
where $\tilde{q}_t=\tilde{p}_{T\!\!-\!t}$ is the marginal density of the process $\{\overbar{\bm{G}}^{ref}_t\}_{t\in[0,T]}$. 
Then the bridge process of Eq.~\eqref{eq:reverse_reference} with fixed end point $\bm{x}\sim\Gamma$ can be derived by using the Doob's h-transform~\citep{doob1984h-transform} as follows:
\begin{align}
    \overbar{\mathbb{Q}}^{\bm{x}} \!: \mathrm{d}\overbar{\bm{G}}_t = \!\Big[\! -\! \mu(\overbar{\bm{G}}_t,T\!\!-\!t) 
    +\! \sigma_{T\!\!-\!t}^2\nabla_{\overbar{\bm{G}}_t}\!\log \tilde{q}_t(\overbar{\bm{G}}_t) 
    +\! \sigma_{T\!\!-\!t}^2\nabla_{\overbar{\bm{G}}_t}\! \log \tilde{q}_{T|t}(\bm{x}|\overbar{\bm{G}}_t)
    \Big]\!\mathrm{d}t +\! \sigma_{T\!\!-\!t}\mathrm{d}\mathbf{W}_t ,
    \label{eq:reverse_bridge_before}
\end{align}
which is a reverse process for the conditioned process of $\bm{G}^{ref}$ with starting point $\bm{x}$ and endpoint $\bm{g}\sim\Pi^{\ast}\!$ fixed.
Here using the fact that $\tilde{q}_t=\tilde{p}_{T\!\!-\!t}$, we can see that 
\begin{align}
    \tilde{q}_t(y) \tilde{q}_{T|t}(x|y) = \tilde{q}(\overbar{\bm{G}}_T\!=\!x, \overbar{\bm{G}}_t\!=\!y) = \frac{\tilde{q}(\overbar{\bm{G}}_T\!=\!x, \overbar{\bm{G}}_t\!=\!y)}{\tilde{q}_T(x)}\tilde{q}_T(x) = \tilde{p}_{T\!\!-\!t|0}(y|x)\tilde{q}_T(x),
\end{align}
and since $\nabla_{\overbar{\bm{G}}_t} \log\tilde{q}_T(\bm{x})=0$ for fixed $\bm{x}$, Eq.~\eqref{eq:reverse_bridge_before} can be simplified as follows:
\begin{align}
    \overbar{\mathbb{Q}}^{\bm{x}} : \mathrm{d}\overbar{\bm{G}}_t = \Big[ -\mu(\overbar{\bm{G}}_t,T\!\!-\!t) 
    + \sigma_{T\!\!-\!t}^2 \nabla_{\overbar{\bm{G}}_t}\log \tilde{p}_{T\!\!-\!t|0}(\overbar{\bm{G}}_t|\bm{x})
    \Big]\mathrm{d}t + \sigma_{T\!\!-\!t}\mathrm{d}\mathbf{W}_t .
    \label{eq:reverse_bridge}
\end{align}
Finally, the mixture of the bridge processes $\{ \overbar{\mathbb{Q}}^{\bm{x}} : \bm{x}\!\sim\!\Gamma \}$ can be derived using the diffusion mixture representation as follows:
\begin{align}
    \overbar{\mathbb{Q}} : \mathrm{d}\overbar{\bm{G}}_t = \left[ -\mu(\overbar{\bm{G}}_t,t) + \sigma_{T\!\!-\!t}^2 \int \nabla_{\overbar{\bm{G}}_t}\log \tilde{p}_{T\!\!-\!t|0}(\overbar{\bm{G}}_t|\bm{x}) \frac{q^{\bm{x}}_t(\overbar{\bm{G}}_t)}{q_t(\overbar{\bm{G}}_t)}\Gamma(\mathrm{d}\bm{x}) \right]\mathrm{d}t + \sigma_{T\!\!-\!t}\mathrm{d}\mathbf{W}_t ,
    \label{eq:reverse_mixture}
\end{align}
where $q^{\bm{x}}_t$ is the marginal density of $\overbar{\mathbb{Q}}^{\bm{x}}$ and $q_t$ is the marginal density of the mixture process $\overbar{\mathbb{Q}}$ defined as $q_t(\cdot) \coloneqq \int q^{\bm{x}}_t(\cdot)\Gamma(\mathrm{d}\bm{x})$.

Using the result of Eq.~\eqref{eq:reverse_mixture}, now we can derive the time reversal of the OU bridge mixture by setting $\mu(z,t)=\alpha\sigma_t^2z$. 
Since the transition distributions of the OU process satisfy the following (we provide closed-form mean and covariance of the transition distribution in Eq.~\eqref{eq:transition_distribution}):
\begin{align}
    \tilde{p}_{T\!-\!t|0}(\bm{z}|\bm{x}) = \mathcal{N}\left(\bm{z}; \; \overbar{u}_t \bm{x},\; \overbar{u}_t^2\overbar{v}_t \mathbf{I} \right) \;\text{ for }\;\;
    \overbar{u}_t \!= \exp\! \left(\!\alpha\int^{T\!\!-\!t}_0\sigma_{\tau}^2\mathrm{d}\tau \!\right), 
    \; \overbar{v}_t \!= \frac{1}{2\alpha}\left(1 - \overbar{u}_t^{-2}\right) ,
\end{align}
the log gradient of the transition distribution can be computed as follows:
\begin{align}
    \nabla_{\bm{z}}\log \tilde{p}_{T\!-\!t|0}(\bm{z} | \bm{x}) 
    = -\frac{1}{\overbar{u}_t^2\overbar{v}_t}\left( \bm{z} - \overbar{u}_t \bm{x} \right).
\end{align}
Thereby, the reverse-time diffusion process of the OU bridge mixture is given by:
\begin{align}
    \overbar{\mathbb{Q}} :
    \mathrm{d}\overbar{\bm{G}}_t = \left[ -\alpha\sigma_{T\!\!-\!t}^2\overbar{\bm{G}}_t + \frac{\sigma_{T\!\!-\!t}^2}{\overbar{u}_t^2\overbar{v}_t}\Big( \overbar{u}_t\bm{D}^{\Gamma}(\overbar{\bm{G}}_t,t) - \overbar{\bm{G}}_t \Big) \right]\mathrm{d}t + \sigma_{T\!\!-\!t}\mathrm{d}\mathbf{W}_t, \quad \overbar{\bm{G}}_0\sim\Pi^{\ast},
    \label{eq:reverse_ou_mixture}
\end{align}
where $\bm{D}^{\Gamma}(\cdot,t)$ is the graph mixture of $\overbar{\mathbb{Q}}$ defined as follows:
\begin{align}
    \bm{D}^{\Gamma}(\overbar{\bm{G}}_t,t) = \int \bm{x}\frac{q^{\bm{x}}_t(\overbar{\bm{G}}_t)}{q_t(\overbar{\bm{G}}_t)} \Gamma(\mathrm{d}\bm{x}).
\end{align}
Since $\overbar{\mathbb{Q}}$ describes the diffusion process from the data distribution to the prior distribution, it can be considered a perturbation process. Further, we can observe that the time reversal of the OU bridge mixture is non-linear with respect to $\overbar{\bm{G}}_t$ in general, and completely different from the forward process (i.e. perturbation process) of denoising diffusion models, i.e. the VESDE or VPSDE~\citep{song21sde}. 

Note that the reverse process of the OU bridge mixture perfectly transports the data distribution $\Pi^{\ast}\!$ to the arbitrary prior distribution $\Gamma$ in the sense that the terminal distribution exactly matches $\Gamma$ for finite terminal time $T$.
On the other hand, the forward process of denoising diffusion models, for example, VPSDE~\citep{song21sde}, does not perfectly transport the data distribution to the prior distribution. The terminal distribution of the forward process is approximately Gaussian but not exactly a Gaussian distribution for finite $T$, although the mismatch is small for sufficiently large $T$. This is because the forward process requires infinite $T$ in order to decouple the prior distribution $\Gamma$ from the data distribution $\Pi^{\ast}$.

In conclusion, the generative process of GruM is different from denoising diffusion models which naturally follows from the fact that the time reversal of the OU bridge mixture is different from the forward processes of denoising diffusion models.

\subsection{Derivation of the graph mixture matching objective \label{sec:app:derivation:objective}}
We provide the derivation of our graph mixture matching objective, corresponding to Eq.~\eqref{eq:objective}. First, we leverage the Girsanov theorem~\cite{oksendal03sde} to upper bound the KL divergence between the data distribution $\Pi^{\ast}\!$ and the terminal distribution of $\mathbb{P}^{\theta}$ denoted as $p^{\theta}_T$: 
\begin{align}
    D_{KL}(\Pi^{\ast}\|p^{\theta}_T) 
    &\leq D_{KL}(\mathbb{Q}^{\Pi^{\ast}} \|\mathbb{P}^{\theta})  \\ 
    &= D_{KL}(\mathbb{Q}^{\Pi^{\ast}}_0\|\mathbb{P}^{\theta}_0) + \mathbb{E}_{\mathbb{Q}^{\Pi^{\ast}}}\! \bigg[ \log\frac{\mathrm{d}\mathbb{Q}^{\Pi^{\ast}}\!}{\mathrm{d}\mathbb{P}^{\theta}} \bigg]  \\
    &= \mathbb{E}_{\bm{G}\sim\mathbb{Q}^{\Pi^{\ast}}}\! \bigg[\! -\log p^{\theta}_0(\bm{G}_0) + \frac{1}{2}\int^T_0 \!\!\left\| \sigma_t^{-1} \left(\eta_{\theta}(\bm{G}_t,t) - \eta(\bm{G}_t,t)\right) \right\|^2 \mathrm{d}t \bigg] + C \\
    &= \mathbb{E}_{\bm{G}\sim\mathbb{Q}^{\Pi^{\ast}}}\! \bigg[\! -\log p^{\theta}_0(\bm{G}_0) + \frac{1}{2}\int^T_0 \!\! \gamma_t^2 \left\| \bm{s}_{\theta}(\bm{G}_t,t) - \bm{D}^{\Pi^{\ast}}\!\!(\bm{G}_t,t) \right\|^2 \mathrm{d}t  \bigg] + C , 
    \label{eq:objective_derivation}
\end{align} 
where $p^{\theta}_0$ is a predetermined prior distribution that is easy to sample from, for instance, Gaussian distribution, and $C$ is a constant independent of $\theta$. Note that the first inequality is known as the data processing inequality. The expectation in Eq.~\eqref{eq:objective_derivation} corresponds to Eq.~\eqref{eq:objective}.

Furthermore, the expectation of Eq.~\eqref{eq:objective_derivation} can be written as follows:
\begin{align}
    &\mathbb{E}_{\bm{G}\sim\mathbb{Q}^{\Pi^{\ast}}}\! \bigg[ \frac{1}{2}\int^T_0 \!\! \gamma_t^2 \left\| \bm{s}_{\theta}(\bm{G}_t,t) - \bm{D}^{\Pi^{\ast}}\!\!(\bm{G}_t,t) \right\|^2 \mathrm{d}t  \bigg] \notag \\
    &= \mathbb{E}_{\bm{G}\sim\mathbb{Q}^{\Pi^{\ast}}}\! \bigg[ \frac{1}{2}\int^T_0 \!\! \gamma_t^2 \left\| \Big( \bm{s}_{\theta}(\bm{G}_t,t) - \bm{G}_T \Big) + \Big( \bm{G}_T - \bm{D}^{\Pi^{\ast}}\!\!(\bm{G}_t,t)  \Big) \right\|^2 \mathrm{d}t  \bigg] \notag \\
    &= \mathbb{E}_{\bm{G}\sim\mathbb{Q}^{\Pi^{\ast}}}\! \bigg[ \frac{1}{2}\int^T_0 \!\! \gamma_t^2 \left\| \bm{s}_{\theta}(\bm{G}_t,t) - \bm{G}_T \right\|^2 \! \mathrm{d}t  \bigg] 
    + \mathcal{E} + \mathcal{E}^T + C_1,
\end{align}
where $\mathcal{E}$ and $C_1$ are defined as:
\begin{equation}
\begin{split}
    \mathcal{E} &= \mathbb{E}_{\bm{G}\sim\mathbb{Q}^{\Pi^{\ast}}}\! \bigg[ \frac{1}{2}\int^T_0 \!\! \gamma_t^2 \,  \Big( \bm{s}_{\theta}(\bm{G}_t,t) - \bm{G}_T \Big)^T \Big( \bm{G}_T - \bm{D}^{\Pi^{\ast}}\!\!(\bm{G}_t,t) \Big) \mathrm{d}t  \bigg], \\
    C_1 &= \mathbb{E}_{\bm{G}\sim\mathbb{Q}^{\Pi^{\ast}}}\! \bigg[ \frac{1}{2}\int^T_0 \!\! \gamma_t^2 \left\| \bm{G}_T - \bm{D}^{\Pi^{\ast}}\!\!(\bm{G}_t,t) \right\|^2 \! \mathrm{d}t \bigg].
\end{split}
\end{equation}
From the definition of the graph mixture (Eq.~\eqref{eq:graph_mixture}), the following identity holds for all $t\in [0,T]$:
\begin{align}
    \mathbb{E}_{\bm{G}\sim\mathbb{Q}^{\Pi^{\ast}}}  \bm{D}^{\Pi^{\ast}}\!\!(\bm{G}_t,t) = \mathbb{E}_{\bm{G}\sim\mathbb{Q}^{\Pi^{\ast}}} \bm{G}_T ,
\end{align}
which gives the following result:
\begin{align}
    \mathcal{E} &= \mathbb{E}_{\bm{G}\sim\mathbb{Q}^{\Pi^{\ast}}}\! \bigg[ \frac{1}{2}\int^T_0 \!\! \gamma_t^2 \,  \Big( \bm{s}_{\theta}(\bm{G}_t,t) - \bm{G}_T \Big)^T \Big( \bm{G}_T - \bm{D}^{\Pi^{\ast}}\!\!(\bm{G}_t,t) \Big) \mathrm{d}t  \bigg] \\
    &= \mathbb{E}_{\bm{G}\sim\mathbb{Q}^{\Pi^{\ast}}}\! \bigg[ \frac{1}{2}\int^T_0 \!\! \gamma_t^2 \,  \Big( \bm{s}_{\theta}(\bm{G}_t,t) - \bm{G}_T \Big)^T \Big( \bm{G}_T - \bm{G}_T \Big) \mathrm{d}t  \bigg]
    = 0
\end{align}
Therefore, we can conclude that minimizing Eq.~\eqref{eq:objective_derivation} is equivalent to minimizing the following loss:
\begin{align}
    \mathbb{E}_{\bm{G}\sim\mathbb{Q}^{\Pi^{\ast}}}\! \bigg[ \frac{1}{2}\int^T_0 \!\! \gamma_t^2 \left\| \bm{s}_{\theta}(\bm{G}_t,t) - \bm{G}_T \right\|^2 \! \mathrm{d}t  \bigg] 
    \label{eq:graph_mixture_matching}
\end{align}
which corresponds to the graph mixture matching presented in Eq.~\eqref{eq:objective}.

\subsection{Analytical computation of graph mixture matching
\label{sec:app:derivation:prob}}
In order to practically use the graph mixture matching (Eq.~\eqref{eq:objective}), we provide the analytical form of the distribution $p_{t|0,T}(\bm{G}_t|\bm{G}_0,\bm{G}_T)$. Notice that by construction, the OU bridge mixture with a fixed starting point $\bm{G}_0$ and an endpoint $\bm{G}_T$ coincides with the reference OU process in Eq.~\eqref{eq:general_ou_process} with a fixed starting point $\bm{G}_0$ and an endpoint $\bm{G}_T$. Thereby, $p_{t|0,T}(\bm{G}_t|\bm{G}_0,\bm{G}_T)$ is equal to $\tilde{p}_{t|0,T}(\bm{G}_t|\bm{G}_0,\bm{G}_T)$ where $\tilde{p}$ denotes the marginal probability of the reference OU process of Eq.~\eqref{eq:general_ou_process}. Using the Bayes theorem, we can derive the following:
\begin{align}
    \tilde{p}(\bm{G}_t|\bm{G}_0,\bm{G}_T) = \frac{\tilde{p}(\bm{G}_t,\bm{G}_T|\bm{G}_0)}{\tilde{p}(\bm{G}_T|\bm{G}_0)} 
    = \frac{\tilde{p}(\bm{G}_T|\bm{G}_t,\bm{G}_0) \; \tilde{p}(\bm{G}_t|\bm{G}_0)}{\tilde{p}(\bm{G}_T|\bm{G}_0)}
    = \frac{\tilde{p}(\bm{G}_T|\bm{G}_t) \; \tilde{p}(\bm{G}_t|\bm{G}_0)}{\tilde{p}(\bm{G}_T|\bm{G}_0)} ,
    \label{eq:bayes}
\end{align}
where the last equality is due to the Markov property of the OU process. Note that the transition distributions of the reference OU process are Gaussian with the mean and the covariance as follows:
\begin{equation}
\begin{split}
    &\tilde{p}_{b|a}(\bm{G}_b|\bm{G}_a) = \mathcal{N}(\bm{G}_b; u_{b|a}\bm{G}_a, u_{b|a}^2v_{b|a} \mathbf{I}) \;\;\text{for} \;\; 0\leq a<b\leq T, \\
    & \text{where} \;\; u_{b|a} \coloneqq \exp\Big(\alpha\int^b_a \sigma_{\tau}^2\mathrm{d}\tau \Big) \;\;,\;\; v_{b|a} \coloneqq \frac{1}{2\alpha}\Big( 1 - u_{b|a}^{-2}\Big) .
\end{split}\label{eq:transition_distribution}
\end{equation}
Therefore, the distribution $p(\bm{G}_t|\bm{G}_0,\bm{G}_T)$ is also Gaussian resulting from the product of Gaussian distributions, where the mean $\mu^{\ast}_t$ and the covariance $\mathbf{\Sigma}^{\ast}_t$ have analytical forms as follows:
\begin{align}
    \mu^{\ast}_t = \frac{v_{T|t}}{u_{t|0}v_{T|0}}\bm{G}_0 + \frac{v_{t|0}}{u_{T|t}v_{T|0}}\bm{G}_T 
    \;\;, \;\; \mathbf{\Sigma}^{\ast}_t = \frac{v_{T|t}v_{t|0}}{v_{T|0}}\mathbf{I} .
\end{align}
The mean and the covariance can be simplified by using the hyperbolic sine function as follows:
\begin{align}
    \mu^{\ast}_t = \frac{\sinh\left( \varphi_{T}-\varphi_{t} \right)}{\sinh\left( \varphi_{T} \right)} \bm{G}_0 
    + \frac{\sinh\left( \varphi_{t} \right)}{\sinh\left( \varphi_{T} \right)} \bm{G}_T ,
    \quad \mathbf{\Sigma}^{\ast}_t =  \frac{1}{\alpha}\frac{\sinh\left( \varphi_{T}-\varphi_{t} \right) \sinh\left( \varphi_{t} \right)}{ \sinh\left( \varphi_{T} \right)} \mathbf{I} ,
    \label{eq:transition_mean_variance}
\end{align}
where $\varphi_t \coloneqq \alpha\beta_t = \alpha\int^{t}_{0}\sigma_{\tau}^2\mathrm{d}\tau $.

\subsection{GruM as a stochastic interpolant \label{sec:app:derivation:interpolant}}
Recently, \citet{albergo23interpolant} introduced the concept of \emph{stochastic interpolant} which unifies the framework for diffusion models from the perspective of continuous-time stochastic processes.

From the results of Eq.~\eqref{eq:transition_mean_variance}, we can represent the OU bridge mixture as a stochastic interpolant between the distributions $\Gamma$ and $\Pi^{\ast}$ as follows:
\begin{align}
    &\bm{G}_t = \frac{\sinh\left( \varphi_{T}-\varphi_{t} \right)}{\sinh\left( \varphi_{T} \right)} \bm{G}_0 
    + \frac{\sinh\left( \varphi_{t} \right)}{\sinh\left( \varphi_{T} \right)} \bm{G}_T 
    + \left( \frac{1}{\alpha}\frac{\sinh\left( \varphi_{T}-\varphi_{t} \right) \sinh\left( \varphi_{t} \right)}{ \sinh\left( \varphi_{T} \right)} \right)^{1/2} \bm{Z} .
    \label{eq:interpolant}
\end{align}
where $\bm{G}_0$, $\bm{G}_T$, and $\bm{Z}$ are random variables sampled independently from the distributions $\Gamma$, $\Pi^{\ast}\!$, and $\mathcal{N}(\bm{0},\mathbf{I})$, respectively.
Eq.~\eqref{eq:interpolant} shows that $\bm{G}_t$ is linear in both the starting point $\bm{G}_0\sim\Gamma$ and the endpoint $\bm{G}_T\sim\Pi^{\ast}\!$. 
Note that our proposed graph mixture matching is different from the loss introduced in \citet{albergo23interpolant}, as graph mixture matching does not require estimation of the score function. Additionally, we further derive the score function of our GruM in Section~\ref{sec:app:derivation:probability_flow}.

\subsection{Understanding the informative prior as regularizing the graph mixture}
\citet{wu22bridge} introduces incorporating prior information into the generative process, for example injecting physical and statistical information.
To be specific, given a generative process:
\begin{align*}
    \mathrm{d}\bm{G}_t = \eta(\bm{G}_t,t)\mathrm{d}t + \sigma_t\mathbf{W}_t ,
\end{align*}
\citet{wu22bridge} modifies the drift by adding a prior function $f(\cdot,t)$ as follows:
\begin{align}
    \mathrm{d}\bm{G}_t = \Big( \underbrace{\sigma_t f(\bm{G}_t,t) + \eta(\bm{G}_t,t)}_{\eta_{R}(\bm{G}_t,t)} \Big)\mathrm{d}t + \sigma_t\mathbf{W}_t , 
    \label{eq:informative_prior}
\end{align}
where $f(\cdot,t)$ is designed to be a force defined as $f(\cdot, t)=-\nabla \bm{E}(\cdot)$ where $E(\cdot)$ is a problem-dependent energy function.
Although \citet{wu22bridge} shows that incorporating prior information is beneficial for the generation of stable molecules or realistic 3D point clouds, how this modification leads to better performance was not fully explained. 

Notably, from the perspective of our framework, we can interpret the incorporation of the prior information as modifying the generative path toward an energy-regularized result.
To be precise, given a generative process modeled by the OU bridge mixture as in Eq.~\eqref{eq:ou_bridge_mixture}, adding the prior function $f(\cdot,t)$ to the drift can be written as follows:
\begin{align}
   \eta_{R}(\bm{G}_t,t) = \alpha\sigma^2_t\bm{G}_t + \frac{\sigma_t^2}{v_t}\left[ \frac{1}{u_t}\Big( \bm{D}^{\Pi^{\ast}}\!\!(\bm{G}_t,t) + \frac{u_tv_t}{\sigma_t}f(\bm{G}_t,t) \Big) - \bm{G}_t \right] ,
\end{align}
which is equivalent to regularizing the graph mixture with the weighted prior function as follows:
\begin{align}
    \bm{D}^{\Pi^{\ast}}_{R}\!(\bm{G}_t,t) \coloneqq \bm{D}^{\Pi^{\ast}}\!\!(\bm{G}_t,t) + \frac{u_tv_t}{\sigma_t}f(\bm{G}_t,t) .
\end{align}
Since the weight of the prior function converges to 0 through the generative process:
\begin{align*}
    \frac{u_tv_t}{\sigma_t} = \frac{\exp\left({\alpha\int^T_t\sigma_{\tau}^2\mathrm{d}\tau}\right)-\exp\left({-\alpha\int^T_t\sigma_{\tau}^2\mathrm{d}\tau}\right)}{2\alpha\sigma_t} \rightarrow 0 \quad \text{as} \quad t\rightarrow T,
\end{align*}
we can see that $\bm{D}^{\Pi^{\ast}}_{R}\!$ converges to the original graph mixture $\bm{D}^{\Pi^{\ast}}\!$ where the convergence is determined by the prior function. 
By defining $f(\cdot, t)=-\nabla \bm{E}(\cdot)$ where $\bm{E}$ is an energy function, for example, potential energy for the 3D molecules or Riesz energy for the 3D point cloud, the regularized graph mixture has the following representation:
\begin{align}
    \bm{D}^{\Pi^{\ast}}_{R}\!(\bm{G}_t,t) = \bm{D}^{\Pi^{\ast}}\!\!(\bm{G}_t,t) - \frac{u_tv_t}{\sigma_t}\nabla\bm{E}(\bm{G}_t) .
\end{align}
Thereby, $\bm{D}^{\Pi^{\ast}}_{R}\!$ follows a path that minimizes the energy function $\bm{E}$ through the generative process. Therefore, the generative process is guided toward the regularized graph mixture which results in samples that achieve desired physical properties, for instance, stable 3D-structured molecules or point clouds.

\subsection{Associated probability flow ODE of GruM \label{sec:app:derivation:probability_flow}}
Since we have derived the reverse-time diffusion process of the OU bridge mixture in Section~\ref{sec:app:derivation:reverse}, we can further derive its associated probability flow ODE~\citep{song21sde}, i.e. a deterministic process that shares the same marginal density with the OU bridge mixture.

First, the OU bridge mixture is modeled by the following SDE:
\begin{align*}
    \mathrm{d}\bm{G}_t = \left[ \alpha\sigma_t^2\bm{G}_t + \frac{\sigma_t^2}{v_t} \left( \frac{1}{u_t} \bm{D}^{\Pi^{\ast}\!}(\bm{G}_t,t) - \bm{G}_t \right) \right] \mathrm{d}t + \sigma_t\mathrm{d}\mathbf{W}_t ,
\end{align*}
where the scalar functions $u_t$ and $v_t$, and the graph mixture $\bm{D}^{\Pi^{\ast}}\!$ are defined as:
\begin{align*}
     u_t = \exp\Big( \alpha\int^T_t \sigma^2_{\tau} \mathrm{d}\tau \Big) ,\;\; 
     v_t = \frac{1}{2\alpha}(1-u_t^{-2}) ,\;\; 
     \bm{D}^{\Pi^{\ast}\!}(\bm{G}_t,t) = \int \bm{g} \frac{p^{\bm{g}}_t(\bm{G}_t)}{p_t(\bm{G}_t)} \Pi^{\ast}(\mathrm{d}\bm{g}) .
\end{align*}
Then using the results of Section~\ref{sec:app:derivation:reverse}, the reverse-time diffusion process of the OU bridge mixture is modeled by the following SDE:
\begin{align*}
    \mathrm{d}\overbar{\bm{G}}_t = \left[ -\alpha\sigma_{T\!\!-\!t}^2\overbar{\bm{G}}_t + \frac{\sigma_{T\!\!-\!t}^2}{\overbar{u}_t^2\overbar{v}_t}\Big( \overbar{u}_t\bm{D}^{\Gamma}(\overbar{\bm{G}}_t,t) - \overbar{\bm{G}}_t \Big) \right]\mathrm{d}t + \sigma_{T\!\!-\!t}\mathrm{d}\mathbf{W}_t, 
\end{align*}
where the scalar functions $\overbar{u}_t$ and $\overbar{v}_t$, and the reversed graph mixture $\bm{D}^{\Gamma}$ are defined as:
\begin{align*}
    \overbar{u}_t = \exp\left(\alpha\int^{T\!\!-\!t}_0\sigma_{\tau}^2\mathrm{d}\tau\right), 
    \quad \overbar{v}_t = \frac{1}{2\alpha}\left(1 - \overbar{u}_t^{-2}\right),
    \quad \bm{D}^{\Gamma}(\overbar{\bm{G}}_t,t) = \int \bm{x}\frac{q^{\bm{x}}_t(\overbar{\bm{G}}_t)}{q_t(\overbar{\bm{G}}_t)} \Gamma(\mathrm{d}\bm{x}).
\end{align*}
From the relation between the diffusion process and its reverse-time diffusion process (for instance, Eq.~\eqref{eq:general_reference} and Eq.~\eqref{eq:reverse_reference}), the score function of the OU bridge mixture can be computed as follows:
\begin{align}
    \nabla_{\bm{G}_t}\!\log p_t(\bm{G}_t) = \frac{1}{v_t} \bigg( \frac{1}{u_t} \bm{D}^{\Pi^{\ast}\!}(\bm{G}_t,t) - \bm{G}_t \bigg) + 
    \frac{1}{\overbar{u}_{T\!-\!t}^2\overbar{v}_{T\!-\!t}}\bigg( \overbar{u}_{T\!-\!t}\bm{D}^{\Gamma}(\bm{G}_t, T\!-\!t) - \bm{G}_t \bigg) .
    \label{eq:score}
\end{align}
Therefore, the associated probability flow ODE can be derived as follows:
\begin{align}
    \frac{\mathrm{d}\bm{G}_t}{\mathrm{d}t} &=  \alpha\sigma_t^2\bm{G}_t + \frac{\sigma_t^2}{v_t} \left( \frac{1}{u_t} \bm{D}^{\Pi^{\ast}\!}(\bm{G}_t,t) - \bm{G}_t \right) - \frac{1}{2}\sigma_t^2 \nabla_{\bm{G}_t}\!\log p_t(\bm{G}_t)  \\
    &= \alpha\sigma_t^2\bm{G}_t + \frac{\sigma_t^2}{2v_t} \left( \frac{1}{u_t} \bm{D}^{\Pi^{\ast}\!}(\bm{G}_t,t) \!-\! \bm{G}_t \!\right)  
    - \frac{\sigma_t^2}{2\overbar{u}_{T\!-\!t}^2\overbar{v}_{T\!-\!t}}\bigg( \overbar{u}_{T\!-\!t} \bm{D}^{\Gamma}(\bm{G}_t, T\!-\!t) \!-\! \bm{G}_t \!\bigg)  .
\end{align}
To practically use the probability flow ODE as a generative model, the graph mixtures $\bm{D}^{\Pi^{\ast}\!}(\cdot, t)$ and $\bm{D}^{\Gamma}(\cdot, t)$ should be approximated by the neural networks $\bm{s}_{\theta}(\cdot, t)$ and $\bm{s}_{\phi}(\cdot, t)$, respectively. $\bm{s}_{\theta}$ can be trained using the graph mixture matching (Eq.~\eqref{eq:graph_mixture_matching}). $\bm{s}_{\phi}$ also can be trained in a similar way where the trajectories are sampled from the reverse-time process of the OU bridge mixture.

In particular, from the result of Eq.~\eqref{eq:score}, we can see that learning the score function of the mixture process is not interchangeable with learning the graph mixture since the score function additionally requires the knowledge of the reversed graph mixture $\bm{D}^{\Gamma}$. Our mixture process differs from the denoising diffusion processes for which learning the score function is equivalent to recovering clean data from its corrupted version~\citep{kingma21variance-preserving}. The difference originates from the difference in the construction of the generative process, where denoising diffusion processes are derived by reversing the forward noising processes while our mixture process is built as a mixture of bridge processes without relying on the time-reversal approach. We further discuss the difference between our framework and the denoising diffusion models in Section~\ref{sec:app:derivation:comparison}.

\subsection{Comparison with Denoising Diffusion Models \label{sec:app:derivation:comparison}}
Here we explain in detail the difference between our framework and previous denoising diffusion models.

\paragraph{Comparison of the generative processes}
The main difference with the denoising diffusion models~\citep{ho20ddpm, song21sde} is in the different generative processes. 
While denoising diffusion models derive the generative process by reversing the forward noising process, our method constructs the generative process using the mixture of OU bridge processes described in Eq.~\eqref{eq:ou_bridge} which does not rely on the time-reversal approach. Due to the difference in the generative process, our method demonstrates two distinct properties: First, the mixture process defines an exact transport from an arbitrary prior distribution to the data distribution by construction. In contrast, denoising diffusion processes are not an exact transport to the data distribution since the forward noising processes require infinitely long diffusion time in order to guarantee convergence to the prior distribution~\citep{franzese2023much}.

Furthermore, our framework does not suffer from the restrictions of denoising diffusion models. Denoising diffusion models require $p_{prior}$ to be approximately independent of the data distribution $\Pi^{\ast}$, e.g. Gaussian, as the perturbation process decouples $p_{prior}$ from $\Pi^{\ast}$, and further this decoupling requires infinitely long diffusion time $T$. On the contrary, our framework does not have any constraints on the prior distribution $p_{prior}$ and does not require large $T$, since the OU bridge mixture can be defined between two arbitrary distributions for any $T>0$, where its drift forces the process to the terminal distribution regardless of the initial distribution. 
Therefore, the OU bridge mixture provides flexibility for our generative framework in choosing the prior distribution and the finite terminal time while maintaining the generative process to be an exact transport from the prior to the data distribution.

\paragraph{Comparison of the training objectives}
We further compare our training objective in Eq.~\eqref{eq:graph_mixture_matching} with the training objectives of denoising diffusion models.
First, we clarify that learning the graph mixture is not equivalent to learning the score function for the mixture process of GruM. As derived in Eq.~\eqref{eq:score} of Section~\ref{sec:app:derivation:probability_flow}, the score function of the OU bridge mixture additionally requires the knowledge of the reversed graph mixture $\bm{D}^{\Gamma}$, thus learning the score function needs to predict not only the graph mixture but also the reversed graph mixture. In contrast, the training objectives of denoising diffusion models are interchangeable~\citep{kingma21variance-preserving}, i.e., learning the score function of the denoising diffusion process is equivalent to recovering clean data from its corrupted version.
This difference in the training objective originates from the difference in the generative process, which we have discussed in detail in the previous paragraphs.

Furthermore, our training objective differs from the objectives of previous works~\citep{diffusion/images/3} that aim to recover clean data from its corrupted version. While our method learns the graph mixture, i.e. the probable graph represented as the weighted mean of data, \citet{diffusion/images/3} aims to predict the exact final result which could be problematic as the prediction would be highly inaccurate in early steps which may lead the process in the wrong direction.
It should be noted that the goal of Eq.~\eqref{eq:graph_mixture_matching} is to estimate the graph mixture, i.e. the weighted mean of data, not to predict the exact graph as in \citet{diffusion/images/3}. This is because Eq.~\eqref{eq:graph_mixture_matching} is derived from Eq.~\eqref{eq:objective} which minimizes the difference between our model prediction and the ground truth graph mixture. We emphasize that learning graph mixture (Eq.~\eqref{eq:objective}) is only feasible for the OU bridge mixture and cannot be used for denoising diffusion models due to the difference in the generative processes.

In the perspective of the mathematical formulation of the training objective, Eq.~\eqref{eq:graph_mixture_matching} differs from the objective of \citet{diffusion/images/3} in two parts:
(1) The computation of the expectation for the squared error loss term is different. The expectation is computed by sampling from the trajectory of the diffusion process, where our GruM uses the OU bridge mixture while previous works use the denoising diffusion process. These two processes are not the same and therefore result in different objectives. 
(2) The weight function in the loss is different. The weight function $\gamma_t$ of Eq.~\eqref{eq:objective} is different from the weight function used in denoising diffusion models, and $\gamma_t$ is derived to guarantee that minimizing Eq.~\eqref{eq:graph_mixture_matching} is equivalent to minimizing the KL divergence between the data distribution and the terminal distribution of our approximated process.

Another line of works on discrete diffusion models~\citep{austin21discrete, hoogeboom21argmax, vignac22digress} aims to predict the probabilities of each state of the final data to be generated. Since these works predict the probabilities, they are limited to data with a finite number of states and cannot be applied to data with continuous features. In contrast, our method directly predicts the weighted mean of the data (i.e., graph mixture) instead of the probabilities, which can be applied to data with continuous features, for example, 3D molecules as well as the discrete data, which we experimentally validate to be effective. It is worth noting that our GruM is a continuous diffusion model, and thereby our framework can leverage the advanced sampling strategies that reduce the sampling time or improve sample quality~\citep{campbell2022continuous}, whereas the discrete diffusion models are forced to use a simple ancestral sampling strategy.

\begin{table*}[t]
    \caption{\textbf{Comparison of graph diffusion models.} }\label{tab:comparison}
    \vspace{-0.075in}
    \centering
    \resizebox{1.0\textwidth}{!}{
    \renewcommand{\arraystretch}{1.0}
    \renewcommand{\tabcolsep}{5pt}
    \begin{tabular}{l@{\hskip -2pt} c c c c c c c}
    \toprule
    & Explicitly model & Simulation-free & Arbitrary prior & Does not require large & Learning object & Model \\
    & graph topology   & training        & distribution    & diffusion time $T$     & is bounded & prediction \\
    \midrule
    EDM~\citep{hoogeboom22edm} & \redx & \greencheck & \redx & \redx & \greencheck & Noise \\
    GDSS~\citep{jo22gdss} & \redx & \greencheck & \redx & \redx & \redx & Score \\
    Bridge~\citep{wu22bridge} & \redx & \redx & \greencheck & \greencheck & \redx & Drift \\
    \midrule
    \textbf{GruM} (Ours) & \greencheck & \greencheck & \greencheck & \greencheck & \greencheck & Data \\
    \bottomrule
    \end{tabular}}
\vspace{-0.1in}
\end{table*}

\subsection{Additional explanation on relevant works}

\paragraph{Related works on diffusion bridges \label{sec:app:derivation:comparison_bridge}}
Recent works~\citep{peluchetti21mixture, wu22bridge, ye22firsthitting, liu23bridge} introduce learning the generative process using a mixture of diffusion bridge processes, instead of learning to reverse the noising process as in denoising diffusion models.
\citet{peluchetti21mixture} introduces a diffusion mixture representation that constructs a generation process as a mixture of the bridge processes. \citet{wu22bridge} injects physical information into the process by adding informative prior to the drift, while \citet{ye22firsthitting} and \citet{liu23bridge} extend the bridge process to constrained domains.

\paragraph{Related works on graph generative models}
Recently, graph diffusion models~\citep{niu20edpgnn, jo22gdss, hoogeboom22edm, vignac22digress} have made large progress on generating general graphs as well as molecular graphs.
EDP-GNN~\citep{niu20edpgnn} aims to generate the adjacency matrix by learning the score function of the denoising diffusion process, while GDSS~\citep{jo22gdss} proposes a framework for generating the nodes and edges simultaneously by learning the joint score function that captures the node-edge dependency. However, learning the score function is ill-suited for modeling the graph topology as it does not explicitly consider the graph structures, and further could be problematic due to the diverging score function.
On the other hand, discrete diffusion model~\citep{vignac22digress} proposes to model the noising process as successive graph edits, preserving the discrete structure during the diffusion process. However, this is not a desirable solution for real-world graph generation tasks since it applies to graphs with categorical node and edge attributes, and cannot be alone applied to graphs with continuous features, such as the 3D coordinates of atoms. 
We summarize the comparison between closely related graph diffusion models in Table~\ref{tab:comparison}.

\section{Details of GruM \label{sec:app:details}}
In this section, we provide the details of the training and sampling methods of GruM, describe the models used in our experiments, and further discuss the hyperparameters of GruM.

\subsection{Overview~\label{sec:app:details:overview}}
We provide the pseudo-code of the training and sampling of our generative framework in Algorithm~\ref{alg:training} and \ref{alg:sampling}, respectively.
We further provide the implementation details of the training and sampling for each generation task in Section~\ref{sec:app:exp}.

\begin{figure}[t!]
\vspace{-0.12in}
\centering
\begin{minipage}{0.49\linewidth}
\centering
\begin{algorithm}[H]
    \caption{ Training of GruM}\label{alg:training}
        \textbf{Input:} Model $\bm{s}_{\theta}$, constant $\epsilon$ \\
        \textbf{For each epoch:} \phantom{a}
    \begin{algorithmic}[1]
        \STATE Sample graph $\bm{G}$ from the training set
        \STATE $N \leftarrow$ \text{number of nodes of} $\bm{G}$
        \STATE Sample $t\sim[0,T-\epsilon]$ and $\bm{G}_0\sim\mathcal{N}(0,\mathbf{I}_N)$
        \STATE Sample $\bm{G}_t\sim p_{t|0,T}(\bm{G}_t|\bm{G}_0,\bm{G})$ 
        \COMMENT{Section~\ref{sec:app:derivation:prob}}
        \STATE $\gamma_t \leftarrow {\sigma_{t}}/{u_tv_t}$
        \STATE $\mathcal{L}_{\theta} \leftarrow \gamma_t^2 \| \bm{s}_{\theta}(\bm{G}_t,t) - \bm{G} \|^2$ 
        \COMMENT{Eq.~\eqref{eq:graph_mixture_matching}}
        \STATE Update $\theta$ using $\mathcal{L}_{\theta}$
    \end{algorithmic}
\end{algorithm}
\vspace{-0.275in}
\begin{algorithm}[H]
    \caption{Sampling of GruM}\label{alg:sampling}
        \textbf{Input:} Trained model $\bm{s}_{\theta}$, number of sampling steps $K$, diffusion step size $\mathrm{d}t$
    \begin{algorithmic}[1]
        \STATE Sample number of nodes $N$ from the training set.
        \STATE $\bm{G}_0\sim \mathcal{N}(0,\mathbf{I}_N)$
        \COMMENT{Start from noise}
        \STATE $t\leftarrow 0$
        \FOR{$k=1$ \textbf{to} $K$}
            \STATE $\eta\leftarrow \alpha\sigma_t^2\bm{G}_t + \frac{\sigma_t^2}{v_t} \left(\frac{1}{u_t}\bm{s}_{\theta}(\bm{G}_t,t) - \bm{G}_t\right)$ 
            \STATE $\mathbf{w}\sim\mathcal{N}(0,\mathbf{I}_N)$ 
            \STATE $\bm{G}_{t+\mathrm{d}t} \!\leftarrow \eta\mathrm{d}t +  \sigma_t \sqrt{\mathrm{d} t} \mathbf{w}$
            \COMMENT{Euler-Maruyama}
            \STATE $t \leftarrow t+\mathrm{d}t$
        \ENDFOR
        \STATE $\bm{G} \leftarrow \texttt{quantize}(\bm{G}_t)$
        \COMMENT{Quantize if necessary}
        \STATE \textbf{Return:} Graph $\bm{G}$
    \end{algorithmic}
\end{algorithm}
\end{minipage}
\hfill
\begin{minipage}{0.49\linewidth}
\centering
\begin{algorithm}[H]
    \caption{ PC Sampler for GruM }\label{alg:pc_sampler}
        \textbf{Input:} Trained models $\bm{s}_{\theta}$ and $\bm{s}_{\phi}$ (described in Section~\ref{sec:app:derivation:probability_flow}), number of sampling steps $K$, number of correction steps per prediction $M$, diffusion step size $\mathrm{d}t$, score-to-noise ratio $r$  \\
        \textbf{Output:} Sampled graph $\bm{G}$
    \begin{algorithmic}[1]
        \STATE Sample number of nodes $N$ from the training set.
        \STATE $\bm{G}_0\sim \mathcal{N}(0,\mathbf{I}_N)$
        \COMMENT{Start from noise}
        \STATE $t\leftarrow 0$
        \FOR{$k=1$ \textbf{to} $K$}
            \STATE $\eta \leftarrow \alpha\sigma_t^2\bm{G}_{t} + \frac{\sigma_t^2}{v_t}\left(\frac{1}{u_t}\bm{s}_{\theta}(\bm{G}_{t},t) - \bm{G}_{t}\right)$
            \STATE $\mathbf{w}\sim\mathcal{N}(0,\mathbf{I}_N)$ 
            \STATE $\tilde{\bm{G}}_t \leftarrow \eta\mathrm{d}t +  \sigma_t \sqrt{\mathrm{d} t} \mathbf{w}$
            \COMMENT{Predictor}
            \FOR[Corrector loop]{$m=1$ \textbf{to} $M$}
                \STATE $\bm{D},\, \bar{\bm{D}} \leftarrow \bm{s}_{\theta}(\tilde{\bm{G}_{t}},t),\; \bm{s}_{\phi}(\tilde{\bm{G}_{t}},T\!-\!t)$
                \STATE $s \!\leftarrow\! \texttt{Compute\textunderscore Score}(\bm{D},\bar{\bm{D}}, \tilde{\bm{G}_{t}})$\!\!
                \COMMENT{Eq.\eqref{eq:score}}
                \STATE $\mathbf{w}\sim\mathcal{N}(0,\mathbf{I}_N)$ 
                \STATE $\epsilon \leftarrow 2\left(r\|\mathbf{w}\|_2/\|s\|_2  \right)^2$ 
                \STATE $\tilde{\bm{G}}_t \!\leftarrow\! \texttt{Corrector}(\tilde{\bm{G}_{t}},s,\epsilon)$ \!
            \ENDFOR
            \STATE $\bm{G}_{t+\mathrm{d}t} \leftarrow \tilde{\bm{G}}_t$
            \STATE $t \leftarrow t+\mathrm{d}t$
        \ENDFOR
        \STATE $\bm{G}\leftarrow \texttt{quantize}(\bm{G}_t)$
        \COMMENT{Quantize if necessary}
        \STATE \textbf{Return:} Graph $\bm{G}$ 
    \end{algorithmic}
\end{algorithm}
\end{minipage}
\vspace{-0.1in}
\end{figure}

\subsection{Training objectives \label{sec:app:details:objective}}

\paragraph{Random permutation}
The general graph datasets, namely Planar and SBM, contain only 200 graphs. Thus to ensure the permutation invariant nature of the graph dataset, we apply random permutation to the graphs of the training set during training. To be specific, for a graph data $\bm{G}=(\bm{X},\bm{A})$ in the training set and random permutation matrix $\bm{P}$, we use the permuted data $(\bm{P}^{T}\bm{X}, \bm{P}^{T}\bm{A}\bm{P})$ for training, where $\bm{P}^T$ denotes the transposed matrix. We empirically found that this leads to more stable training.

\paragraph{Simplified loss \label{sec:app:details:simplified}}
We provide the explicit form of simplified loss explained in Section~\ref{sec:method:objective}, which uses constant loss coefficient $c$ instead of the time-dependent $\gamma_t$ as follows:
\begin{align}
    \mathcal{L}(\theta) = \mathbb{E}_{\bm{G}\sim\mathbb{Q}^{\Pi^{\ast}}}\! \bigg[ \frac{1}{2}\int^T_0 \!\! c^2 \left\| \bm{s}_{\theta}(\bm{G}_t,t) - \bm{G}_T \right\|^2 \! \mathrm{d}t  \bigg] .
    \label{eq:simplified_objective}
\end{align}
We empirically found that using this loss is beneficial for the generation of continuous features such as eigenvectors of the graph Laplacian or 3D coordinates.

\paragraph{Attributed graphs \label{sec:app:details:objective:attributed}}
Especially for the generation of attributed graphs $\bm{G}=(\bm{X},\bm{A})$, the graph mixture matching for $\bm{X}$ and $\bm{A}$ can be derived from Eq.~\eqref{eq:objective}. Specifically, for the model $\bm{s}_{\theta}(\cdot,t)=(\bm{s}^{X}_{\theta}(\cdot,t),\bm{s}^{A}_{\theta}(\cdot,t))$, we use the following objective:
\begin{align}
    \mathcal{L}(\theta) =  \mathbb{E}_{\bm{G}\sim\mathbb{Q}^{\Pi^{\ast}}}\! \bigg[ &\frac{1}{2}\int^T_0 \!\! \gamma_{1,t}^2 \left\| \bm{s}^{X}_{\theta}(\bm{G}_t,t) - \bm{X}_T \right\|^2 \! \mathrm{d}t +
    \frac{\lambda}{2}\int^T_0 \!\! \gamma_{2,t}^2 \left\| \bm{s}^{A}_{\theta}(\bm{G}_t,t) - \bm{A}_T \right\|^2 \! \mathrm{d}t
    \bigg]
\end{align}
where $\lambda$ is the hyperparameter. We use $\lambda=5$ for all our experiments and empirically observed that changing $\lambda$ did not make much difference for sufficient training epochs.

\subsection{Model architecture~\label{sec:app:details:model}}
For the general graph and 2D molecule generation tasks, we leverage the graph transformer network introduced in \citet{dwivedi20transformer} and \citet{vignac22digress}. The node features and the adjacency matrices are updated with multiple attention layers with global features obtained by the self-attention-based FiLM layers~\citep{perez18film}. We additionally use the higher-order adjacency matrices following \citet{jo22gdss}. 
For general graph generation tasks, we add the sigmoid function to the output of the model since the entries of the weighted mean of the data are supported in the interval $[0,1]$.
For 2D molecule generation tasks, we apply the softmax function to the output of the node features to model the one-hot encoded atom types, and further apply the sigmoid function to the output of the adjacency matrices. Note that we do not use the structural augmentation as in \citet{vignac22digress}.
For the 3D molecule generation task, we use EGNN~\citep{satorras21enflow} to model the E(3) equivariance of the molecule data. We additionally add a softmax function at the last layer to model the one-hot encoded atom types.

\subsection{Sampling from GruM \label{sec:app:details:sampler}}
Sampling from the generative model requires solving the SDE of Eq.~\eqref{eq:generative_model}. If our model $\bm{s}_{\theta}$ can closely approximate the graph mixture, a simple Euler-Maruyama method would be enough to simulate the generative model, which is true for most of the experiments.
Since $\bm{s}_{\theta}$ is trained on the marginal density $p_t$, $\bm{G}_t$ outside of $p_t$ could cause incorrect predictions that lead to an undesired result. 
To address the limitation, we may leverage the predictor-corrector (PC) sampling method introduced in \citet{song21sde}. Using the corrector method such as Langevin dynamics~\citep{song21sde}, we force $\bm{G}_t$ to be drawn from $p_t$ which ensures a more accurate estimation of the graph mixture. The score function to be used for the corrector can be computed as in Eq.~\eqref{eq:score} of Section~\ref{sec:app:derivation:probability_flow}. We provide the pseudo-code of the predictor-only sampler and the PC sampler for our GruM in Algorithm~\ref{alg:sampling} and \ref{alg:pc_sampler}.
Note that our GruM does not require additional time for sampling compared to the denoising diffusion models, since the generation is equivalent to solving the SDE where the drift is computed each step from the forward pass of the model.

\subsection{Hyperparameters of GruM \label{sec:app:details:hyperparameter}}
The generative process of GruM modeled as the OU bridge mixture is uniquely determined with the noise schedule $\sigma_t$ and constant $\alpha$. Through our experiments, we use $\alpha=-1/2$ and a decreasing linear noise schedule, starting from $\sigma_0^2$ and ends in $\sigma_0^2$ defined as follows:
\begin{align}
    \sigma_t^2 = (1-t)\sigma_0^2 + t\sigma_1^2 \;\; \text{for} \;\; 0 < \sigma_1 < \sigma_0 < 1
\end{align}
We perform a grid search for the hyperparameters $\sigma_0$ and $\sigma_1$ in $\{ 0.4, 0.6, 0.8, 1.0 \}$ and $\{ 
0.1, 0.2, 0.3 \}$, respectively, where the search space slightly differs for each generation task.

\vspace{-0.075in}
\section{Experimental Details \label{sec:app:exp}}

\vspace{-0.05in}
\subsection{General graph generation \label{sec:app:exp:general}}

\vspace{-0.05in}
\paragraph{Datasets and evaluation metrics}
We evaluate the quality of generated graphs on three graph datasets used as benchmarks in \citet{martinkus22spectre}: 
\textbf{Planar} graph dataset consists of 200 synthetic planar graphs where each graph has 64 nodes. We determine that a graph is a valid Planar graph if it is connected and planar. 
\textbf{Stochastic Block Model} (SBM) graph dataset consists of 200 synthetic stochastic block model graphs with the number of communities uniformly sampled between 2 and 5, where the number of nodes in each community is uniformly sampled between 20 and 40. The probability of the inter-community edges and the intra-community edges are 0.3 and 0.05, respectively. We determine that a graph is a valid SBM graph if it has the number of communities between 2 and 5, the number of nodes in each community between 20 and 40, and further using the statistical test introduced in \citet{martinkus22spectre}.
\textbf{Proteins} graph dataset~\citep{protein} consists of 918 real protein graphs with up to 500 nodes in each graph. The protein graph is constructed by considering each amino acid as a node and connecting two nodes if the corresponding amino acids are less than 6 Angstrom. For our experiments, we use the datasets provided by \citet{martinkus22spectre}. 

We follow the evaluation setting of \citet{liao2019gran} using total variation (TV) distance for measuring MMD which is considerably fast compared to using the earth mover's distance (EMD) kernel, especially for large graphs. Moreover, we use the V.U.N. metric following \citet{martinkus22spectre} that measures the proportion of valid, unique, and novel graphs among the generated graphs, where the validness is defined as satisfying the specific property of the dataset explained above. The baseline results are taken from \citet{vignac22digress} or obtained by running the open-source codes.

\vspace{-0.075in}
\paragraph{Implementation details \label{sec:app:exp:general_implementation}}
We follow the standard setting of \citet{martinkus22spectre} using the same data split and evaluation procedures. We report the baseline results taken from \citet{martinkus22spectre} and \citet{vignac22digress}, or the results obtained from running the open-source codes using the hyperparameters given by the original work. We could not report the results of EDP-GNN~\citep{niu20edpgnn} and DiGress~\citep{vignac22digress} on the Proteins dataset as they took more than 2 weeks. For the diffusion models including our proposed method, we set the diffusion steps to 1000 for a fair comparison. 

For our proposed GruM, we train our model for 30,000 epochs for all datasets using a constant learning rate with AdamW optimizer~\citep{loshchilov17adamw} and weight decay $10^{-12}$, applying the exponential moving average (EMA) to the parameters~\citep{song20improved_smld}.
We perform the hyperparameter search explained in Section~\ref{sec:app:details:hyperparameter} for the lowest MMD and highest V.U.N. results. We initialize the node features with the eigenvectors of the graph Laplacian of the adjacency matrices, which we further scale with constant. 
During training (Algorithm~\ref{alg:training}), we sample the noise for the adjacency matrices to be symmetric with zero diagonals. During generation (Algorithm~\ref{alg:sampling}), we generate both the node features and adjacency matrices starting from noise, and we quantize the entries of the resulting adjacency matrices.
Empirically, we found that the entries of the resulting sample lie very close to the desired values 0 and 1, for which the L1 distance between the resulting sample and the quantized sample is smaller than $10^{-2}$.

In Figure~\ref{fig:analysis} (Left), we measure the Spec. MMD and V.U.N. of our method and the baselines as follows: 
For GruM we evaluate the predicted graph mixture. For DiGress, we evaluate the prediction obtained from the predicted probability of each state. For GDSS and ConGress, we evaluate the implicit prediction computed from the estimated score or noise following Eq. (16) of \citet{hoogeboom22edm}. The Spec. MMD and the V.U.N. are measured after quantizing the predicted adjacency matrix with thresholding at 0.5.

\vspace{-0.075in}
\subsection{2D molecule generation \label{sec:app:exp:2d_mol}}

\vspace{-0.05in}
\paragraph{Datasets and evaluation metrics}
We evaluate the quality of generated 2D molecules on two molecule datasets used as benchmarks in \citet{jo22gdss}.
\textbf{QM9}~\cite{ramakrishnan2014qm9} dataset consists of 133,885 molecules with up to 9 heavy atoms of four types. \textbf{ZINC250k}~\cite{irwin2012zinc} dataset consists of 249,455 molecules with up to 38 heavy atoms of 9 types. Molecules in both datasets have 3 edge types, namely single bond, double bond, and triple bond. 
For our experiments, we follow the standard procedure~\cite{shi2020graphaf, luo2021graphdf, jo22gdss} of kekulizing the molecules using the RDKit library~\citep{landrum2016rdkit} and removing the hydrogen atoms from the molecules in the QM9 and ZINC250k datasets. 

We evaluate the models with four metrics: \textbf{Validity} is the percentage of the valid molecules among the generated without any post-hoc correction such as valency correction or edge resampling. \textbf{Fr\'{e}chet ChemNet Distance} (FCD)~\citep{preuer2018frechet} measures the distance between the feature vectors of generated molecules and the test set using ChemNet, evaluating the chemical properties of the molecules. \textbf{Neighborhood subgraph pairwise distance kernel} (NSPDK) MMD~\citep{nspdk} measures the MMD between the generated molecular graphs and the molecular graphs from the test set, assessing the quality of the graph structure. \textbf{Scaffold similarity} (Scaf.) measures the cosine similarity of the frequencies of Bemis-Murcko scaffolds~\citep{bemis96scaf}, evaluating the ability to generate similar substructures. See Section~\ref{sec:app:exp:2d_mol} for more details.
Among these, FCD and NSPDK MMD metrics measure the distribution similarity between the test dataset and generated samples while scaffold similarity evaluates the similarity of the generated scaffolds. The baseline results are taken from \citet{jo22gdss} or obtained by running open-source codes.

\vspace{-0.075in}
\paragraph{Implementation details}
We report the results of the baselines taken from \citet{jo22gdss}, except for the results of the Scaffold similarity (Scaf.) and the results of DiGress, which we obtained by running the open-source codes. Especially, the 2D molecule generation results of DiGress on the QM9 dataset are different compared to the results reported in its original paper, since we have used the preprocessed dataset following the setting of \citet{jo22gdss} for a fair comparison with other baselines. We set the diffusion steps to 1000 for all the diffusion models.

For our GruM, we train our model $\bm{s}_{\theta}$ with a constant learning rate with AdamW optimizer and weight decay $10^{-12}$, applying the exponential moving average (EMA) to the parameters. We perform the hyperparameter search similar to that of the general graph generation tasks. Especially for GruM, we follow \citet{jo22gdss} by using the adjacency matrices in the form of $\bm{A}\in\{0,1,2,3\}^{N\!\times\!N}$ where $N$ is the maximum number of atoms in a molecule and each entries indicating the bond types: 0 for no bond, 1 for the single bond, 2 for the double bond and 3 for the triple bond. Further, we scale the entries with a constant scale of 3 in order to bound the input of the model in the interval $[0,1]$, and rescale the final sample of the generation process to recover the bond types. We also sample the noise for the adjacency matrices to be symmetric with zero diagonals during training. We quantize the entries of the resulting adjacency matrices to obtain the discrete bond types $\{0,1,2,3\}$. Empirically, we found that the entries of the resulting sample lie very close to the desired bond types $\{0,1,2,3\}$, for which the L1 distance between the resulting sample and the quantized sample is approximately 0.

\subsection{3D molecule generation \label{sec:app:exp:3d_mol}}
\vspace{-0.05in}
\paragraph{Datasets and evaluation metrics}
We evaluate the quality of generated 3D molecules on two molecule datasets used as benchmarks in \citet{hoogeboom22edm}.
\textbf{QM9}~\cite{ramakrishnan2014qm9} dataset consists of 133,885 molecules with up to 29 atoms of five types including hydrogen atoms. The node features of the QM9 dataset include the one-hot representation of atom type and atom number. \textbf{GEOM-DRUGS}~\cite{axelrod22geom-drugs} dataset consists of 430,000 molecules with up to 181 atoms of fifteen types including hydrogen atoms. GEOM-DRUGS dataset contains different conformations for each molecule. Among many conformations, the 30 lowest energy conformations for each molecule are retained. For the GEOM-DRUGS dataset, we utilize the one-hot representation of atom type without the atom number.
To evaluate the generated molecules, we measure the \textbf{atom} \textbf{and molecule stability} by predicting the bond type between atoms with the standard bond lengths and then checking the valency.

\begin{table*}[t!]
    \caption{\textbf{2D molecule generation results on the QM9 dataset.} The baseline results are taken from \citet{jo22gdss} or obtained by running the open-source codes. Best results are highlighted in bold.}
\vspace{-0.1in}
    \centering
    \resizebox{\textwidth}{!}{
    \renewcommand{\arraystretch}{1.1}
    \renewcommand{\tabcolsep}{9pt}
    \begin{tabular}{l c c c c c c }
    \toprule
        Method & Valid (\%)$\uparrow$ & FCD $\downarrow$ & NSPDK $\downarrow$ & Scaf. $\uparrow$ & Uniq (\%) $\uparrow$ & Novelty (\%) $\uparrow$ \\
    \midrule
        MoFlow~\citep{zang2020moflow} & 91.36~\small{$\pm$1.23} & \phantom{0}4.467~\small{$\pm$0.595} & 0.017~\small{$\pm$0.003} &  0.1447~\small{$\pm$0.0521} & 98.65~\small{$\pm$0.57} & \textbf{94.72}~\small{$\pm$0.77} \\
        GraphAF`\citep{shi2020graphaf} & 74.43~\small{$\pm$2.55} & \phantom{0}5.625~\small{$\pm$0.259} & 0.021~\small{$\pm$0.003} &  0.3046~\small{$\pm$0.0556} & 88.64~\small{$\pm$2.37} & 86.59~\small{$\pm$1.95} \\
        GraphDF~\citep{luo2021graphdf} & 93.88~\small{$\pm$4.76} & 10.928~\small{$\pm$0.038} & 0.064~\small{$\pm$0.000} &  0.0978~\small{$\pm$0.1058} & 98.58~\small{$\pm$0.25} & 98.54~\small{$\pm$0.48} \\
        \midrule
        EDP-GNN~\citep{niu20edpgnn} & 47.52~\small{$\pm$3.60} & \phantom{0}2.680~\small{$\pm$0.221} & 0.005~\small{$\pm$0.001} &  0.3270~\small{$\pm$0.1151} & \textbf{99.25}~\small{$\pm$0.05} & 86.58~\small{$\pm$1.85} \\
        GDSS~\citep{jo22gdss} & 95.72~\small{$\pm$1.94} & \phantom{0}2.900~\small{$\pm$0.282} & 0.003~\small{$\pm$0.000} & 0.6983~\small{$\pm$0.0197} & 98.46~\small{$\pm$0.61} & 86.27~\small{$\pm$2.29} \\
        DiGress~\citep{vignac22digress} & 98.19~\small{$\pm$0.23} & \phantom{0}\textbf{0.095}~\small{$\pm$0.008} & 0.0003~\small{$\pm$0.000} & 0.9353~\small{$\pm$0.0025} & 96.67~\small{$\pm$0.24} & 25.58~\small{$\pm$2.36} \\
    \midrule
        GruM (ours) & \textbf{99.69}~\small{$\pm$0.19} & \phantom{0}0.108~\small{$\pm$0.006} & \textbf{0.0002}~\small{$\pm$0.000} & \textbf{0.9449}~\small{$\pm$0.0054} & 96.90~\small{$\pm$0.15} & 24.15~\small{$\pm$0.80} \\
    \bottomrule
    \end{tabular}}
    \vspace{-0.05in}
    \label{tab:qm9_variance}
\end{table*}

\begin{table*}[t!]
\vspace{-0.1in}
    \caption{\textbf{2D molecule generation results on the ZINC250k dataset.} The baseline results are taken from \citet{jo22gdss} or obtained by running the open-source codes. Best results are highlighted in bold.}
\vspace{-0.1in}
    \centering
    \resizebox{\textwidth}{!}{
    \renewcommand{\arraystretch}{1.1}
    \renewcommand{\tabcolsep}{9pt}
    \begin{tabular}{l c c c c c c }
    \toprule
        Method & Valid (\%)$\uparrow$ & FCD $\downarrow$ & NSPDK $\downarrow$ & Scaf. $\uparrow$ & Uniq (\%) $\uparrow$ & Novelty (\%) $\uparrow$ \\
    \midrule
        MoFlow~\citep{zang2020moflow} & 63.11~\small{$\pm$5.17} & 20.931~\small{$\pm$0.184} & 0.046~\small{$\pm$0.002} & 0.0133~\small{$\pm$0.0052} & \textbf{99.99}~\small{$\pm$0.01} & \textbf{100.00}~\small{$\pm$0.00} \\
        GraphAF~\citep{shi2020graphaf} & 68.47~\small{$\pm$0.99} & 16.023~\small{$\pm$0.451} & 0.044~\small{$\pm$0.005} & 0.0672~\small{$\pm$0.0156} & 98.64~\small{$\pm$0.69} & 99.99~\small{$\pm$0.01} \\
        GraphDF~\citep{luo2021graphdf} & 90.61~\small{$\pm$4.30} &  33.546~\small{$\pm$0.150} & 0.177~\small{$\pm$0.001} & 0.0000~\small{$\pm$0.0000} & 99.63~\small{$\pm$0.01} & \textbf{100.00}~\small{$\pm$0.00} \\
    \midrule
        EDP-GNN~\cite{niu20edpgnn} & 82.97~\small{$\pm$2.73} &  16.737~\small{$\pm$1.300} & 0.049~\small{$\pm$0.006} & 0.0000~\small{$\pm$0.0000} & 99.79~\small{$\pm$0.08} & \textbf{100.00}~\small{$\pm$0.00} \\
        GDSS~\citep{jo22gdss} & 97.01~\small{$\pm$0.77} &  14.656~\small{$\pm$0.680} & 0.019~\small{$\pm$0.001} & 0.0467~\small{$\pm$0.0054} & 99.64~\small{$\pm$0.13} & \textbf{100.00}~\small{$\pm$0.00} \\
        DiGress~\citep{vignac22digress} & 94.99~\small{$\pm$2.55} & \phantom{0}3.482~\small{$\pm$0.147} & 0.0021~\small{$\pm$0.0004} & 0.4163~\small{$\pm$0.0533} & 99.97~\small{$\pm$0.01} & 99.99~\small{$\pm$0.01} \\
    \midrule
        GruM (ours) & \textbf{98.65}~\small{$\pm$0.25} & \phantom{0}\textbf{2.257}~\small{$\pm$0.084} & \textbf{0.0015}~\small{$\pm$0.0003} & \textbf{0.5299}~\small{$\pm$0.0441} & 99.97~\small{$\pm$0.03} & 99.98~\small{$\pm$0.02} \\
    \bottomrule
    \end{tabular}}
    \vspace{-0.2in}
    \label{tab:zinc_variance}
\end{table*}
\begin{figure*}[t!]
    \caption{\textbf{(Left) Generation results on the Planar dataset.} Best results are highlighted in bold, where smaller MMD and larger V.U.N. indicate better results. \textbf{(Right) Generated graphs by learning the drift.} Visualized graphs are randomly sampled without curation.} \label{fig:ablation_drift}
\vspace{-0.075in}
    \begin{minipage}{0.58\linewidth}
        \centering
        \resizebox{1.0\textwidth}{!}{
\renewcommand{\arraystretch}{1.1}
\renewcommand{\tabcolsep}{8pt}
\begin{tabular}{l c c c c a}
\toprule
    & \multicolumn{5}{c}{Planar}\\
\cmidrule(l{2pt}r{2pt}){2-6}
    & Deg. & Clus. & Orbit & Spec. & V.U.N. \\
\midrule
    Training set & 0.0002 & 0.0310 & 0.0005 & 0.0052 & 100.0 \\
\midrule
    GraphRNN~\citep{you2018graphrnn} & 0.0049 & 0.2779 & 1.2543 & 0.0459 & 0.0  \\
    SPECTRE~\citep{martinkus22spectre} & 0.0005 & 0.0785 & 0.0012 & 0.0112 & 25.0 \\
\midrule
    EDP-GNN~\citep{niu20edpgnn} & 0.0044 & 0.3187 & 1.4986 & 0.0813 & 0.0 \\
    GDSS~\citep{jo22gdss} & 0.0041 & 0.2676 & 0.1720 & 0.0370 & 0.0 \\
    DiGress~\citep{vignac22digress} & \textbf{0.0003} & 0.0372 & \textbf{0.0009} & 0.0106 & 75 \\
\midrule
    Drift & 0.0008 & 0.0845 & 0.0075 & 0.0126 & 15 \\
\midrule
    \textbf{GruM (Ours)} & 0.0005 & \textbf{0.0353} & \textbf{0.0009} & \textbf{0.0062} & \textbf{90.0} \\
\bottomrule
\end{tabular}}
    \end{minipage}
    \hfill
    \begin{minipage}{0.38\linewidth}
        \centering
        \includegraphics[width=1\linewidth]{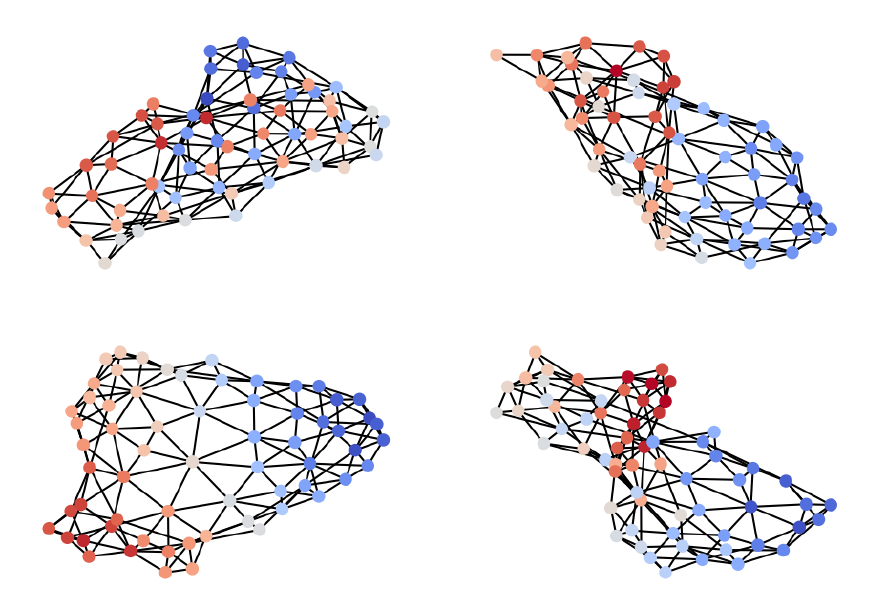}
    \end{minipage}
\vspace{0.1in}
\centering
    \caption{\textbf{MMD results of graph mixture of GruM} through the generative process. Dotted lines indicate MMDs of training set.} \label{fig:early_stop}
\vspace{-0.05in}
    \includegraphics[width=1\linewidth]{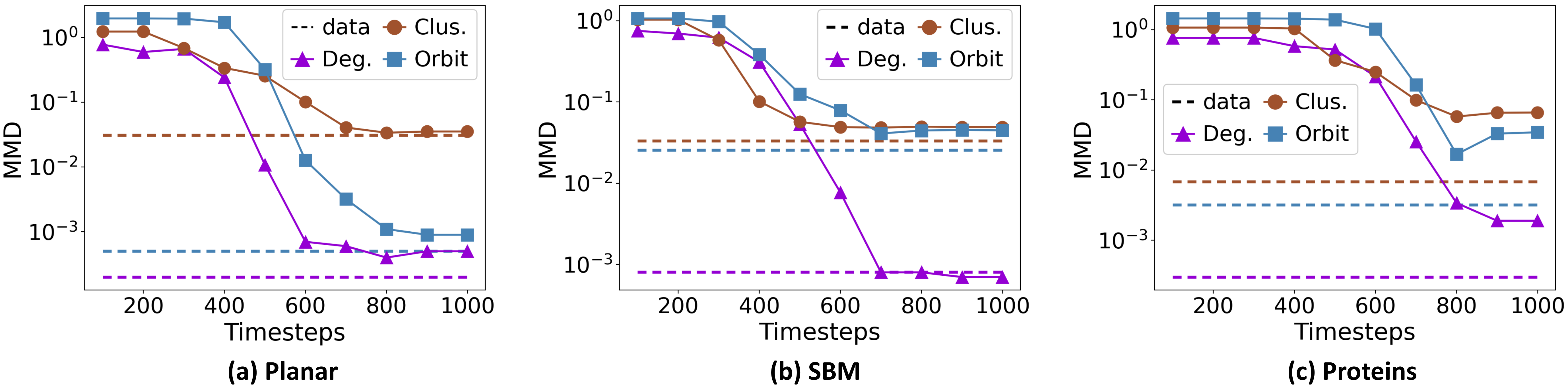}
\vspace{0.2in}
    \includegraphics[width=1\linewidth]{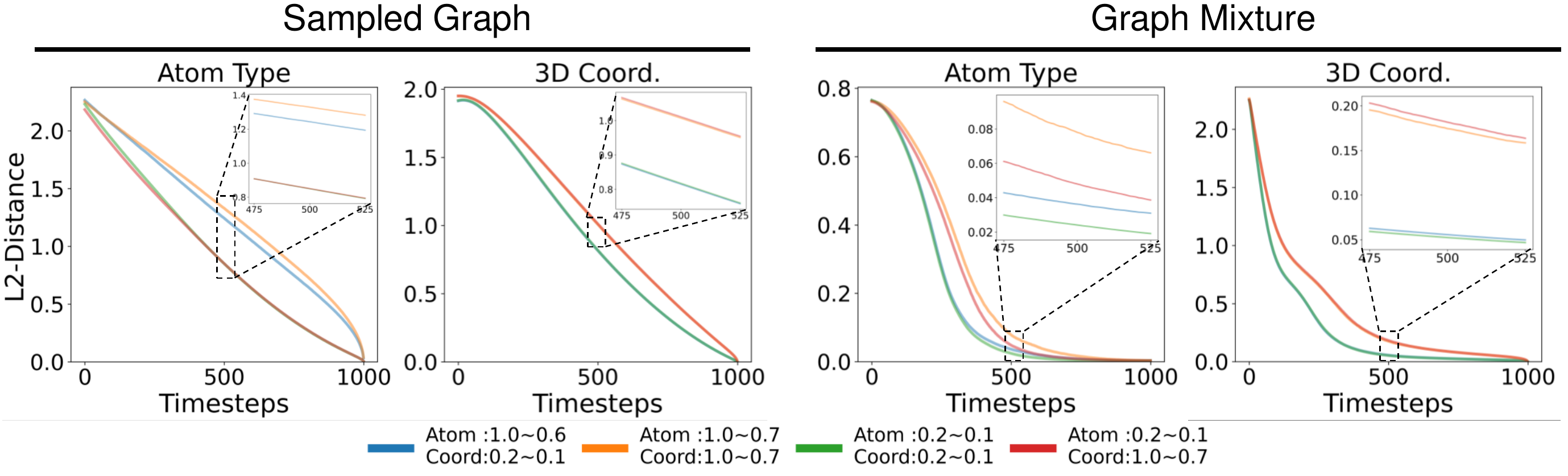}
\vspace{-0.1in}
    \caption{\textbf{Convergence of sampled graphs and graph mixtures with varying $\sigma_0$ and $\sigma_1$ values.}} \label{fig:conv_sigma}
\vspace{-0.2in}
\end{figure*}

\vspace{-0.075in}
\paragraph{Implementation details}
We follow the standard setting of \citet{hoogeboom22edm} for a fair comparison: for the QM9 experiment, we use EGNN with 256 hidden features and 9 layers and train the model, and for the GEOM-DRUGS experiment, we use EGNN with 256 hidden features and 4 layers and train the model. We report the results of the baselines taken from \citet{hoogeboom22edm} and \citet{wu22bridge}. 
In Figure~\ref{tab:3d_mol} (Right), we compute the implicit prediction using the estimated noise following Eq. (16) of \citet{hoogeboom22edm}.

For our GruM, we train our model $\bm{s}_\theta$ for 1,300 epochs with batch size 256 for the QM9 experiment, and for 13 epochs with batch size 64 for the GEOM-DRUGS experiment. We apply EMA to the parameters of the model with a coefficient of 0.999 and use AdamW optimizer with learning rate $10^{-4}$ and weight decay $10^{-12}$. 
The 3D coordinates and charge values are scaled as $\times$4 and $\times$0.1, respectively, and we use the simplified loss with a constant $c=100$. We perform the hyperparameter search with smaller values for coordinates in \{0.1, 0.2, 0.3\} and higher values for node features in \{0.6, 0.7, 0.8, 0.9, 1.0\}. For the generation, we use the Euler-Maruyama predictor.

\vspace{-0.075in}
\subsection{Computing resources}
For all experiments, we use NVIDIA GeForce RTX 3090 and 2080 Ti and implement the source code with PyTorch~\cite{pytorch}.

\vspace{-0.075in}
\section{Additional Experimental Results \label{sec:app:additional}}

\vspace{-0.075in}
\subsection{2D molecule generaation \label{sec:app:additional_2d_mol}}
We provide the standard deviation results in Table~\ref{tab:qm9_variance} and Table~\ref{tab:zinc_variance}. We additionally report the following two metrics: \textbf{Novelty} is the proportion of the molecules generated that are valid and not in the training set, and \textbf{Uniqueness} is the proportion of the molecules generated that are valid and unique, where valid molecules are the ones that do not violate the chemical valency rule.

\begin{figure*}[t]
\centering
\caption{The experimental results for the variant of EDM where it aims to predict the final result (EDM-Var.). \textbf{(Left) Generation results on the 3D molecule QM9 datasets.} Best results are highlighted in bold where the higher stability indicates better results. \textbf{(Right) Convergence of the generative process.} We compare the convergence of the graph mixture from GruM, the implicit prediction computed from the estimated noise of EDM, and the predicted result of EDM-Var. We measure the convergence with L2 distance and further visualize the molecule stability of the predictions through the generative process.} 
\vspace{-0.1in}
\begin{minipage}{0.6\linewidth}
    \resizebox{\textwidth}{!}{
    \renewcommand{\arraystretch}{1.0}
    \renewcommand{\tabcolsep}{10pt}
    \begin{tabular}{l c c}
    \toprule
    & \multicolumn{2}{c}{\phantom{\scriptsize{($|V|\leq 29$)}}QM9 ~\scriptsize{($|V|\leq 29$})}  \\
    \cmidrule(l{2pt}r{2pt}){2-3}
        Method & Atom Stab.(\%) & Mol. Stab.(\%) \\
    \midrule
    G-Schnet~\citep{gebauer19gschnet} & 95.7 & 68.1 \\
    GDM~\citep{hoogeboom22edm} & 97.0 & 63.2 \\
    EDM~\citep{hoogeboom22edm} & 98.7 & 82.0 \\
    Bridge~\citep{wu22bridge} & 98.7 & 81.8 \\
    Bridge+Force~\citep{wu22bridge} & 98.8 & 84.6 \\
    \midrule
    EDM-Var. & 94.02 & 35.95 \\
    \midrule
    \textbf{GruM (Ours)} & \textbf{98.81} & \textbf{87.34}\\
    \bottomrule
    \end{tabular}}
\end{minipage}
\hfill
\begin{minipage}{0.36\linewidth}
    \vspace{0.05in}
    \includegraphics[width=1\linewidth]{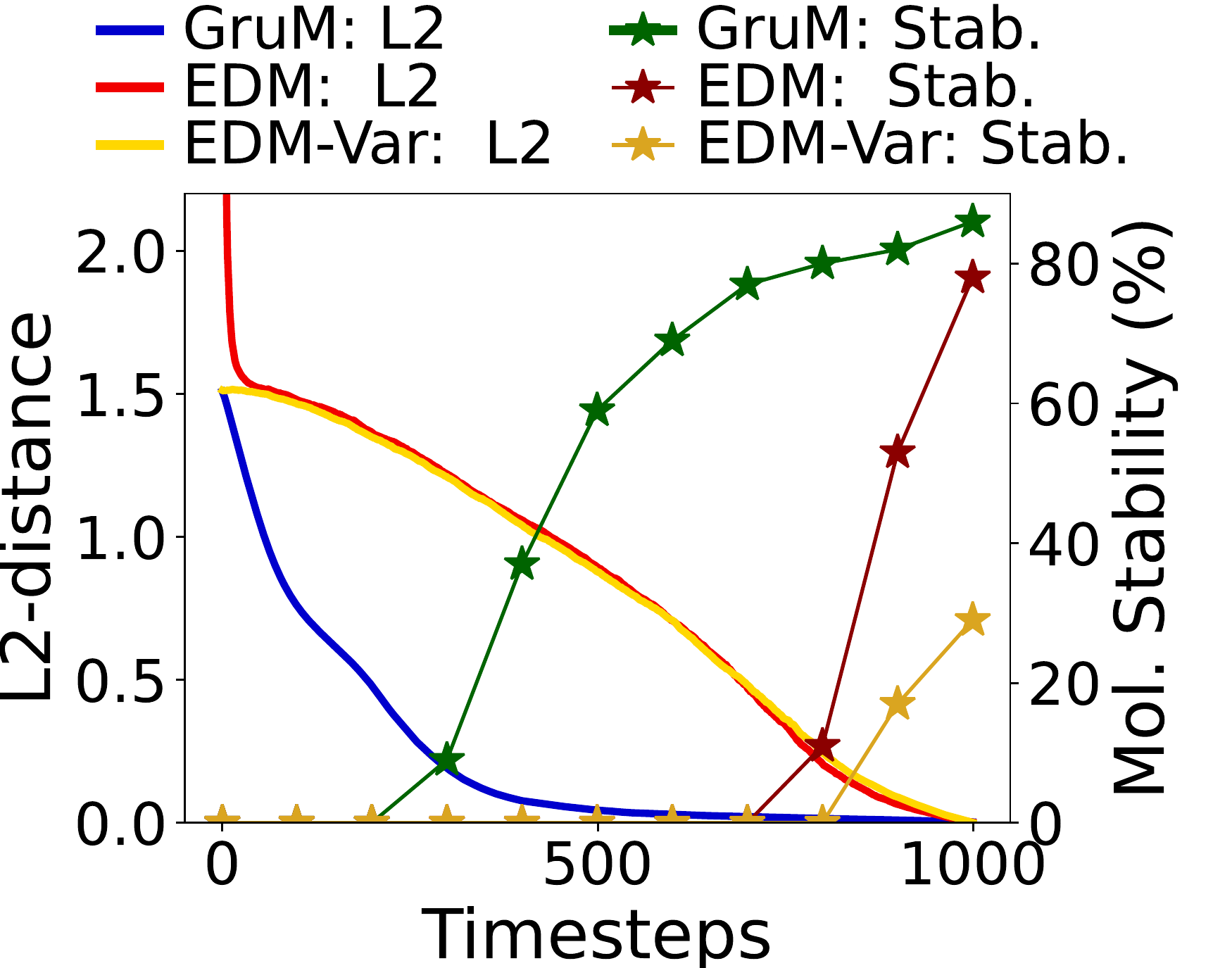}
\end{minipage}
\vspace{-0.2in}
\label{fig:3d_mol_dest}
\end{figure*}

\vspace{-0.075in}
\subsection{Further analysis}

\vspace{-0.075in}
\begin{wrapfigure}{r}{0.25\textwidth}
\vspace{-0.25in}
  \centering
  \includegraphics[width=0.24\textwidth]{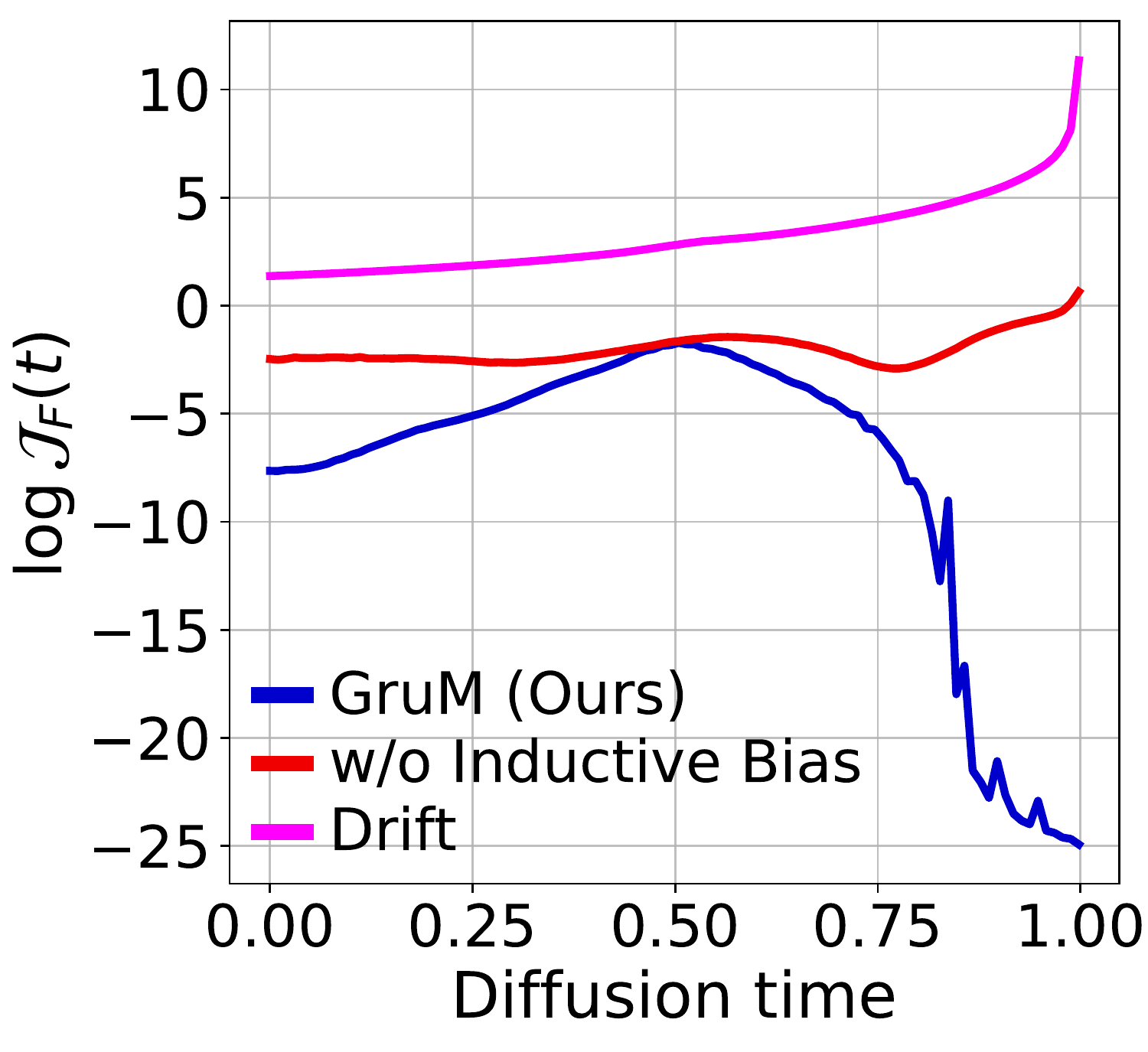}
\vspace{-0.15in}
  \caption{Model complexity comparison of GruM and Drift.}
\vspace{-0.35in}
  \label{fig:complexity}
\end{wrapfigure}
\paragraph{Comparison with learning the drift \label{sec:app:additional:drift}}
To verify that learning the graph mixture as in our GruM is superior compared to learning the drift, we additionally report the generation result of the variant of GruM which learns the drift, similar to \citet{wu22bridge}, on the Planar dataset. Table in Figure~\ref{fig:ablation_drift} shows that learning 
the drift, denoted as Drift in the table, performs poorly generating only 15\% valid, novel, and unique graphs. The generated Planar graphs in Figure~\ref{fig:ablation_drift} demonstrate that learning the drift fails to capture the correct topology.

Further, to verify why learning the drift fails to capture the correct topology, we compare the complexity of the models for learning different objectives. As shown in Figure~\ref{fig:complexity}, the complexity of learning the drift (Drift) is significantly higher than learning the graph mixture (GruM) for all time steps. Moreover, learning the drift is much harder compared to learning the graph mixture without exploiting the graph structure (w/o Inductive Bias). In particular, the complexity gap dramatically increases at the late stage of the diffusion process, because the drift diverges approaching the terminal time while the graph mixture is supported inside the data space, as discussed in Section~\ref{sec:method:advantages}.

\vspace{-0.1in}
\paragraph{Early Stopping for Generative Process \label{sec:app:additional:early_stopping}}
In Figure~\ref{fig:analysis} (Left) and (Middle), the V.U.N. and the MMD results of GruM in the Planar dataset demonstrate that the estimated graph mixture converges to the final result at early sampling steps, accurately capturing both the global topology and local graph characteristics. This allows us to early-stop the diffusion process, which reduces the generation time by up to 20\% on this task. 


We provide additional MMD results of the generative processes in Figure~\ref{fig:early_stop}, which show that the estimated graph mixture converges to the final result around 800 diffusion steps for all datasets.

\paragraph{Role of the diffusion coefficient \label{sec:app:additional:coeff}} 
We can observe that the generative process of GruM is uniquely determined by the constant $\alpha$ and the diffusion coefficient $\sigma_t$. These two coefficients control the convergence behavior of the diffusion process: large $\alpha$ and small $\sigma_t$ lead to a drift with a large norm that forces the trajectory to converge quickly.
Here, we demonstrate the effect of the diffusion coefficient $\sigma_t$ on the convergence of the generative process. Figure \ref{fig:conv_sigma} (Sampled Graph) shows that the smaller values of $\sigma_t$ (i.e. 0.2$\sim$0.1) lead to faster convergence of the trajectory to the final result, compared to the larger $\sigma_t$. 
This is due to the fast convergence of each bridge process with small $\sigma_t$.
Especially, as shown in Figure \ref{fig:conv_sigma} (Graph Mixture), large $\sigma_t$ for the continuous feature (i.e., 3D coordinates) leads to slower convergence of the graph mixture since it destroys the topology of graphs and makes it hard to predict the final result.

\paragraph{Graph prediction through EDM \label{sec:app:additional:explicit_EDM}}
Additionally, we compare our GruM with the variant of EDM~\cite{hoogeboom22edm} which learns to predict the final result of the denoising process instead of learning the noise. Table of Figure~\ref{fig:3d_mol_dest} shows the generation result of this variant, denoted as EDM-Var., on the 3D molecule QM9 dataset. EDM-Var. exhibits the lowest atom stability and extremely low molecule stability of less than 40\%, which performs significantly worse than GruM as well as the original EDM. This is because EDM-Var. depends on a single deterministic prediction during the generative process, and the inaccurate prediction of the final result at the early step of the generative process leads the process in the wrong direction resulting in invalid molecules, as discussed in the Introduction and Section~\ref{sec:method:design}.

On the other hand, our GruM predicts the final grpah of the generative process using the graph mixture which represents the probable graph as a weighted mean of the data, thereby guiding the process in the right direction resulting in valid molecules with correct topology.
We further provide the convergence results of EDM-Var. in Figure~\ref{fig:3d_mol_dest}, which demonstrates that the prediction of GruM converges significantly faster than that of EDM and EDM-Var. The inaccurate prediction of EDM-Var. results in slower convergence and low molecule stability.

\paragraph{Analysis on the model architecture \label{sec:app:additional:arch}}
As shown in Table~\ref{tab:general_graph_transformer} and \ref{tab:2d_mol_graph_transformer}, GDSS using graph transformer architecture shows improved performance over original GDSS but is still outperformed by our GruM with a large margin in V.U.N, FCD, and NSPDK. These results verify that the superior performance of GruM comes from its ability to accurately model the topology of the final graph to be generated.

\begin{table*}[t!]
\vspace{0.1in}
    \caption{\textbf{General graph generation results with GDSS using graph transformer.}}
    \label{tab:general_graph_transformer}
\vspace{-0.1in}
    \centering
    \resizebox{\textwidth}{!}{
    \renewcommand{\arraystretch}{1.1}
    \renewcommand{\tabcolsep}{8pt}
    \begin{tabular}{l c c c c a c c c c a}
    \toprule
        &
        \multicolumn{5}{c}{Planar} &
        \multicolumn{5}{c}{SBM} \\
    \cmidrule(l{2pt}r{2pt}){2-6}    
    \cmidrule(l{2pt}r{2pt}){7-11}
        &
        \multicolumn{5}{c}{Synthetic, $|V| = 64$} &
        \multicolumn{5}{c}{Synthetic, $44\leq|V|\leq187$} \\
    \cmidrule(l{2pt}r{2pt}){2-6}    
    \cmidrule(l{2pt}r{2pt}){7-11}
        & Deg. & Clus. & Orbit & Spec. & V.U.N.  & Deg. & Clus. & Orbit & Spec. & V.U.N. \\
    \midrule
        Training set & 0.0002 & 0.0310 & 0.0005 & 0.0052 & 100.0 & 0.0008 & 0.0332 & 0.0255 & 0.0063 & 100.0 \\
    \midrule
        GDSS & 0.0041 & 0.2676 & 0.1720 & 0.0370 & 0.0 & 0.0212 & 0.0646 & 0.0894 & 0.0128 & 5.0 \\
        GDSS+Transformer & 0.0036 & 0.1206 & 0.0525 & 0.0137 & 5.0 & 0.0411 & 0.0565 & 0.0706 & 0.0074 & 27.5 \\
        ConGress & 0.0048 & 0.2728 & 1.2950 & 0.0418 & 0.0 & 0.0273 & 0.1029 & 0.1148 & - & 0.0  \\
        DiGress & \textbf{0.0003} & 0.0372 & \textbf{0.0009} & 0.0106 & 75 & 0.0013 & 0.0498 & 0.0434 & 0.0400 & 74 \\
    \midrule
        \textbf{GruM (Ours)} & 0.0005 & \textbf{0.0353} & \textbf{0.0009} & \textbf{0.0062} & \textbf{90.0} & \textbf{0.0007} & \textbf{0.0492} & 0.0448 & \textbf{0.0050} & \textbf{85.0} \\
    \bottomrule
    \end{tabular}}
    \vspace{-0.075in}
\end{table*}
\begin{table*}[t!]
    \caption{\textbf{2D molecule generation results with GDSS using graph transformer.}} \label{tab:2d_mol_graph_transformer}
    \vspace{-0.075in}
    \centering
    \resizebox{\textwidth}{!}{
    \renewcommand{\arraystretch}{1.1}
    \renewcommand{\tabcolsep}{8pt}
    \begin{tabular}{l c c c c c c c c}
    \toprule
    & \multicolumn{4}{c}{\phantom{\scriptsize{($|V|\leq 9$)}}QM9 ~\scriptsize{($|V|\leq 9$)}} & \multicolumn{4}{c}{\phantom{\scriptsize{($|V|\leq 38$)}}ZINC250k \scriptsize{($|V|\leq 38$)}} \\
    \cmidrule(l{2pt}r{2pt}){2-5}
    \cmidrule(l{2pt}r{2pt}){6-9}
        Method & Valid (\%)$\uparrow$ & FCD$\downarrow$ & NSPDK$\downarrow$ & Scaf.$\uparrow$ & Valid (\%)$\uparrow$ & FCD$\downarrow$ & NSPDK$\downarrow$ &  Scaf.$\uparrow$ \\
    \midrule
    Training set & 100.0 & 0.0398 & 0.0001 & 0.9719 & 100.0 & 0.0615 & 0.0001 & 0.8395 \\
    \midrule
        GDSS & 95.72 & 2.900 & 0.0033 & 0.6983 & 97.01 & 14.656 & 0.0195 & 0.0467 \\
        GDSS+Transformer & 99.68 & 0.737 & 0.0024 & 0.9129 & 96.04 & \phantom{0}5.556 & 0.0326 & 0.3205 \\
        DiGress & 98.19 & \textbf{0.095} & 0.0003 & 0.9353 & 94.99 & \phantom{0}3.482 & 0.0021 & 0.4163 \\
    \midrule
        \textbf{GruM (Ours)} & \textbf{99.69} & 0.108 & \textbf{0.0002} & \textbf{0.9449} & \textbf{98.65} & \phantom{0}\textbf{2.257} & \textbf{0.0015} & \textbf{0.5299} \\
    \bottomrule
    \end{tabular}}
\end{table*}

\newpage
\section{Visualization~\label{sec:app:vis}}
In this section, we visualize the generated graphs and molecules of GruM, and further provide visualization of the diffusion processes for diverse generation tasks.

\subsection{Generated samples of GruM~\label{sec:app:vis:samples}}
\begin{figure*}[ht]
\vspace{0.1in}
    \centering
    \includegraphics[width=0.95\linewidth]{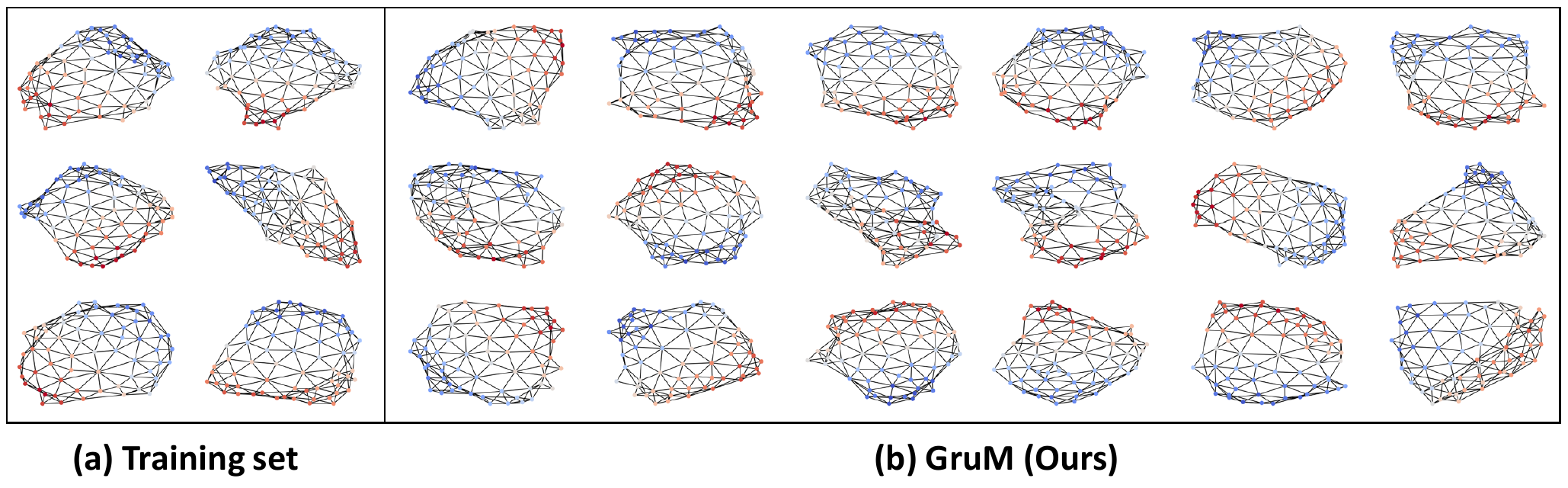}
    \caption{\textbf{Visualization of graphs from the Planar dataset and the generated graphs of GruM.}}
    \label{fig:planar}
\end{figure*}

\begin{figure*}[ht]
\vspace{0.1in}
    \centering
    \includegraphics[width=0.95\linewidth]{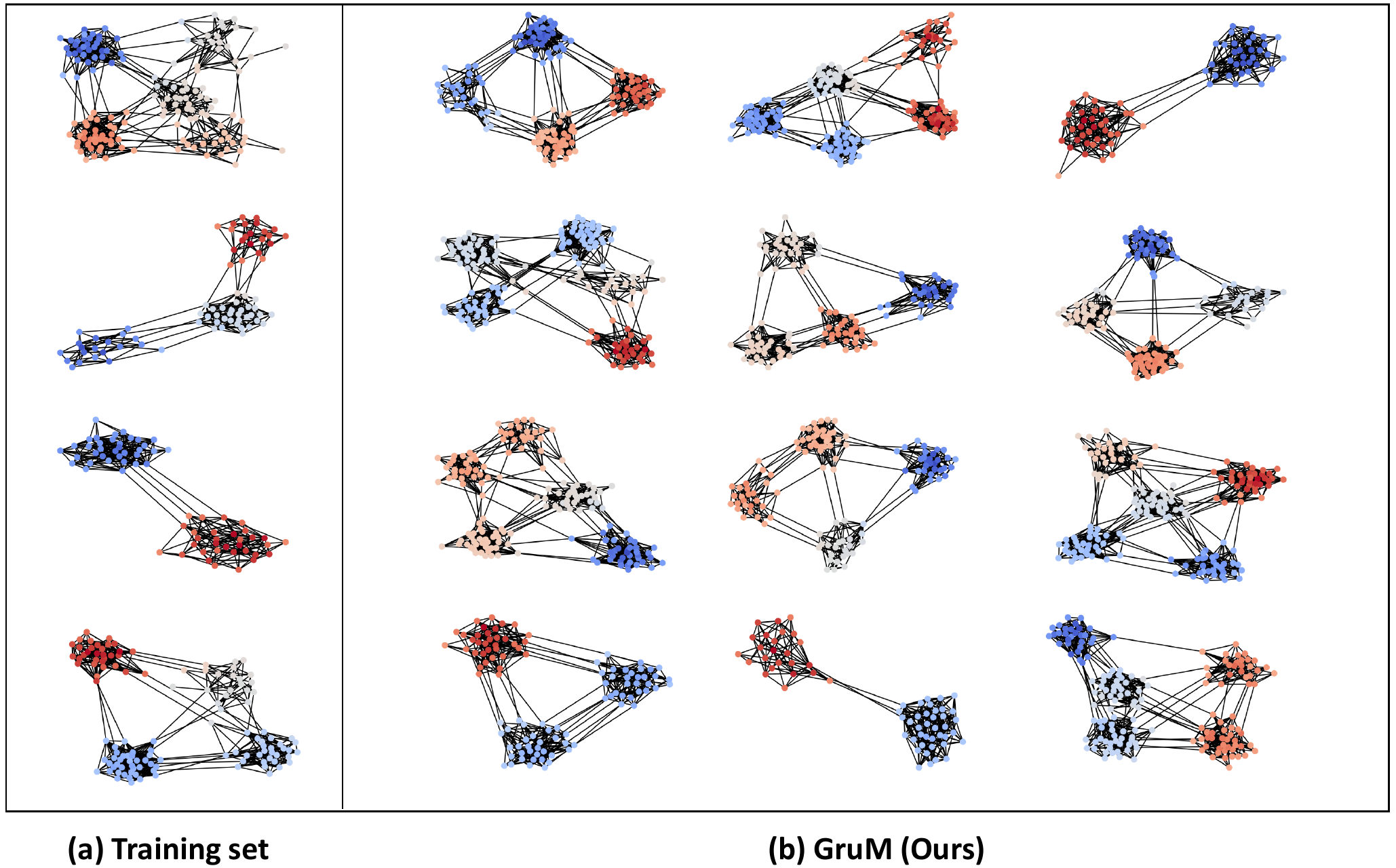}
    \caption{\textbf{Visualization of graphs from the SBM dataset and the generated graphs of GruM.}}
    \label{fig:sbm}
\end{figure*}

\begin{figure*}[ht!]
\vspace{0.1in}
    \centering
    \includegraphics[width=0.95\linewidth]{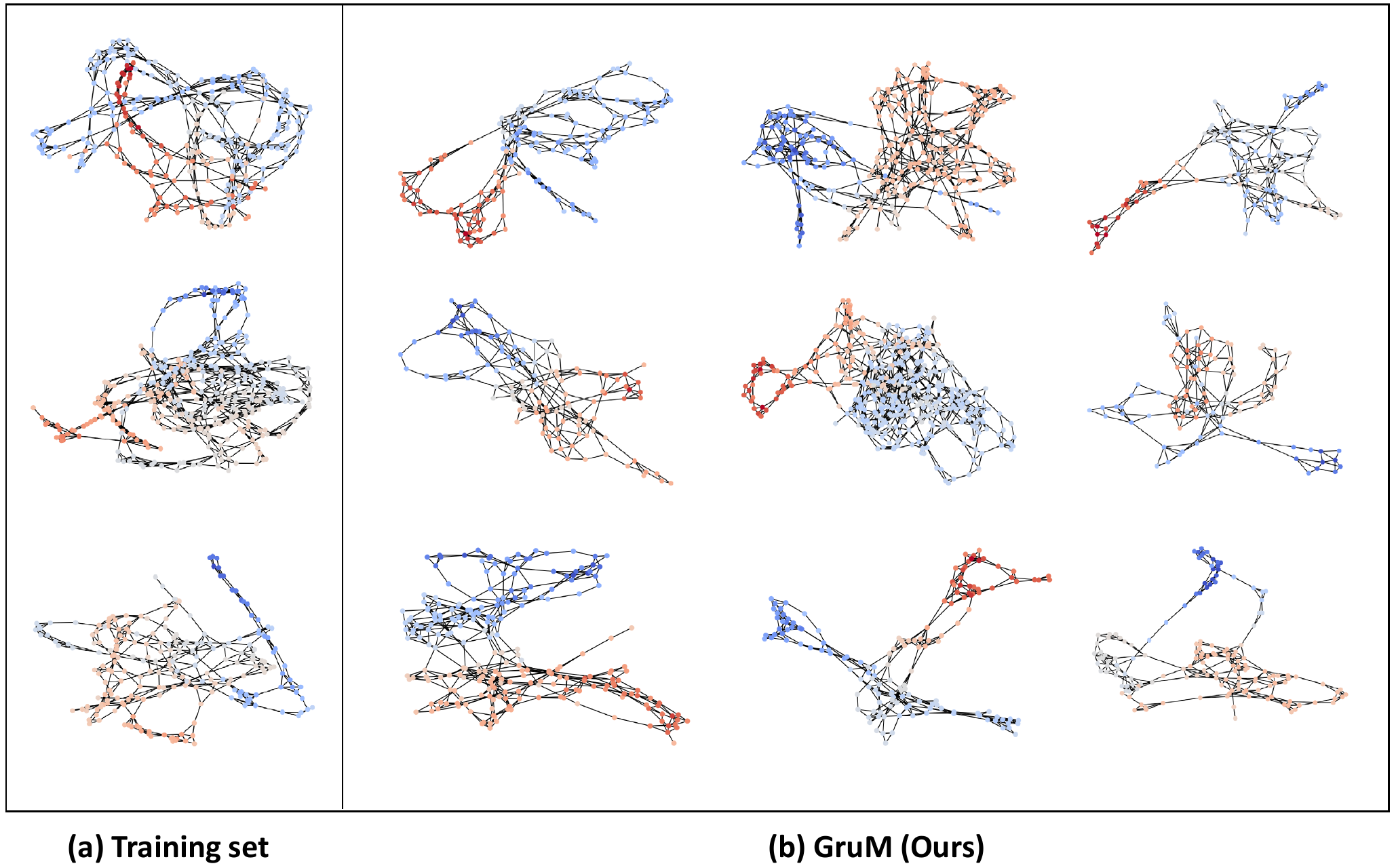}
    \caption{\textbf{Visualization of graphs from the Proteins dataset and the generated graphs of GruM.}}
    \label{fig:proteins}
\end{figure*}

\paragraph{General graphs}
Graphs from the training set and the generated graphs of GruM are visualized in Figure~\ref{fig:planar}, \ref{fig:sbm}, and \ref{fig:proteins}. The visualized graphs are randomly selected from the training set and the generated graph set. Note that we visualize the entire graph for the Proteins dataset, unlike \citet{martinkus22spectre} which visualizes the largest connected component since it fails to consistently generate connected graphs. For GruM, we found that 92\% of the generated Proteins graphs are connected.

\newpage

\begin{figure*}[ht]
\vspace{0.3in}
    \centering
    \includegraphics[width=0.98\linewidth]{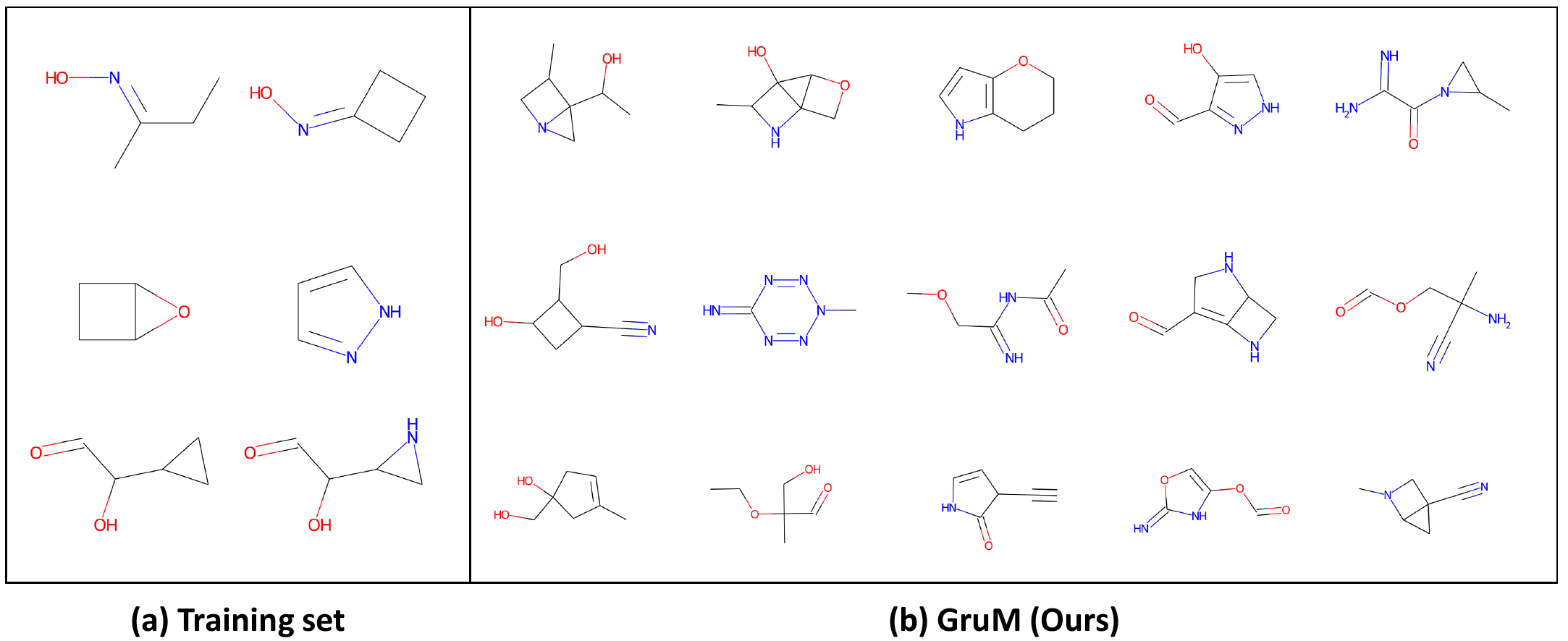}
    \caption{\textbf{Visualization of molecules from the QM9 dataset and the generated molecules of GruM for the 2D molecule generation experiment.}}
    \label{fig:qm9}
\end{figure*}

\begin{figure*}[ht]
\vspace{0.2in}
    \centering
    \includegraphics[width=0.98\linewidth]{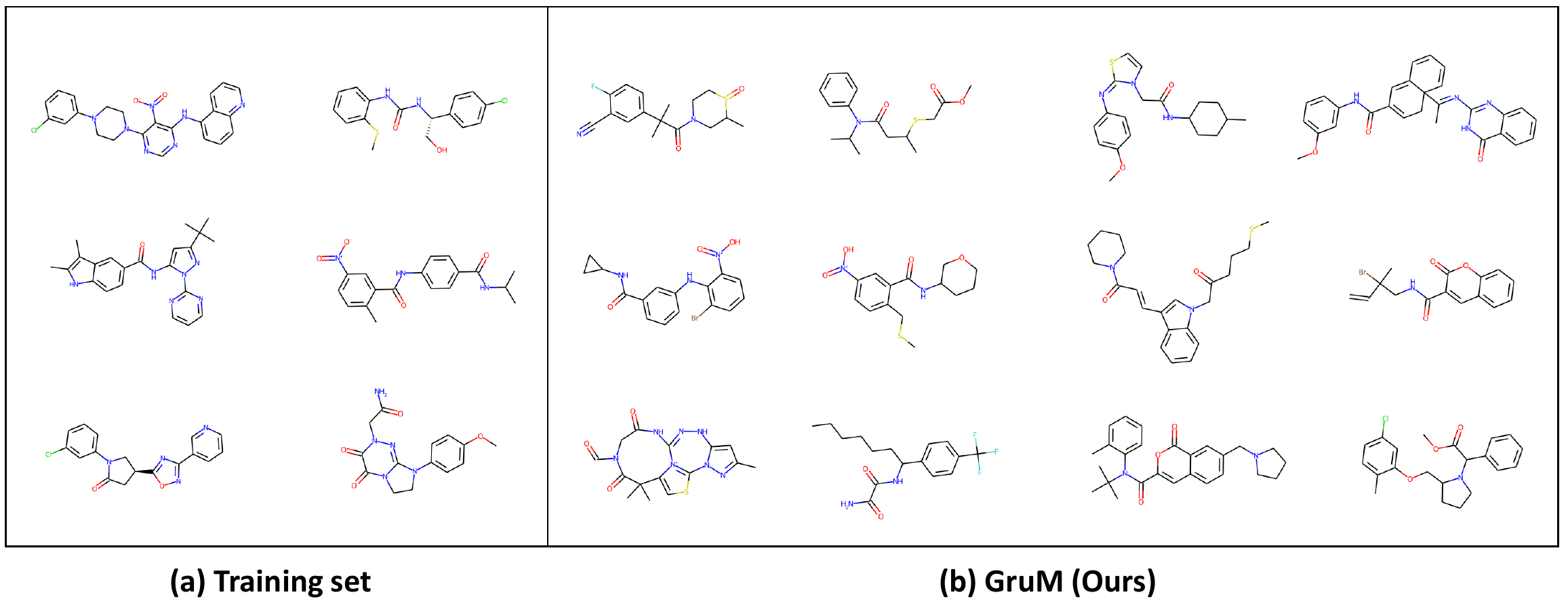}
    \caption{\textbf{Visualization of the molecules from the ZINC250k dataset and the generated molecules of GruM for the 2D molecule generation experiment.}}
    \label{fig:zinc}
\end{figure*}

\FloatBarrier
\newpage
\paragraph{2D molecules}
We provide the visualization of the molecules from the training set and the generated 2D molecules in Figure~\ref{fig:qm9} and \ref{fig:zinc}. These molecules are randomly selected from the training set and the generated molecule set.

\begin{figure*}[h]
\vspace{0.4in}
    \centering
    \includegraphics[width=1\linewidth]{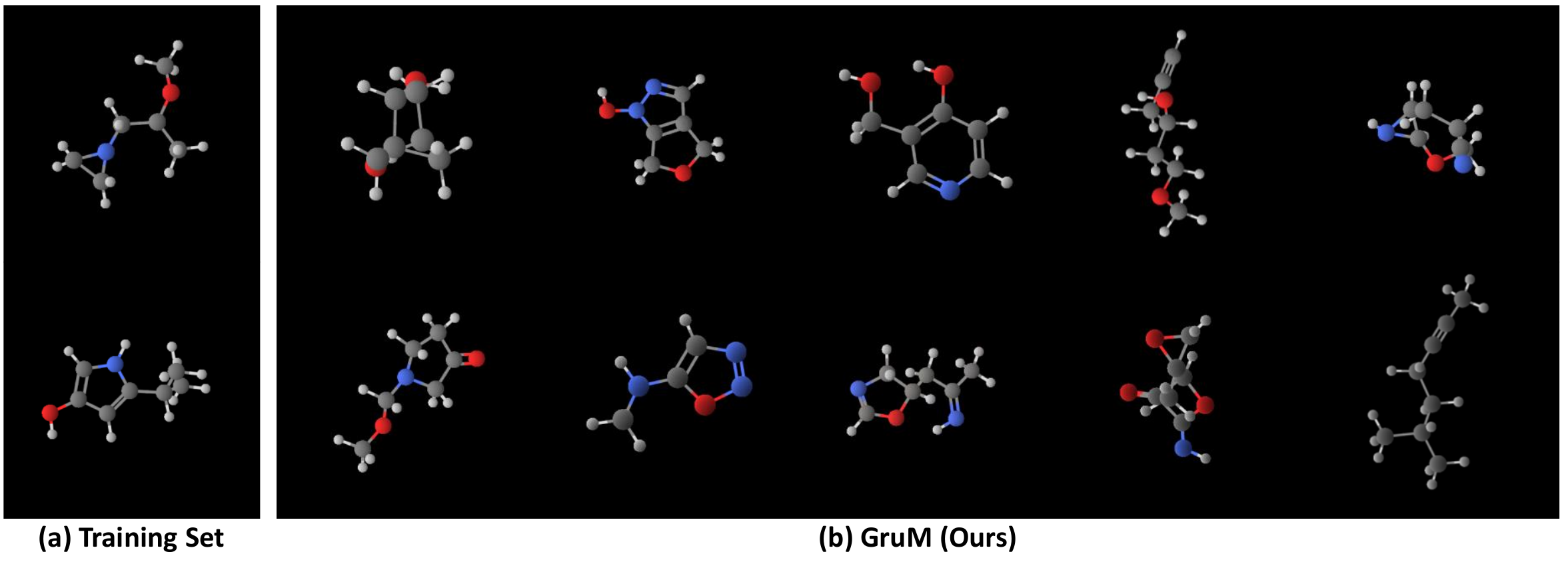}
    \vspace{-0.2in}
    \caption{\textbf{Visualization of the molecules from the QM9 dataset and the generated molecules of GruM for the 3D molecule generation experiment.}}
    \label{fig:qm9_3d}
\end{figure*}

\begin{figure*}[h]
\vspace{0.4in}
    \centering
    \includegraphics[width=1\linewidth]{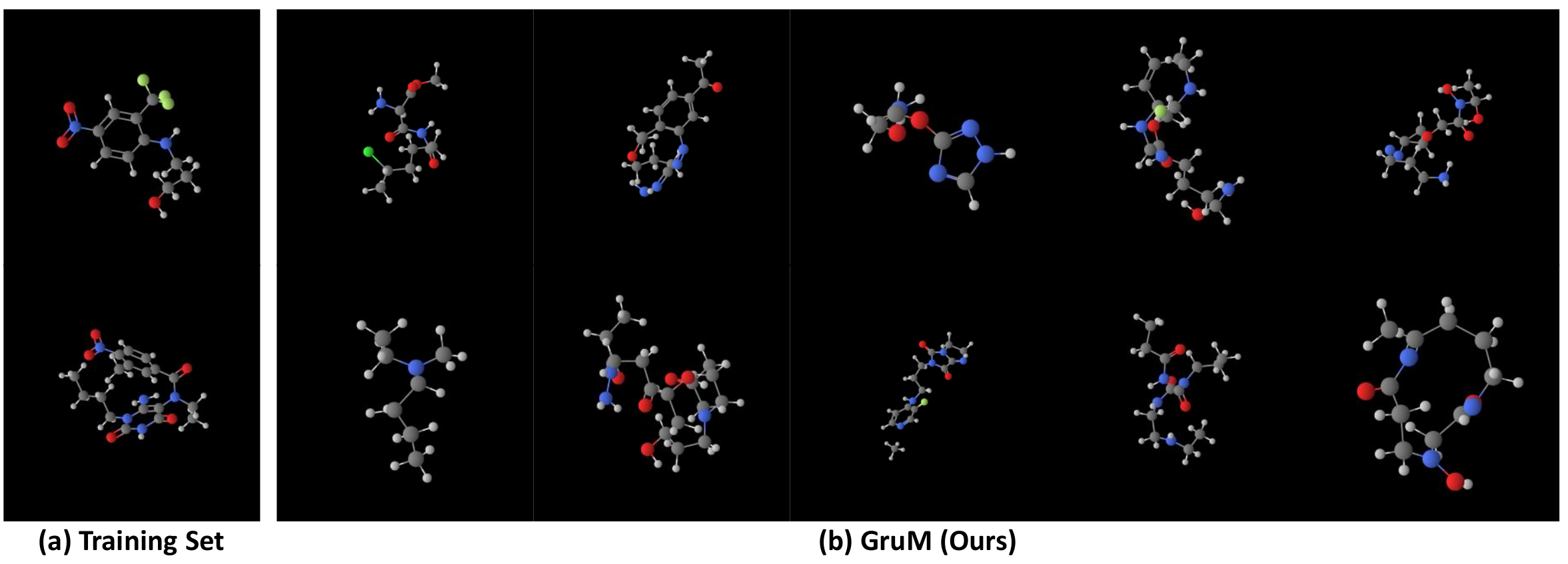}
    \vspace{-0.2in}
    \caption{\textbf{Visualization of the molecules from the GEOM-DRUGS dataset and the generated molecules of GruM for the 3D molecule generation experiment.}}
    \label{fig:geom_3d}
\end{figure*}

\newpage
\begin{wraptable}{t}{0.3\textwidth}
    \centering
    \caption{Fraction of connected graphs on GEOM-DRUGS experiment.}
    \vspace{-0.1in}
    \begin{adjustbox}{width=\linewidth}
    \begin{tabular}{lc}
        \toprule
        Methods & Connected (\%) \\
        \midrule
        EDM  & 37.70 \small{$\pm$0.79} \\
        \textbf{GruM (Ours)} & \textbf{56.57} \small{$\pm$0.31} \\
        \bottomrule
    \end{tabular}
    \end{adjustbox}
    \label{tab:geom_conn}
\end{wraptable}
\paragraph{3D molecules}
We visualize the generated molecules for the 3D molecule generation experiment in Figure~\ref{fig:qm9_3d} and \ref{fig:geom_3d}. Note that the visualized molecules are all stable. For the GEOM-DRUGS experiment, we observe that a few of the generated molecules are not connected as pointed out in \citet{hoogeboom22edm}. To measure how many graphs are connected, we report the fraction of the connected graphs, taking the average of 3 different runs. Table~\ref{tab:geom_conn} shows that GruM can generate a significantly larger number of connected molecules compared to EDM~\cite{hoogeboom22edm}.

\newpage
\subsection{Generative process of GruM~\label{sec:app:vis:process}}
\begin{figure*}[ht!]
\vspace{-0.1in}
    \centering
    \includegraphics[width=1\linewidth]{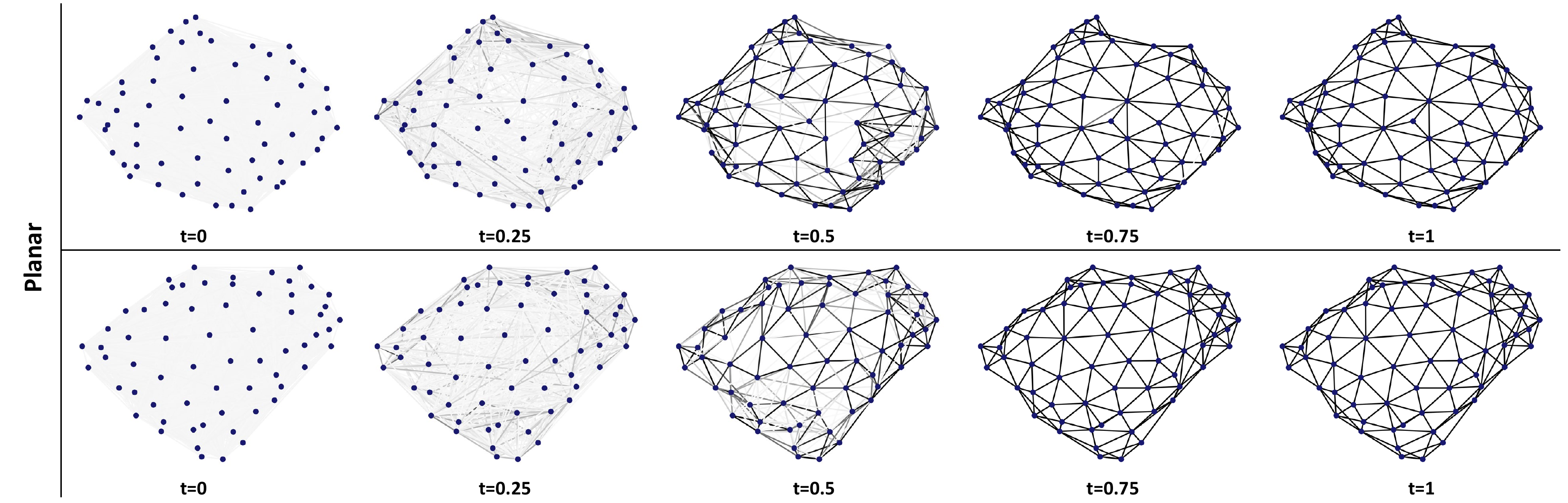}
    \vspace{-0.2in}
    \caption{\textbf{Visualization of the generative process of GruM.} We visualize the graph mixture from GruM on the Planar dataset.}
    \label{fig:vis_process_planar}
\end{figure*}

\begin{figure*}[ht!]
\vspace{-0.1in}
    \centering
    \includegraphics[width=1\linewidth]{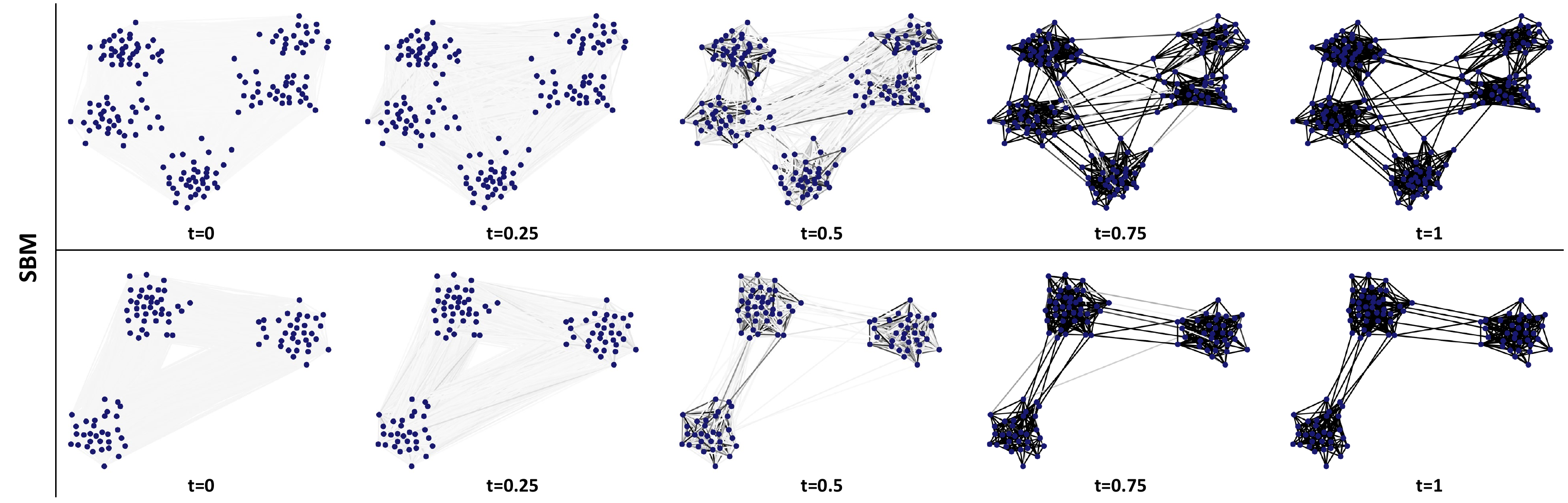}
\vspace{-0.2in}
    \caption{\textbf{Visualization of the generative process of GruM.} We visualize the graph mixture from GruM on the SBM dataset.}
    \label{fig:vis_process_sbm}
\end{figure*}

\begin{figure*}[ht!]
\vspace{-0.1in}
    \centering
    \includegraphics[width=1\linewidth]{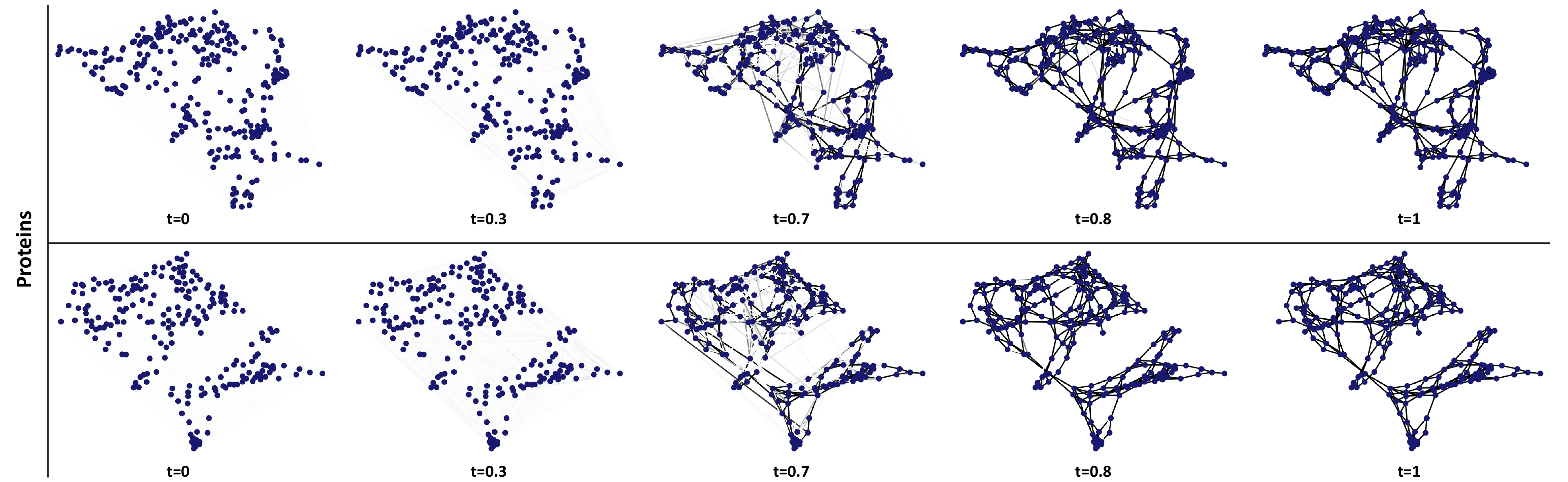}
\vspace{-0.2in}
    \caption{\textbf{Visualization of the generative process of GruM.} We visualize the graph mixture from GruM on the Proteins dataset.}
    \label{fig:vis_process_proteins}
\end{figure*}

\begin{figure*}[t!]
    \centering
    \includegraphics[width=1\linewidth]{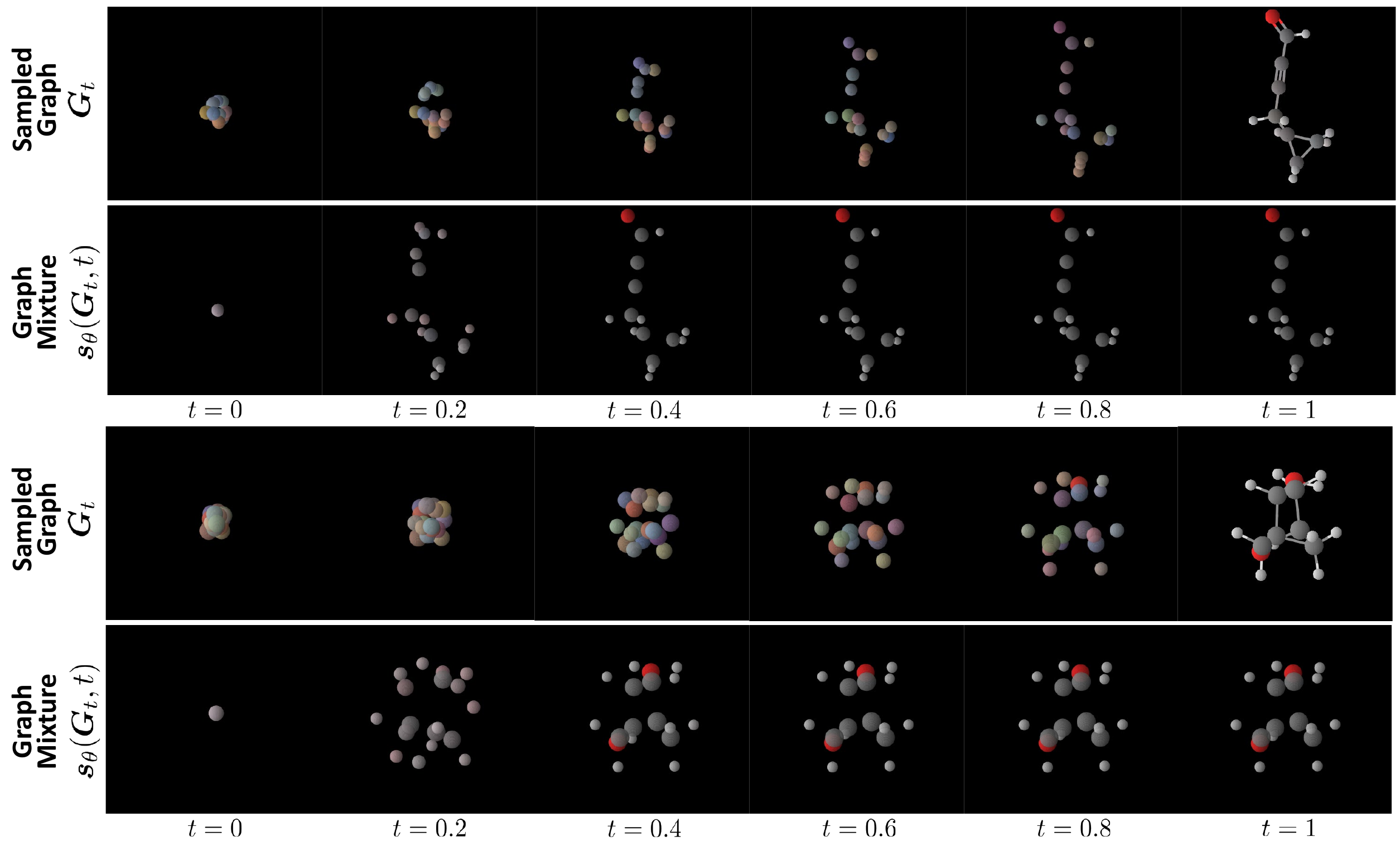}
\par
\vspace{0.37in}
    \includegraphics[width=1\linewidth]{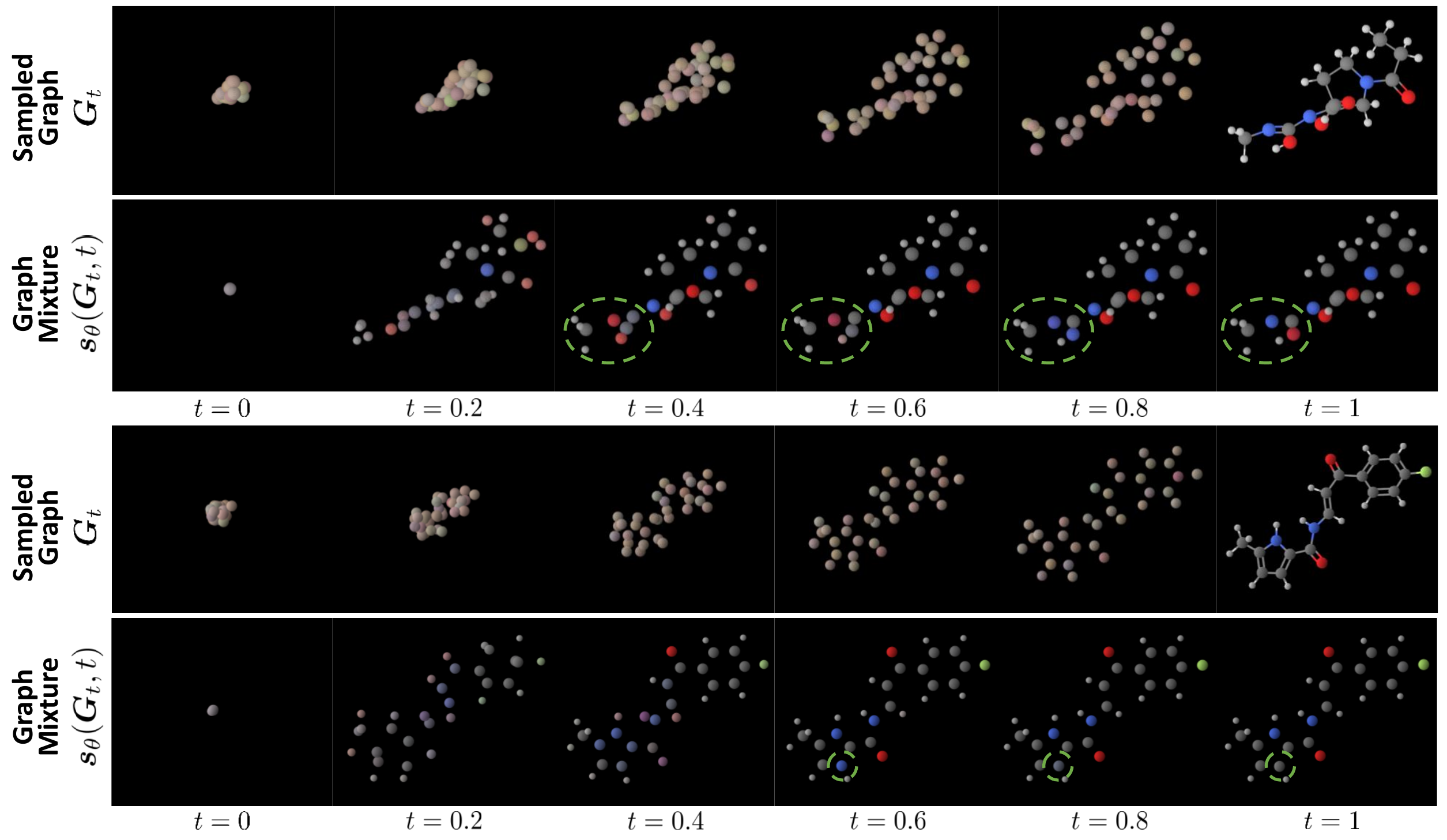}
    \caption{\textbf{Visualizations of the 3D molecule generative process} of GruM on QM9 dataset (Top) and GEOM-DRUGS dataset (Bottom). For each dataset, we visualize the trajectory $\bm{G}_t$ in the first row, and we visualize the estimated graph mixtures from GruM in the second row. Note that the visualized molecules are stable. 
    The atom types and the 3D coordinates of the atoms inside the green circles are calibrated after the convergence of the graph mixtures, where the convergence is achieved at an early stage.}
    \label{fig:3d_process}
\vspace{-0.6in}
\end{figure*}

Here we visualize the generative process of GruM. We visualize the generative process of general graphs in Figure~\ref{fig:vis_process_planar}, \ref{fig:vis_process_sbm}, and \ref{fig:vis_process_proteins}. We also visualize the generative process of the 3D molecules in Figure~\ref{fig:3d_process}.
We further provide the animation of the generative process in https://github.com/harryjo97/GruM.

\newpage
\section{Limitation} \label{sec:app:limitation}

\paragraph{Limitation}
We proposed a novel diffusion-based graph generation framework that directly predicts the final graph of the generative process as a weighted mean of data, thereby accurately capturing the valid structures and the topological properties. We have shown that our framework is able to generate graphs with correct topology for diverse graph generation tasks, including 2D/3D molecular generation, on which ours significantly outperforms previous graph generation methods. While GruM shows superior performance, future work would benefit from improving our framework.

First, the likelihood of the generative process of GruM cannot be directly computed from the training objective. In order to compute the likelihood, one could derive an associated probability flow ODE of GruM as described in Section~\ref{sec:app:derivation:probability_flow}, but this requires training an additional model for estimating the reverse graph mixture. 

Furthermore, the proposed framework is focused on unconditional graph generation tasks. We could design a conditional framework of GruM by training a model $\bm{s}_{\theta}(\bm{G}_t,t,\bm{c})$ for a given condition (i.e., class label) $\bm{c}$ for estimating the $\bm{c}$-conditional graph mixture defined as follows:
\begin{align}
    \bm{D}^{\Pi^{\ast}_{\bm{c}}}\, (\bm{G}_t,t) \coloneqq \int \bm{g} \frac{p^{\bm{g}}_t(\bm{G}_t)}{p_t(\bm{G}_t)} \Pi^{\ast}_{\bm{c}}(\mathrm{d}\bm{g}), \quad \Pi^{\ast}_{\bm{c}}\!\coloneqq\{\bm{g}:\bm{g}\sim\Pi^{\ast}\! \text{ with label } \bm{c}\} .
\end{align}
Intuitively, the generative process of the modified OU bridge mixture, for which the graph mixture is replaced by $\bm{D}^{\Pi^{\ast}_{\bm{c}}}\, (\bm{G}_t,t)$ is guided by the conditional graph mixture that terminates in the conditioned distribution $ \Pi^{\ast}_{\bm{c}}$. We leave this conditional framework as future work.

\end{document}